# Not cheating on the Turing Test: towards grounded language learning in Artificial Intelligence

by

Lize Alberts

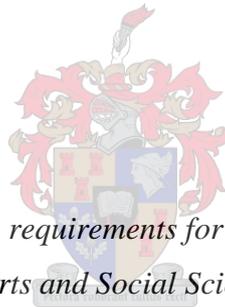

*Thesis presented in fulfilment of the requirements for the degree of Master of Arts Philosophy (Thesis) in the Faculty of Arts and Social Sciences at Stellenbosch University*

Supervisor: Prof. J.P. Smit

December 2020

## Declaration

By submitting this thesis electronically, I declare that the entirety of the work contained therein is my own, original work, that I am the sole author thereof (save to the extent explicitly otherwise stated), that reproduction and publication thereof by Stellenbosch University will not infringe any third party rights and that I have not previously in its entirety or in part submitted it for obtaining any qualification.

December 2020

# Abstract

In this thesis, I carry out a novel and interdisciplinary analysis into various complex factors involved in human natural-language acquisition, use and comprehension, aimed at uncovering some of the basic requirements for if we were to try and develop artificially intelligent (AI) agents with similar capacities. Inspired by a recent publication wherein I explored the complexities and challenges involved in enabling AI systems to deal with the *grammatical* (i.e. syntactic and morphological) irregularities and ambiguities inherent in natural language (Alberts, 2019), I turn my focus here towards appropriately inferring the content of symbols themselves—as 'grounded' in real-world percepts, actions, and situations.

I first introduce the key theoretical problems I aim to address in theories of mind and language. For background, I discuss the co-development of AI and the controverted strands of computational theories of mind in cognitive science, and the *grounding problem* (or 'internalist trap') faced by them. I then describe the approach I take to address the grounding problem in the rest of the thesis. This proceeds in chapter I.

To unpack and address the issue, I offer a critical analysis of the relevant theoretical literature in philosophy of mind, psychology, cognitive science and (cognitive) linguistics in chapter II. I first evaluate the major philosophical/psychological debates regarding the nature of concepts; theories regarding how concepts are acquired, used, and represented in the mind; and, on that basis, offer my own account of conceptual structure, grounded in current (cognitively plausible) connectionist theories of thought. To further explicate how such concepts are acquired and communicated, I evaluate the relevant embodied (e.g. cognitive, perceptive, sensorimotor, affective, etc.) factors involved in grounded human (social) cognition, drawing from current scientific research in the areas of 4E Cognition and social cognition. On that basis, I turn my focus specifically towards grounded theories of language, drawing from the cognitive linguistics programme that aims to develop a naturalised, cognitively plausible understanding of human concept/language acquisition and use. I conclude the chapter with a summary wherein I integrate my findings from these various disciplines, presenting a general theoretical basis upon which to evaluate more practical considerations for its implementation in AI—the topic of the following chapter.

In chapter III, I offer an overview of the different major approaches (and their integrations) in the area of Natural Language Understanding in AI, evaluating their respective strengths and shortcomings in terms of specific models. I then offer a critical summary wherein I contrast and contextualise the different approaches in terms of the more fundamental theoretical convictions they seem to reflect.

On that basis, in the final chapter, I re-evaluate the aforementioned grounding problem and the different ways in which it has been interpreted in different (theoretical and practical) disciplines, distinguishing between a stronger and weaker reading. I then present arguments for why implementing the stronger version in AI seems, both practically and theoretically, problematic. Instead, drawing from the theoretical insights I gathered, I consider some of the key requirements for 'grounding' (in the weaker sense) as much as possible of natural language use with robotic AI agents, including implementational constraints that might need to be put in place to achieve this. Finally, I evaluate some of the key challenges that may be involved, if indeed the aim were to meet all the requirements specified.

# Opsomming

In hierdie tesis stel ek 'n oorspronklike en interdissiplinêre ondersoek in na die verskeie komplekse faktore wat betrokke is by die mens se aanleer, gebruik en begrip van natuurlike taal, met oog op die identifikasie van die basiese vereistes om kunsmatig intelligente (KI)-agente met soortgelyke vermoëns te ontwikkel. Geïnspireer deur 'n onlangse publikasie waarin ek die kompleksiteite en uitdagings oorweeg wat betrokke is in die prosessering van die *grammatikale* (d.i. sintaktiese en morfologiese) onreëlmatighede en onduidelikhede inherent aan natuurlike taal, fokus ek hier op die semantiese inhoud van die simbole self— as 'gegrond' in persepsies, aksies en situasies in die werklike wêreld.

In Hoofstuk I stel ek die hoof teoretiese probleme bekend wat ek beoog om aan te spreek. Ter agtergrond bespreek ek die mede-ontwikkeling van KI en verskeie berekeningsteorieë van kognisie in kognitiewe wetenskap, asook die *begrondingsprobleem* (of 'internalistiese strik') waarmee hulle te make het. Daarna beskryf ek die benadering wat ek gebruik om dit in die res van die tesis aan te spreek.

Om verskeie aspekte van die probleem uit te lig, bied ek 'n kritiese ontleding van die relevante teoretiese literatuur in filosofie, kognitiewe wetenskap en (kognitiewe) taalkunde in hoofstuk II. Hier evalueer ek die belangrikste filosofiese/sielkundige debatte rakende die aard van konsepte; teorieë oor hoe konsepte aangeleer, gebruik en gerepresenteer word in die mense se verstand; en bied op grond hiervan my eie weergawe aan van konseptuele struktuur, gegrond in huidige (kognitief aanneemlike) 'konneksionistiese' teorieë van kognisie. Om verder te verduidelik hoe sulke konsepte aangeleer en gekommunikeer word, stel ek ondersoek in na die relevante *beliggaamde* (bv. kognitiewe, perseptuele, sensoriese, affektiewe, ens.) faktore wat betrokke is by menslike (sosiale) kognisie, gebaseer op huidige wetenskaplike navorsing in die velde van 4E Kognisie en beliggaamde sosiale kognisie. Na aanleiding hiervan fokus ek spesifiek op teorieë oor die begronding van taal, aan die hand van die kognitiewe taalkunde navorsingsprogram wat daarop gemik is om 'n genaturaliseerde, kognitief-aanneemlike begrip van die menslike konsep- en taalverwerwing en -gebruik te ontwikkel. Ek sluit die hoofstuk af met 'n samevatting waarin ek my bevindings uit hierdie verskillende dissiplines krities integreer. Dit vorm die breë teoretiese basis waarvolgens ek die meer praktiese oorwegings vir die implementering daarvan in AI kan evalueer in die volgende hoofstuk.

In hoofstuk III bied ek 'n oorsig aan van die verskeie hoofbenaderings in die navorsingsgebied van Natuurlike Taal Begrip in KI, en evalueer hul onderskeie sterkpunte en tekortkominge na aanleiding van spesifieke modelle. Daarna bied ek 'n kritiese samevatting aan waarin ek die verskillende benaderings kontrasteer en kontekstualiseer in terme van die meer fundamentele teoretiese oortuigings wat hulle blyk te weerspieël.

Op grond hiervan evalueer ek, in die laaste hoofstuk, die voorgenoemde begrondingsprobleem en die maniere waarop dit in verskillende (teoretiese en praktiese) dissiplines geïnterpreteer word, en onderskei tussen 'n sterker en swakker lees. Daarna voer ek argumente aan waarom die implementasie van die sterker weergawe in KI (prakties en teoreties) problematies mag wees. In stede daarvan, oorweeg ek, na aanleiding van die teoretiese insigte wat ek versamel het, sommige van die sleutelvereistes om soveel as moontlik van die menslike vermoë van taalgebruik te 'begrond' (in die swakker sin) met robotiese KI-agente, sowel as die implementeringsbeperkings wat daarvoor benodig sal word. Laastens evalueer ek sommige van die hoof tegniese uitdagings wat daaraan verbonde mag wees, indien dit wel die doel is om aan al die gespesifiseerde vereistes te voldoen.

# Acknowledgements

This thesis was written under rather abnormal circumstances and time pressures during the 2020 COVID-19 Lockdown period, and there are many who deserve my deepest gratitude for their support and understanding in my state of frenzy.

Firstly, I would like to thank my parents and uncle for supporting me financially during my studies, none of this would have been possible without you. I would also like to specifically thank my mother for her patience, love, and care in letting me hermit at her house and keeping me fed.

I would also like to thank my friends, Geoff and Aadil, for offering their valuable time and feedback, as well as Nick, whose thoughtful critique and ongoing encouragement has been invaluable.

Finally, I would like to thank my supervisor, Prof. J.P. Smit, for his unending support and faith in me during my postgraduate studies; his tolerance of my odd methods and frequent changes of plan; and for his always insightful, and often brilliant, commentary and advice. I could not have asked for a better mentor.

# Table of contents

**Key abbreviations**

AI: Artificial Intelligence

GOFAI: Good Old-Fashioned AI

NLU: Natural Language Understanding

CCTM: Classic Computational Theory of Mind

LOT/LOTH: Language of Thought (Hypothesis)

4E Cognition: Embodied, Embedded, Extended, and/or Enacted Cognition

Ch.: Chapter

§: Section

# Chapter I: Introduction and background

In this chapter, I introduce the main (theoretical and practical) issues I aim to address in this thesis. I first offer an introduction of the title issue—what is meant with 'not cheating' on the Turing Test—followed by an overview of the structure of my thesis and a brief description of the relevance of the different chapters. Secondly, I offer some background regarding some of the key terms, debates and theories that I address, particularly regarding the development of the field of Artificial Intelligence (AI), cognitive science, and computational theories of mind. Within that context, I introduce the *grounding problem* that forms a key issue I grapple with in this thesis. Finally, I conclude the chapter with a brief summary of the issues introduced.

## 1. Thesis introduction

In 1950, Turing wrote his influential article, *Computing Machinery and Intelligence*, in which he proposed a measure for determining whether a computer can be deemed 'intelligent'[1]. To pass Turing's test, or 'Imitation Game', a computer would have to provide indistinguishably human-like responses to a series of interrogations. Initially, towards this end, most research in AI ambitiously tried to identify law-like principles behind human behaviour so as to model our causal reasoning, language and vision abilities formally, in what became known as Good Old-Fashioned AI (GOFAI), putting to test centuries of philosophical theorising on human reason and behaviour. However, in putting theory to practice, we soon realised that we know less about our own cognition than we thought, and time-consuming approaches that relied on hand-written rules and exceptions failed to deliver viable solutions for tasks that we found simple ourselves. Instead, since the so-called *statistical revolution*[2] of the nineties, researchers have increasingly settled for more practical approaches that infer statistical correlations from ample examples of how we behave, which have enabled systems to sufficiently *emulate* the desired behaviour in multiple narrow applications—even without 'understanding' the causal logic behind it (Dreyfus, 1992:203; Steels, 2007:21; Christianini, 2019:1). Thus, the aim of most AI applications largely shifted from trying to model everything about human behaviour at once, the so-called artificial general intelligence we find in science fiction literature (or *Strong AI*), to finding practical and efficient solutions for narrow-scope tasks like categorisation, pattern recognition, etc. (*Weak AI*), by essentially trying to copy what we do. Such kinds of shortcuts and trade-offs are precisely what have allowed statistical systems to achieve the narrower-scope successes we see in the field today.

However, as statistical methods for processing natural language (and particularly the area of Natural Language Understanding, NLU) have been increasing in sophistication and areas of application[3], there is a growing demand for AI technologies that take decisions and execute tasks on our behalf—including social robots that collaborate with us in our homes and workspaces (Hermann et al., 2017:1). As such, there is a pressing need for, and great commercial interest in, enhancing NLU technologies further so as to enable seamless communication between humans and machines. Recent successes in the area of (corpus-based) statistical language processing have, for some at least, renewed optimism in artificially emulating all the essential aspects of human-like linguistic ability so as to 'pass' the hypothetical *Turing Test*. Certainly, statistical methods can get us far, as much of (what is common in) our behaviour can be emulated purely by extracting surface-level patterns from ample data—the backbone of current machine learning methods—and resolving uncertainties using probability. Whilst some have argued that such approaches will ultimately be sufficient (e.g. Bryson, 2001), others have expressed doubts that human-like intelligence can be reduced to disembodied (rule-based or statistical) algorithms (e.g. Dreyfus, 1972, 1992; Harnad, 1990), and insist, for instance, that responding as a person would to more open-ended questions demand some (distinctively human) notions of common sense[4], and a deeper understanding of how we use language to engage with our internal and external world—things that statistical 'copycat' approaches notoriously lack[5].

---

[1] That is, defined operationally: not necessarily able to reproduce everything about human cognition internally, but only to exhibit external (linguistic) behaviour that seems sufficiently 'human-like'.

[2] The shift to statistical AI was also enabled by a proliferation of available data (through the internet), as well as greater storage and computational power.

[3] Notable applications include automated reasoning, machine translation, query-answering, news gathering, text-categorisation; as well as sophisticated implementations that range from digital assistants that process voice commands

Arguments in this direction have been further provoked by the recent development and popularisation of embodied theories of cognition, which emphasise the complex ways in which our species-specific bodies (and interactions with a physical/sociocultural environment) factor into our cognitive abilities. These, in turn, have sparked a new movement in AI on 'grounded' approaches that integrate visual/motor activities (particularly for aforementioned social robots), and thereby some new hope in achieving Strong(er) AI (e.g. Steels, 2008, 2009).

Drawing from recent insights in *4E* (Embodied, Embedded, Extended, Enacted) cognition, cognitive linguistics[6], and AI research, this thesis explores key desiderata to consider if we were to take seriously the aim of developing AI agents that are able to 'ground' natural language in knowledge/perception of real-world objects, events, and embodied experiences as we do—and evaluate the difficulties. My aim is thus an interdisciplinary investigation into the relevant requirements for appropriately (i.e. *honestly*) 'passing the test' for a true human-like command of language; that is, to *replicate* rather than *mimic* human language use.

Developing on previous work (Alberts, 2019) in which I explored the complexities and challenges involved in enabling AI systems to deal with the *grammatical* (i.e. syntactic and morphological) irregularities and ambiguities inherent in natural language, I turn my focus here towards appropriately inferring the content of symbols themselves, as 'grounded' in real-world phenomenal experiences. Taking seriously the notion that our use of language is fundamentally linked to the kinds of bodies (and body-based experiences) we have[7], my investigation involves an extensive exploration of those evolutionary (cognitive/perceptual/motor) capacities and biases that allow us to process and store complex (external and internal) perceptual phenomena in broadly similar ways, and the factors involved in communicating them through conventions of natural language. Beyond theory, my task also involves an exploration of the practical capabilities (and shortcomings) of current AI (NLU) systems, as well as some deeper philosophical reflection on future possibilities and challenges.

---

[4] (e.g. Siri or Alexa), to chatbots that you can have simple conversations with, to full-blown robots that can coordinate speech with facial expressions and hand gestures (e.g. Hristov et al., 2017).

[4] Regarding what he terms the 'commonsense-knowledge problem' Dreyfus (1992:xvii) identifies three related issues: (i) how to organise knowledge in order to make inferences from it, (ii) how to represent skills or 'know-how' as 'knowing that', and (iii) how relevant knowledge can be brought to bear in certain situations.

[5] Some illustrative examples are the often-nonsensical scripts that have been written by AI algorithms, for instance Austin McConnell's (2019) short science fiction film, *Today Is Spaceship Day*.

[6] That is, a growing research enterprise in linguistics that aims to construct a naturalist, cognitively plausible understanding of human concept and language acquisition (based on relevant findings in science).

This investigation proceeds in four chapters. In the rest of this chapter, I discuss the co-development of the fields of AI and cognitive science. This includes a discussion of the development of (different strands of) computational theories of mind, and the 'grounding problem' faced by them. I conclude the chapter with a brief summary of the key issues and a description of the approach I take to address the problem of grounding in the rest of the thesis.

To unpack and evaluate the relevant aspects, in chapter II, I carry out an extensive critical investigation into the relevant theoretical literature in philosophy of mind, cognitive science, and cognitive linguistics, in hopes of uncovering some basic (bodily) factors that play a role in human concept and language acquisition and use. Firstly, I evaluate some major philosophical/psychological debates regarding the nature of concepts; theories regarding how concepts are acquired, used, and represented in the mind; and, on that basis, offer my own account of conceptual structure, grounded in current (cognitively plausible) connectionist theories of thought (which I discuss later in this chapter). To further explicate how such concepts are acquired and communicated, I then evaluate the relevant embodied (e.g. cognitive, perceptive, sensorimotor, environmental, affective) factors involved in human (social) cognition, drawing from current scientific research in the areas of 4E cognition and embodied social cognition. On that general basis, I turn my focus specifically towards *grounded* theories of language, drawing from the cognitive linguistics research programme that aims to develop a naturalised, cognitively plausible understanding of human concept/language acquisition and use. I conclude the chapter with a summary wherein I integrate my findings from these various disciplines, presenting a general theoretical basis upon which to evaluate more practical considerations for the implementation of those processes in AI—the topic of the following chapter.

In chapter III, I offer an overview of the different major approaches in area of Natural Language Understanding (and their integrations) in AI, evaluating their respective strengths and shortcomings, in terms of specific models. I then offer a critical summary wherein I contrast and contextualise the different approaches in terms of the more fundamental theoretical convictions they seem to reflect.

---

[7] This is a key argument of recent theories in 4E Cognition and cognitive linguistics.

Based on all my findings, in the final chapter, I re-evaluate the aforementioned grounding problem and the different ways in which it has been interpreted in different (theoretical and practical) disciplines, distinguishing between a stronger and weaker reading. I then present arguments for why implementing the stronger version in AI seems, both practically and theoretically, problematic. Instead, drawing from the theoretical insights I gathered in chapter II, I consider some of the key requirements for 'grounding' (in the weaker sense) as much as possible of human language-use capability in robotic AI agents, as well as some implementational constraints that might need to be put in place to achieve this. Finally, I evaluate some of the key challenges that may be involved, if indeed the aim were to meet all the requirements specified.

## 2. Computationalism and cognitive science

A common theme throughout Western history has been to understand the human mind in terms of whichever technological invention is the most advanced at the time (Dreyfus, 1992): in ancient Greece, the mind was understood as a hydraulic clock; in the fourteenth to nineteenth century, as a clockwork mechanism; in the industrial revolution, as a steam engines; and since the 1930s, the predominant view has been of the human mind as a computer (in some form or other). As such, the successes and failures of modern computer science have been inspiring new insights regarding human cognition, developing in tandem with cognitive science—an interdisciplinary enterprise combining research from psychology, philosophy, neuroscience, AI, and linguistics, aimed at gaining a deeper understanding of the human mind. In this subsection I discuss two main theories of mind that feature prominently in cognitive science, the classical computational and connectionist accounts, both of which have been inspired by, and inspired, advances in computer science.

### 2.1. Classic Computational Theory of Mind

Since the invention of geometry and logic, the rationalist idea that all of human reasoning is reducible to some form of calculation has been prominent in Western philosophical thought (Dreyfus, 1992:67-69). From Socrates' demand for moral certainty and Plato's demand for explicit definitions, to Leibniz's binary system, Boole's logical operators, and Frege's concept notation, the belief in a total formalisation of knowledge has fascinated many influential figures in the tradition. Practice started to catch up to theory with Charles Babbage's (1837) theoretical 'Analytical Engine' and Turing's (1950) influential idea of a 'thinking machine', which inspired the development of the digital computer.

Turing's idea was a theoretical machine that could do any job done by human 'computers' (i.e. people that carry out computations) purely through reading/writing discrete symbols on an (infinitely long) tape, and subsequently changing its internal state, according to definite rules[8]. According to what became known as the Church-Turing thesis[9], for any well-defined cognitive task requiring the processing of discrete symbols (or even analogue computations, given that the process is describable in terms of a precise mathematical function) it is theoretically possible to construct a Turing Machine that can solve that problem (Rescorla, 2020). Moreover, they argued that one could theoretically construct a Universal Turing Machine that can be programmed to run any such machine, and thus compute *any computable function*. To many, this seemed a promising way to model what occurs in the minds of humans when they act upon their environment, and it was believed that the only requirement for simulating human intelligent activity would be suitable rules and a long enough tape. The idea coincided especially well with central work in analytical philosophy which construed human behaviour as the result of (symbolic) propositional attitudes[10], and thought occurring in terms of formal transitions between such propositions—which, on Turing's argument, a machine could implement (Ward et al., 2017:367).

---

[8] For example, 'If you are in state B and read 1, stay in B, write 0 and move one square to the left' or 'If in state A and read 0, change to state B, write 1 and stop'.

The theoretical Turing Machine gave rise to two ideas that became central to cognitive science: *physical computation*; that is, that systems similar to the Turing Machine can be physically instantiated (in practice), and *information processing*; that is, that such systems can adequately process input (and give convincing output) based on structural properties alone[11] (Gładziejewski & Miłkowksi, 2017). This relied on two basic assumptions: the functionalist idea that mental states can be defined purely in terms of (nested) internal processes that transform inputs into output, irrespective of their physical substrates; and the rationalist assumption that any intelligent activity boils down to a set of clearly-defined instructions. Since computational states are multiply realisable[12] and mental states are understood as computational states, mental states are considered multiply realisable (a central idea in the popular philosophical view of *functionalism*[13]). This multiple realisability gives rise to different levels of explanation that are typically considered, to some extent, independent of each other: the psychological, the computational/algorithmic, and the implementational/neurological (Pecher & Zwaan, 2005:1). That is, given a particular mental state (e.g. a belief that *X*), there may be multiple algorithms that can process the relevant function, and for each of those algorithms there may be multiple possible physical substrates that can realise it. As a result, many computational theories of cognition aim to explain the mind without regards to neuroscience[14], or any particular form of embodiment[15].

---

[9] This refers both to Turing and mathematician Alonzo Church.

[10] That is, mental states such as beliefs, desires, hopes, fears, etc.

[11] That is, relying on structural (surface-level) features like syntax, pixel patterns, etc., rather than the more abstract (semantic) content it may represent.

[12] That is to say, the same program can run on different computational systems (e.g. Mac or Windows) and the same system can be realised by different physical substrates (e.g. electrical circuits, mechanics, materials, etc.).

[13] See Levin (2018) for an overview of this approach.

[14] Although, even before Turing, neuroscientists McCulloch and Pitts (1943) argued that the brain resembles a sort of digital computing machine, based on their (simplified and idealised) explanations of the all-or-none signals of neurons in the brain (Piccinini, 2009:517).

[15] As Fodor (1997) puts it, the 'special sciences' of the mind are autonomous.

This view that the mind/brain is literally like (the software of) a digital computer, is called the Classical Computational Theory of Mind (CCTM). This forms part of the general class of thought known as *computationalism*: the view that intelligent behaviour is causally explicable by computations performed by the agent's mind/brain, and has, in some form or another, been the mainstream view of cognition in philosophy, psychology, and neuroscience for decades (Piccinini, 2009:515). There have been many (often competing) computationalist approaches[16], but common among them is a commitment to describing cognition as computation over sensory inputs, internal states and/or representations based purely on their structural (or relational) properties (Pecher & Zwaan, 2005:1; Piccinini, 2009:519).

The dominant computational approach in the philosophy of mind is the cognitivist (symbol-processing) approach. In response to former behaviourist approaches that merely focused on external behaviour, cognitivism seeks to explain cognition in terms of processes involving internal states and representations (Piccinini, 2009:519). Representations, here, are understood as abstract symbols that stand in for the objects they represent, combined as logical expressions that respect logical rules of inference, which we use to reason about those objects. For instance, 'If *A* then *B*' '*A*, therefore *B*'; or 'Desire *A*', 'Believe *B* will cause *A*', therefore 'Do *B*', where *A* and *B* can refer to certain propositions, entities, actions etc. Rather than building up their content from sensory experiences, the meaning of a concept/symbol consists solely of its links to other concepts in the system (Pecher & Zwaan, 2005:1). Harnad summarises eight basic characteristics of a cognitivist symbol system:

> [It consists of] (1) a set of arbitrary physical tokens…that are (2) manipulated on the basis of explicit rules that are (3) likewise physical tokens and strings of tokens. The rule-governed symbol-token manipulation is based (4) purely on the shape of the symbol tokens…and consists of (5) rulefully combining and recombining symbol tokens. There are (6) primitive atomic symbol tokens and (7) composite symbol-token strings. The entire system and all its parts…are all (8) semantically interpretable: The syntax can be systematically assigned a meaning (Harnad, 1990:336).

According to Harnad (1990:336), a combination of all eight of these properties are critical for a meaningful definition of a symbolic system.

A popular version of this view is the Language of Thought Hypothesis (LOT or LOTH), which holds thought to consist of an internal system of symbolic, word-like mental representations stored in memory and manipulated according to law-like mechanical rules (Rescorla, 2020). These LOT (or *Mentalese*) expressions have a language-like syntax and compositional semantics: complex representations are built from basic symbols, and the meaning of a complex representation follows logically from the meanings (and particular structural arrangement) of its symbolic constituents. In particular, there is a presumed isomorphism between a person's mental states and relevant sentences in a LOT (Churchland. 1980:149). The view emerged gradually from a variety of thinkers during the middle ages[17] but largely fell out of favour in the 16th and 17th century (Rescorla, 2019). However, following the emergence of computer science, it was dramatically revived by Fodor (1975) in *The Language of Thought*, where he presumes thought to consist in a system of (innate) primitive representations/concepts, which, when combined (according to innate systematic rules), form complex representations.

---

[16] See Piccinini (2009) for a review.

Assuming a strictly computational understanding of the mind, a central aim of Fodor's work is to defend central folk-psychological intuitions regarding aforementioned propositional attitudes. These can be reduced to two types of states: belief-like states, representing the world, and desire-like states, representing one's goals—both of which we commonly employ to explain human behaviour (Rescorla, 2019). For instance, to explain why Alice opened a biscuit tin, we might note that Alice believed she could find a biscuit inside, and that she had the desire to eat a biscuit. These attitudes have intentionality, in that they are *about* a particular subject matter. Fodor's LOTH deals with this notion of intentionality by postulating symbolic mental representations that *stand in* for objects and their relations, which serve as the contents of propositional attitudes: "For each episode of believing that P, there is a corresponding episode of having, 'in one's belief box', a mental representation which means that P" (Fodor, 1998:8). Fodor (1998:8) explains that 'belief box', here, is understood functionally: having a belief is to have a particular (belief-like) attitude to a given symbolic expression that represents an event in the world (and likewise for other kinds of propositional attitudes). The transition between mental states, as sentential attitudes, is then explicable in terms of the logical relations between those LOT sentences, which, most straightforwardly, consist of (abductive and deductive) inference (Churchland. 1980:149).

Other motivations for Fodor's compositional symbolic approach to CCTM include accounting for what he calls the productivity and systematicity of thought: that we seem able to entertain an unbounded number of thoughts using a finite set of basic constituents (concepts), and that this seems to follow systematic rules of production (Fodor, 1975, 1998). For example, if one understands the concepts ALICE, HATE and BOB[18], one is able to think and comprehend 'Alice hates Bob', just as easily as 'Bob hates Alice' using the same systematic rules of combination. Likewise, if one thinks 'Bob's father is bald', one can also think 'Bob's father's father's father is bald', etc. To Fodor, this systematicity and productivity seem to support—in fact, necessitate—the view that thought consists of systematic combinations of atomic concepts. Moreover, as people can communicate using (what seem to him) the *same* concepts, Fodor maintains that concepts cannot be 'built up' through subjective experiences. Nor can the meanings of concepts depend on those of others, as the possession of slightly different concepts would reverberate through the entire system, he argues (Fodor, 1998:114). Instead, he insists that primitive concepts are innate and atomistic: mental representations get their content causally by combining innate primitive concepts to form complex concepts—representing classes of entities in the world—which are then combined into syntactic structures that isomorphically stand in for the relations between entities in the world (Fodor, 1975; Fodor & Pylyshyn, 1988).

---

[17] Some of these include Augustine, Boethius, Thomas Aquinas, William can Ockham, and John Duns Scotus.
[18] I follow the convention of writing concepts in SMALL-CAPS, and words in *italics* or 'quotes'.

From an implementational perspective, Hurley (2001) explains that CCTM follows what she calls the 'classical sandwich' model, which views the mind as divided into three modules: perception, action, and cognition in the middle. On this view, cognition is a necessary intermediate process between perception and action, which are seen as peripheral and separate both from each other and the higher processes of cognition. Most of the hard work occurs in the middle: conceptual/symbolic outputs are produced from decision-making cognitive systems acting on symbols (representing perceptual information), and those outputs comprise new mental states, which are sometimes converted into motor commands that cause action (Hurley, 2001).

Fodor's theory, as a specific version of CCTM, focuses mainly on the psychological level of thought and the nature of presumed propositional attitudes (i.e. in the cognitive module). Moreover, Fodor (1983) advocates a modular view of the mind wherein each functional domain, like language, occupies its own separate module, and each perceptual system is strictly separated as well. Furthermore, language itself is divided into distinct modules with their own rules, such as a syntactic component with rules that govern how lexical units may be combined, and other components for dealing with sound and sentence meaning (Evans, 2019:132). On this view, body-based perceptions are wholly distinct from the *amodal* (i.e. sensory-neutral) representations in the conceptual system, and are stored in different areas of the brain (Evans, 2019:208). Perceptual and motor processes reach the brain via "informationally encapsulated 'plug-ins'" that provide limited forms of input and output (Wilson, 2002:625). Assuming such hard divisions may be necessary for a computational approach that relies on the systematic manipulation of *discrete* symbols[19], and assumes that similar computations over discrete symbols should be multiply realisable amongst different individuals—and possibly computers[20].

Until the early eighties, it was commonly assumed that computationalism is committed to the existence of a LOT and that cognition has little or nothing to do with the neural architectures that support it. During the eighties, however, connectionism emerged in psychology as a viable—and biologically plausible—contender to the classical picture (Piccinini, 2009).

## 2.2. Connectionism

Whilst classical computationalists explain cognition in terms of logical/syntactic structures, connectionists use *neural networks* that consist of interconnected, simple processing units (or *nodes*), loosely modelled on the network of neurons and synapses in the brain. The links between nodes have weights that model the strength of their connections, and many, if not all, nodes are processed in parallel to compute a single output. Buckner and Garson (2019) explain that, if the human nervous system were modelled as a neural network, the input nodes would correspond to the sensory neurons, the output nodes to the motor neurons, and the intermediate (hidden) layers of nodes would represent all remaining neurons.

---

[19] That is, in order for the symbols representing sensory input to be discrete, the input should perhaps also be discretised.
[20] If CCTM is correct, and the 'software' of the mind is multiply realisable, then everything about our minds can, theoretically, be simulated on a computer: if one could also simulate each module of the classical sandwich, one could build an entirely virtual mind and environment. Many current theorists are working toward this goal, including director of engineering at Google, Ray Kurzweil, who conjectured that it will be possible to 'upload' our entire brains to computers within the following 32 (now 25) years (Woollaston, 2013).

This model was encouraged by the development of artificial neural networks and has been gaining traction due their increasingly successful applications in AI, particularly in tasks involving categorisation and pattern recognition. In a basic (feedforward) neural network, the activation pattern set up by a network is determined by the (positive or negative) weights between nodes, that either strengthen or inhibit the activity received from another node. The activation value of each node is calculated using a simple activation function that is (typically) adjusted to fit between 0 and 1 (depending on which function is used). When this activation exceeds a certain threshold, its value gets calculated into the network. Since it is assumed that all the units calculate essentially the same kind of simple activation function, the system depends primarily on the relative weights between the nodes (Buckner & Garson, 2019). In most current artificial neural networks, these weights are initiated with random values, which the system then automatically adjusts based on its training; that is, for instance, examples of input-output pairs (as in *supervised learning* systems), or being rewarded for certain actions (as in *reinforcement learning* systems) (Russell & Norvig, 2010:695, 830). Through training, the weights in the network are adjusted slightly so as to bring the network's output values closer towards the desired output (or expected rewards). Thereby, through multiple iterations of training over many different examples, or through gaining all the more feedback, the system can eventually find a (seemingly) suitable algorithm that is able to generalise to similar tasks; that is, to pick up which (structurally similar) types of combinations of input values typically correspond to a certain output (as in classification problems: e.g. 'Is this a cat or not?'), or to predict, based on inferred correlations in data, how new data would correlate (as in regression problems: e.g. 'How much would a house cost with these features?'), or to learn which actions are preferable based on user feedback (e.g. 'Should I show this kind of advert?') (see Russell & Norvig, 2010, Ch.18).

Because models can sometimes 'overfit' to the particular set of training data they were given, i.e. find algorithms that are modelled so specifically to familiar examples that they fail to generalise in the right way, models should be trained on data that is sufficiently varied, so that they can easier make the right kind of abstractions for effectively dealing with novel input. To find the optimal algorithm, supervised networks use a method called *backpropagation*, in which the network compares its current output to the desired output (given by a human), and then 'propagates' backwards through the net to adjust each weight in the direction that would bring it closer to, collectively, reaching the right output (see Russell & Norvig, 2010, Ch.18).

Given their successes in emulating—at least in effect—many human cognitive tasks, philosophers have been gaining interest in neural networks as a possible framework for characterising the nature of the (human/mammalian) mind and its relation to the brain (e.g. Rumelhart & McClelland, 1986; Bucker, 2019). Connectionist models have several properties that make the view seem promising. Firstly, neural networks exhibit effective flexibility when confronted with the messiness and complexity of the real world: noisy input or damaged/faulty individual units causes a graceful degradation of accuracy (whereas, in classical computers, any noise or faulty circuitry easily results in catastrophic failure). Secondly, neural networks are particularly apt for dealing with problems that require the resolutions of many (conflicting) constraints in parallel: plenty of evidence from AI suggests that many cognitive processes like pattern (e.g. object) recognition, prediction, planning and coordinated motor movement involve such problems (see Buckner & Garson, 2019). Although classical systems can be used to satisfy multiple constraints, connectionists contend that neural network models offer far more natural/biologically plausible methods for tackling such problems (Buckner & Garson, 2019). As we shall see in the following chapters, the flexibility of neural networks also deals much more naturally with family-resemblance theories of concept formation/use, which have largely come to replace those that relied on strict, formal principles (such as necessary and sufficient conditions).

Most connectionists reject a language of thought or a classic computational/representational theory of mind altogether, although this remains a matter of controversy. On the face of it, the connectionist picture of cognition—as a dynamic and graded evolution of activity in an interdependent, distributed network—seems to directly contradict the classic view of manipulating discrete symbols (representations) according to rigid rules, in a linear fashion. Likewise, many classicalists reject arguments that take connectionism as biological evidence for an associationist picture of cognition (i.e. as distributed networks of associations), as it directly contradicts the folk-psychological intuitions the classical picture supports, and also does not give (as clear) a picture of the systematicity and productivity of thought (Fodor & Pylyshyn, 1988). Yet, some connectionists do not consider their kind of model a challenge to CCTM and even explicitly support a classical view, and naturally vice versa (as the human brain does, in fact, consist of a network of neurons and synapses). So-called implementational connectionists seek to accommodate both paradigms: whilst they concede that the brain is, physically, a neural network, they maintain that it implements a symbolic processor at a higher level of abstraction (Piccinini, 2009:521; Buckner & Garson, 2019). Their role is, then, to determine how exactly the mechanisms required for symbolic processing can be realised by such a network[21].

On the other hand, many connectionists reject the symbolic processing view as a fundamentally flawed guess about how cognition works—owing, perhaps, to ingrained *a priori* folk-theoretical/philosophical assumptions about language, propositional attitudes and the like. One such argument comes from Patricia Churchland (1980), who maintains that most of the information-bearing states of the central nervous system do not have the nature of a sentential (propositional) attitude:

---

[21] Although, as noted, many do not bother, as they consider psychology a special science in its own right.

> [T]hat is, they are not describable in terms of the person's being in a certain functional state whose structure and elements are isomorphic to the structure and elements of sentences. Obviously, for example, the results of information processing in the retina cannot be described as the person's believing that p, or thinking that p, or thinking that he sees an x, or anything of that sort (Churchland, 1980:147).

For instance, if a magician puts a ball underneath one of three cups and shuffles them, my *belief that the ball is under cup X,* more likely presents itself as an expectation of perceiving a ball when cup X is lifted. Although I may be able to convey this in language as a sentential belief, there is no independent empirical reason to suppose that propositional attitudes exist in the mind as we are accustomed to talking about them (not to mention that Fodor's insistence that other theories fail to adequately account for such propositional attitudes or their compositionality begs the question—see Dreyfus, 1992, Part II[22]). As such, Churchland (1980:147) argues that we best abandon the sentential approach for 'all but rather superficial processing', and that we should look towards developing more cognitively realistic theories of the mind, guided by empirical science rather than common-sense intuition. This includes satisfying certain empirical constraints, such as the fact that theories of cognition should be realisable by the neural structures of the human brain; that they should fit into an evolutionary account of how we developed from non-verbal organisms; and should likewise account for the intelligent behaviour of other non-verbal organisms (Churchland, 1980:148). Hence, there should not be radically distinct theories of intelligent behaviour for verbal and non-verbal organisms, as LOT theories typically suppose. Rather, she argues that "we should expect a theory of information processing in humans to be a special case of the theory of information processing in organisms generally" (Churchland. 1980:149). Connectionism, as a more biologically plausible, domain-general account of mental processing, avoids these criticisms.

Another argument is that the classical approach does a poor job of explaining those features like the flexibility, context-sensitivity, and generalisability[23] of connectionist models, which human intelligence seems to exhibit (Buckner & Garson, 2019). Moreover, recent findings in neuroscience fundamentally challenge the modular view of the mind, suggesting that cognitive processing is not encapsulated in discrete modules, but is distributed over the whole brain (more akin to connectionist frameworks)—see Fox and Friston (2012) for a discussion. Findings regarding *neural plasticity[24]* also point to the importance of experience for shaping dynamical cognitive structures (see Cech & Martin, 2012). From a linguistic perspective, modular views of language have also been challenged on independent grounds by findings in cognitive linguistics (discussed in §3 of ch.II).

---

[22] In short, Dreyfus argues that Fodor's theory is founded on a provably wrong assumption that the mind can be modelled as a computer (in a classical sense) rather than a scientific theory about human cognition based on empirical evidence. Moreover, it assumes, without grounding, that laws such as those in science between atoms holds on the abstract level of human knowledge and cognition.

[23] That is, the ability to use the same statistical approach to deal with vast amounts of (different types of) data, without the need for crafting domain-specific rules.

[24] That is, the ability of the central nervous system to adapt in response to lesions or changes in the environment.

Piccinini (2009:521) notes a further distinction between connectionists that consider themselves computationalists, and those maintain that the kind of neural network processing done by our minds is something other than computation; that is, at least in the narrow sense, as the manipulation of digits. On the other hand, if one takes too broad an understanding of computation simply as 'information processing', it becomes trivial, as pretty much all physical processes can be seen as processing information in some sense[25] (see Sprevak, 2018). Therefore, those that consider themselves computationalists need a definition of computation that is broad enough that it applies to actual minds, but narrow enough that it is not trivial to explain the mind/brain as a computer.

Beyond the debate between classic computationalists and connectionists, many criticisms have been raised against the respective views. Many of those raised against CCTM are considered fundamental, and are discussed in an increasing number of contributions (Harnad, 1990; Dreyfus, 1992; Valera et al., 1991; Glenberg, 1997; Barsalou, 1999; Pulvermüller, 1999; Hurley, 2001; Pecher & Zwaan, 2005). Two popular criticisms are Barsalou's (1999) *transduction problem* and Harnad's (1990) *symbol grounding problem* (Harnad, 1990) that have inspired much debate. The transduction problem is the issue of how perceptual experiences can be translated into the amodal, arbitrary symbols (representing concepts) in the mind. In digital computers (and early GOFAI systems) this was achieved by means of 'divine intervention' by a programmer. For instance, for AI programs that dealt with symbolic representation, programmers had to abstract concrete objects, actions, and events as discrete concepts like PERSON, CHAIR, and SIT, and then manually put them into appropriate combinations, such as [CAN[PICK-UP, PERSON, CHAIR]], [CAN[SIT-ON, PERSON, CHAIR]], etc. (example by Brooks, 1987). However, many contended such propositions are extremely limited and do not even approximate an exhaustive description of real-world objects, and a method that relies on so much external intervention does not explain human intelligence in a theoretically plausible way (e.g. Brooks, 1987; Pfeifer & Scheier, 1999; Pecher & Zwaan, 2005:2).

Secondly, CCTM (and sometimes connectionism) has been charged with the *symbol grounding problem*; i.e. the problem of "how to causally connect an artificial agent with its environment such that the agent's behaviour, as well as the mechanisms, representations, etc. underlying it, can be intrinsic and meaningful to itself, rather than dependent on an external designer or observer" (Ziemke, 1999:87). In in its original formulation, Harnad (1990) expands on Searle's (1980) Chinese Room Argument[26], by arguing that symbol meanings cannot all be based on combinations of other symbols, otherwise the mind faces a task akin to trying to learn Chinese (as a first language) using only a Chinese dictionary. That is, with no relation to anything outside of the system. Instead, Harnad maintains that some symbols must get their meaning in virtue of their relation to the world outside the system so as to be *intrinsically* meaningful to the symbol system itself, rather than parasitic on the meanings in our heads (just as symbols in a book are not intrinsically meaningful to the book). Several authors have acknowledged that the grounding problem does not just apply to symbolic representations, but likewise to other forms of representations (e.g. Chalmers, 1992; Dorffner & Prem, 1993), and can be referred to more generally as the *internalist trap* (Sharkey & Jackson, 1994). A number of approaches to grounding have been proposed (which I review in the final section), all of which essentially agree on two points: that escaping this internalist trap is "crucial to the development of truly intelligent behaviour" (Law & Miikkulainen, 1994); and that grounding requires agents to be causally coupled with the external world in some way (without the mediation of an external observer)—although the nature of this coupling is disputed (Ziemke, 1999:88).

---

[25] Searle made the related argument that, even if what happens in physics can in some sense be considered computation, it does not necessarily equate the level of abstraction of symbol-manipulation: "syntax is not intrinsic to physics" (Searle, 1992:210).

[26] Searle (1980) aims to contradict the idea, inspired by Turing, that intelligent behaviour (in the Strong-AI sense) can be the outcome of purely computational (formal and implementation-dependent) processes in physical symbols systems. He suggested a thought experiment wherein a person in a room with a multilingual dictionary is able to translate sentences into Chinese purely by processing the characters using formal rules (given in their native language). Although the resulting output may make it seem like the person *understands* Chinese, in the appropriate sense, Searle argues that he certainly does not.

Connectionist architectures, like those that rely on learning from correctly labelled examples (i.e. *supervised learning*) have been charged on similar grounds, as they merely try to optimise their algorithm to achieve the 'correct' output, as specified by an external observer; that is, without it being intrinsically defined by the system. The extent to which artificial neural networks really resemble processes in the human mind has also been questioned, as intricate mathematical tools like backpropagation do not seem cognitively realistic (although some have argued the contrary, e.g. Whittington & Bogacz, 2019). Moreover, whilst neural networks require vast amounts (thousands or more) of examples before learning an adequate algorithm, many have argued that this is an inadequate model of human learning (e.g. Marcus, 2018). According to Harnad (1990:337), little is known about our brain's structure and the complex interplay between its 'higher' and 'lower' functions. As such, it is not clear that neural networks are, in fact, an accurate model of the human brain. Hence, he argues that judging a cognitive theory based on its ability to account for *brain-like* behaviour is "premature", as not only is it still far from clear what 'brain-like' means, but neither classical nor connectionist models are yet able to account for "a lifesize chunk" of human behaviour (Harnad, 1990:337).

Regarding deep learning[27] in particular, which is behind some of our most sophisticated neural network models, Marcus (2018) lists ten significant challenges still faced by it: that it is (i) data-hungry, (ii) has superficial solutions[28] with a limited ability for transfer to other applications, (iii) has no natural way to deal with hierarchal structure[29], (iv) struggles with open-ended inference, (v) is not sufficiently transparent[30], (vi) is not well-integrated with prior (real-world) knowledge, (vii) cannot inherently distinguish between correlation and causation, (viii) presumes a largely stable world, (ix) cannot be fully trusted[31], and (x) is difficult to engineer with[32]. He considers many of these extensions of the fundamental problem of contemporary (particularly supervised) deep learning systems: that they do well on challenges closely resembling their training data but less well on more open-ended cases or those on the periphery which often occur in the real world (as summarised by Alberts, 2019:105). Finding a way to make systems make the *right kinds* of inferences—to 'understand' what their goal is and why—is a fundamental challenge for approaches relying mainly on statistical correlations.

---

[27] Deep learning, as a subset of *machine learning,* refers to a type of neural network model that is 'deep' in the sense that it has more than two layers of nodes between the input and output layers, which allow for more complex relations between features to be extracted (and, as such, has greater ability to learn from unlabelled data).
[28] Recent experiments have shown that the performance of various deep networks trained on a question-answering task dropped precipitously with the mere insertion of distraction sentences (Marcus 2018:8–9).
[29] That is, syntactic relations between main and embedded clauses in a sentence (Marcus 2018:9)
[30] Rather than using parameters that we can clearly interpret and control, the features extracted by hidden layers are opaque and less straight-forward, which can lead to strange biases in algorithms (Marcus 2018:10–11).

Some have suggested addressing the above problems through integrating explicit symbolic programming with sub-symbolic (that is, using connectionist, deep learning) systems (Marcus, 2018); in fact, Young et al. predict that it "will be key for stepping forward in the path from NLP to natural language understanding" (2018:73). However, this in itself does not address the grounding problem, but merely combines sub-symbolic flexibility with the robustness of externally specified rules. Instead, research in cognitive science has increasingly been turning towards *embodied* theories of cognition to understand how, and in what, our thoughts are grounded.

In opposition to traditional cognitive psychological accounts, theories in embodied cognition reject the view that cognition is implementation-neutral[33]; rather, bodily processes (and their particular integration with an external environment) are taken to be a significant part (either causally or constitutively) of mental operations. Simply put, our particular sensorimotor and perceptual mechanisms have evolved to perceive, interact with, and make sense of, (ourselves and) our environment in a specific way, and should thus be accounted for by any theory that tries to explain how real/imagined entities are made meaningful to us, including how we communicate them to each other. This broad research programme supports the *principle of economy* in the explanation of cognitive processes; that is, to substitute, as far as possible, any postulation of amodal representations with plausible hypotheses about the nature of biological processes in the body (Fenici, 2012:276)—see §2-3 of ch.II. This is also considered consistent with theories of evolutionary continuity: "evolution capitalized on existing brain mechanisms to implement conceptual systems rather than creating new ones" (Yeh & Barsalou, 2006:374).

---

[31] Given how deep learning systems base their inferences on features they pick up on in training data, rather than explicit definitions, they are easily fooled (e.g. mistaking black and yellow stripes for school buses) (Marcus 2018:13–14).

[32] Although machine learning is effective in limited circumstances, it will not necessarily work in others as it yet lacks "the incrementality, transparency and debuggability of classical programming" (Marcus 2018:14).

[33] Further objections to the multiple-realisability of cognitive processes have also been made on independent grounds. For instance, Shapiro (2004) argues that this view fails to take into account the importance of temporal dynamics: if neurons used light signals rather than electricity, then signals would travel much faster around in the brain. If the physical substrates of the brain's 'hardware' alters the nature of cognition (without altering the computations) then cognition cannot just be computation.

The embodied cognition movement challenges CCTM in a number of ways. Firstly, through embodied accounts, cognition is explainable without the need for positing representations, and hence is typically not considered computational in the classical sense (which presumes the existence of representations). Rather, to understand how we use and comprehend language, embodied theories typically look at how language prompts activations in certain functional (e.g. sensorimotor, emotional) regions in the brain. As such, concepts are not considered amodal, but based on an individual's embodied (interoceptive and exteroceptive) experiences. Neither are concepts seen as innate, but, rather, as built up in memory through an embodied agent's interaction with its particular physical/sociocultural environment—which does involve the use of certain innate cognitive/perceptual mechanisms. Accordingly, cognitive processes are not considered implementation-neutral, as in the classical sandwich model (as modelling an organism's particular cognitive function would require recreating its particular embodiment, and not just mental 'software').

Proponents of the embodied cognition view typically disregard all computational approaches as they consider embodied cognitive processes too complex to model computationally, and the analogy between a body and a computer may not be as useful as the analogy between a brain and a computer (Rescorla, 2020). However, although incompatible with a CCTM approach, an embodied view of cognition is arguably still compatible with a connectionist approach, as long as the input signals are appropriately 'embodied', and connections are structured in an appropriate way—in which case, there might still be hope in, theoretically, modelling aspects of human cognition artificially, although total modelling might be more of a theoretical than practical possibility. The processes required, and the extent to which it may be practically feasible, is what I aim to explore here—particularly for the sake of grounding language comprehension in a sufficiently human-like way. In chapter II, I explore the former; that is, according to our best current psychological/cognitive scientifical theories, the key processes involved in human concept/language acquisition and use.

## 3. Chapter summary

In this chapter, I offered an overview of the development of the field of AI and how it has been inspired by, and inspired, theories about human cognition. I first discussed the theoretical background to computational theories of mind in the Western philosophical tradition, many of which were put to test in the practical implementation of GOFAI systems. I distinguished between the CCTM and connectionist theories of mind, their points of conflict and agreement, and their relative strengths and limitations.

From a scientific/naturalist perspective, I supplied some key reasons for preferring a connectionist to a CCTM approach to a general theory of human thought. This included, firstly, the fact that connectionism is more biologically plausible: not just because its structure seems to resemble the structure of the mammalian brain (as well as complementing findings in neural plasticity and distributed neural processing), but also because it fits in better with a general theory of cognitive evolution (i.e. fitting into an evolutionary account of how we developed from non-verbal organisms), rather than presuming radically distinct theories of intelligent behaviour for verbal organisms reliant on sentential mental structures. A part of its cognitive plausibility has to do with the fact that connectionist (neural network) models offer more natural explanations for tackling problems, in various domains, that require the resolutions of many (conflicting) constraints in parallel, including pattern recognition, prediction, planning and coordinated motor movement—as evidenced in AI research (whilst linear, rule-based CCTM approaches to modelling human intelligent behaviour have had much more limited successes). Moreover, a desirable property of neural networks is that they offer a much more flexible, context-sensitive approach to deal with the messiness and complexity of the real world, including issues of 'fuzzy' category (concept) boundaries—discussed in §1, ch.II. Although a CCTM model could, theoretically, be implemented on a connectionist model, I follow Churchland's (1980) argument that we best abandon the sentential approach for 'all but rather superficial processing', and that we should look towards developing a more cognitively realistic theory that is able to naturally account for as much as possible of human cognition (with as few as possible *ad hoc* assumptions).

In terms of shortcomings, some of the major ones I listed against connectionist models is that they, unlike human infants, typically require a lot of data (or training) before being able to infer an appropriate algorithm for dealing with input (executing tasks) effectively. Moreover, they lack human-like commonsense, as they are not usually well integrated with prior (real-world) knowledge and have trouble distinguishing between correlation and causation. This fundamental issue of finding ways to make systems make the right kinds of inferences—to 'understand' what their goal is and why—is arguably tied into the general *grounding problem* faced by (disembodied) computational approaches: the fact that these systems merely try to optimise their algorithm to achieve the 'correct' output over symbols (or input data, like pixels) that are meaningful to an external observer rather than the system itself. In Chapter IV, after my discussion of the relevant theoretical and practical approaches to solving the 'internalist trap', I revisit the problem and evaluate its implications more thoroughly. What most of these approaches agree on, however, is that a key requirement is some form of integration of internal cognitive processes with an external environment, 'grounding' symbols in notions of real-world objects, regions, actions, etc.

In the growing embodied (or more generally, 4E) cognition movement in cognitive science, there is a consensus that an appropriate model of our (grounded) cognitive processes requires an understanding of the particular sensorimotor and perceptual mechanisms we have evolved to perceive, interact with, and make sense of, (ourselves and) our particular (physical/social) environment. This is expanded on by the cognitive linguistics enterprise, which applies such general insights to a study of concept and language acquisition and use. Findings from both of these research areas are discussed in the following chapter. As my purpose is to explore a computationally tractable approach to implementing a (general and cognitively plausible) model of human language use in AI, I also attempt, as far as possible, to formulate a broadly connectionist interpretation of all the relevant theoretical factors that I identify—this proceeds at the end of chapter II.

## Chapter II: Towards a grounded understanding of concepts

This chapter explores the key theoretical foundations for understanding how humans acquire, and employ, concepts, and communicate them in language. This includes an investigation of human conceptual structure: how it is shaped by our embodied experiences, and how it relates to semantic structure in natural language. I first explore major theoretical approaches to understanding conceptual structure, followed by an investigation of embodied (and embedded, extended, enacted) theories of (social) cognition. On that basis, I discuss key arguments in the growing research programme of cognitive linguistics, that seeks to understand how our use of language is shaped by general (embodied) cognitive processes. Finally, I conclude the chapter by summarising and evaluating the key findings, in order to sketch an integrated picture of the basic requirements for recreating grounded, human-like natural language understanding in AI, the topic of chapter III.

## 1. Theories of concepts

Concepts are understood as a fundamental part of our mental lives. We use them to make sense of our experiences and learn from past encounters to guide our interactions with the world, and they are what we (attempt to) communicate through language. They constitute much of our knowledge of the world, including that of objects and their properties—as well as more abstract social and psychological categories like linguistic units, emotions, and situations—and are key to such cognitive capacities as inference, learning, categorisation, memory, and decision-making (Murphy, 2004:1-2; Prinz, 2004:1). In light of this, it should be unsurprising that questions regarding concepts (including words, their meaning and their relation to objects in the world) have been major topics of interest (and debate) in philosophy.

The last thirty years have seen the emergence of multiple competing theories of conceptual structure, many of which remain very much alive (Cain, 2016:96). Before evaluating some of the major ones in turn, it is worth clarifying ways in which the word 'concept' is used, and how I will use it here.

### 1.1. Background

#### 1.1.1. What are concepts?

Whilst there is wide agreement about the significance of concepts to thought, defining what precisely the word refers to is no simple feat. This is partly due to the fact that concepts may take various forms and contents, and the use of the word varies across different people, tasks and disciplines. Reconciling those differences, or at least describing how different types of concepts are coordinated, is a major goal of theorists in this area that is yet to be fully accomplished (Murphy, 2004:3; Cain, 2016:94).

Although research in the area concentrates mainly on concepts of common objects/concrete nouns, the principles that govern concept formation and employment are thought to be generalisable across different domains, and understandings of concepts usually correspond to more fundamental commitments to particular approaches to the study of mind/language (Margolis & Laurence, 2019). In this subsection, I briefly contrast three main understandings of what the word 'concept' refers to. That is, that concepts are mental entities, abstract objects, and abilities respectively. From there, I specify the understanding I will use for my purposes.

The first view, the idea that concepts are mental representations (of some sort), was widely endorsed by philosophers of the seventeenth and eighteenth centuries, and is considered the default position in cognitive science (Cain, 2016:93). This view emerged from introspective approaches to the study of mind that initially understood concepts as 'mental images' that somehow resemble things we encounter in the world. On this account, employing a concept involves having an image of the kind of object the concept represents before one's 'mind's eye' (Cain, 2016:97). Thereby, it accounts for the phenomenology of thought: if someone is asked to think of the concept COW (and they are familiar with cows), they may recall an instance from their long-term memory and experience a mental image akin to seeing a cow. It also offers an account of concept acquisition, tracing mental images back to previous experiences of objects (Cain, 2016:97).

Most contemporary theorists have discarded the view that thoughts are (necessarily) imagistic (Margolis & Laurence, 2019). Apart from the fact that not everyone is even capable of experiencing vivid mental images[34], an obvious critique is that many concepts are abstract in nature (e.g., the concept KNOWLEDGE) and are thus not suited for an imagistic representation (Cain, 2016:97). Moreover, concepts are often general categories which apply equally to many distinct things (Cain, 2016:97). For example, the concept ANIMAL, or even its more specific subclass DOG, could be built up from creatures of different colours, shapes, textures and sizes that an individual had encountered. The imagistic approach remains ambiguous as to which member of a category the concept would resemble, and how would it still apply equally to all of them. Moreover, in pointing at a given object, it is not always clear which aspect of it is being referred to (Wittgenstein, 1953:11; Quine, 1960). For example, a picture of a fish in a bowl could represent the concept FISH, FISHBOWL, or PET, not to mention the colours, shapes, and other features of the individual objects. Hence, the picture alone is insufficient to pick out the intended aspect/element that is referred to. What is needed to fix the content is something non-imagistic beyond the image, like a mental state such as an intention (Cain, 2016:98). However, Cain (2016:98) contends that such a state would require employing further imagistic concepts, which present the same problem, *ad infinitum*.

Another classical treatment is the LOTH discussed earlier, which, again, addresses this problem of intention by holding much of thought (as propositional/intentional attitudes) to occur in an internal system of symbolic, word-like mental representations with a language-like syntax and compositional semantics. Although I have supplied independent reason to reject an approach assuming a CCTM, it is worth briefly discussing Fodor's view of concepts and language acquisition as an influential part of discourse in the field.

---

[34] That is, a condition known as *aphantasia* (see Keogh & Pearson, 2018).

Fodor (1998:7) explains that, similar to the imagistic accounts of empiricist philosophers, his LOTH views concepts as mental particulars endowed with causal powers (e.g. the concept COW is satisfied by all and only cows). However, being amodal, his symbols do not correspond to particular visual experiences. Being strongly opposed to relativism, one of Fodor's *not-negotiable conditions on a theory of concepts* is that concepts are public; that is, they have to be identical between people, and not merely similar (Fodor, 1998:30). For Fodor, to acquire a concept is to get 'nomologically locked' to the property that a concept expresses, with no mediating hypothesis-testing or induction. What allows people to have the same concepts is a combination of an innate system of primitive concepts and "having the right kinds of experiences" (Fodor, 1998:127). Hence, complex concepts—or representations—are 'occasioned' from these experiences (based on innate logical combinatorial rules over primitive concepts) rather than inferred or abstracted (Fodor, 1998:127). In that sense, Fodor (1998:128) describes the relationship between a concept and the experience by which it is elicited as "brute causal" and intentional: something objective/public your thoughts represent, rather than a subjective construction based on varied experience. To acquire a language, then, on Fodor's account, is to *translate* a Mentalese concept into a target language—no concepts are 'learnt' (Churchland, 1980:160).

An immediate worry of this approach is that it limits explanation to thoughts that can be expressed logically: Fodor himself admits that his theory is unable to account for more abstract phenomena such as emotion, creativity, imagination, and consciousness. Yet, he rejects the idea that there could be a science of these phenomena at all, as they are not 'natural kinds'—the sorts of states over which counterfactual, law-like, reliable generalisations can be stated—and hence do not constitute appropriate domains for scientific investigation (Arias, 2019). One might object that, as causally-implicated parts of the natural world, those phenomena should, in principle, be amenable to scientific investigation, and that Fodor's (question-begging) response that they are not suited for the kind of CCTM account his theory assumes, should rather be seen as an attack on that assumption. As contended earlier, another concern with Fodor's anthropocentric account that assumes (at least) most of our linguistic knowledge and concepts to be innate, is that it fails to explain how such a system might have naturally evolved from ancestral structures (Churchland, 1980). Moreover, the strong requirement for having 'the right kind of experiences' for having exactly the same concepts, seems hard to meet given the range of different experiences people have of varied phenomena (as differently embodied/abled and contextually situated individuals) in the real world.

Despite Fodor's insistence that his version of a representational theory of mind is "the only game in town" (Fodor, 1975:406; 1998:23), most popular understandings of concepts take a slightly weaker view of concepts mental representations: not as a set of innate mental symbols, but as some kind of psychologically real phenomena that reflect experience. In fact, *all* of the main theories of concepts that I evaluate in this section describe concepts as mental representations of some sort (perhaps purely because, regardless of one's metaphysical commitments regarding concepts, most theorists accept that we bear *some kind* of psychological relation to them—see Murphy, 2000).

Whereas Fodor views concepts as stable, 'public' (i.e. identical) entities in individual minds, another approach takes concepts as universal abstract objects. As Fodor's, this approach aims to account for the fact that people seem able to communicate about the same topics by postulating concepts that are universally shared. In this view, concepts are the particular semantic contents of words which can be realised, in different ways, by individual minds (Peacocke, 1992). This view is typically associated with the work of Frege, where concepts are understood as different (standard) Fregean senses that can be associated with referents/objects in the world[35] (Margolis & Laurence, 2019).

One such contemporary account is given by Cain (2016), who insists that "concepts don't change or develop their content. If a particular concept changed its content, then it would become a different concept" (Cain, 2016:95). Instead, Cain uses Rey's (1994) distinction between concepts and conceptions: whereas *concept* refers to a fixed, universal content, *conception* refers to individual understandings of that concept that develops in relation to subjective experience[36]. To illustrate, Cain (2016:96) notes that two speakers could debate about a single concept, for instance, AARDVARK, despite having different conceptions about their nature (for example, whether they are herbivores or insectivores), without talking past each other.

Departing from the view that concepts are *things* of some kind, the third major approach considers concepts *abilities* that are particular to cognitive agents (Dummett, 1993; Bennett & Hacker, 2008; Kenny, 2010). In this sort of operational definition (grounded in behaviourism), a concept like PEACOCK is understood as *the ability* to draw certain inferences about what peacocks are, in order to distinguish peacocks from non-peacocks (Margolis & Laurence, 2019). The most prominent reason for adopting this perspective is a deep scepticism regarding the existence and utility of mental or abstract entities, that is tracible to the later writings of Wittgenstein. Whereas traditional philosophical theories of meaning were aimed at describing sense-giving entities that transcend individual expressions, Wittgenstein (1953) contends that meaning emerges as part of an activity, a *language game*, in a particular instance. He rejects the search for a singular, essential core of meaning (common to all instances of use) as dogmatic, maintaining instead that different uses are connected only in terms of *family resemblances*, as "a complicated network of similarities overlapping and criss-crossing" (Wittgenstein, 1953:66). Similar sentiments are expressed by Quine (1960), who questions the utility of appealing to universal objects that reside outside of the causal realm.

As noted, it may also be the case that different disciplines (e.g. formal semantics, philosophy of language, behavioural psychology, cognitive science, etc.) have different aims, and hence, it may not be so much a confusion about what *concepts* are, as much as different senses of the word being employed to play different kinds of explanatory roles for different purposes.

---

[35] For example, the same referent (e.g. Venus) can be associated with, or picked out by, different senses (e.g. 'the Morning Star', 'the Evening Star', 'the second planet from the sun', etc.), where each offers different information about the referent, and can be considered a different concept.

[36] Murphy (2000:5) takes an inverse approach: using *concepts* to refer to mental representations of classes of things, and *categorie*s in reference to the classes themselves.

For my purposes, approaches that regard concepts as universal entities seem unfit, as it seems to deal with (reifications of) abstractions from concept use rather than offering a (computationally tractable) account of individual concept acquisition. Similarly, adding to the apparent untenability of CCTM, Fodor's understanding of concepts in a purely causal-intentional manner is unfit, as its strong dependence on innate cognitive materials (not further explained) for acquiring a universal concept by 'locking' to certain properties in the world, coinciding with the strong requirement for the ideal kind of experiences for locking onto the 'right' properties for each given concept, seems a very *ad hoc* and (ironically) computationally intractable means of modelling all of human language acquisition and use.

As for the abilities view, although it (arguably) rightly disposes of ideal universal entities and gives a more cognitively realistic/grounded account, there does not seem any principled reason it cannot coincide with a weaker, subjective notion of mental representations to account for the phenomenology of thought (e.g. the associations one recall when asked for one's concept of *X*, and the ideas one can describe in language). If proponents can accept that we use inferences about objects of perception to discriminate between them, one might easily refer to those psychological inferences/categories themselves as *concepts*, given that they are what lie behind our concept-abilities—including our ability to call some form of representations of certain 'concepts' to memory and communicate them.

In light of these considerations, the account I advocate here is an understanding of concepts as mental abstractions, associations and inferences from individual experience, that are involved in such abilities as recognition, categorisation, and language comprehension, which tend to be, at least partly, directly accessible to memory. One could consider these mental representations in a weaker sense—i.e. complex psychological states of some sort, rather than clear-cut entities or specific images—as derived from a diverse range of perceptual experiences. Thus, my use of *concept* refers to these subjective mental constructions, which is also a way in which we typically employ the word (and our ways of talking about those states likely impose more order on them than they actually have). As for the central claim of Fodor and abstract entity theorists that concepts must be universal, I aim to provide an alternative account to explain how individually acquired concepts can be sufficiently similar for communication to succeed, even if they are not (instantiations of) exactly the same universally-given concept[37].

To substantiate these claims, I integrate evidence from cognitive science (§2 of ch.II) and cognitive linguistics (§3 of ch.II) to extract insights for a more cognitively realistic, grounded theory of concept (and language) acquisition. In doing so, I aim to account for (i) how we are able to learn language and acquire concepts that are sufficiently similar for communication to succeed; (ii) how that is possible given our unique set of real-world experiences; and (iii) how that is explicable in terms of general cognitive mechanisms and abilities that evolved naturally. As will become clearer through in this chapter, I follow the embodiment theoretic/cognitive linguistic stance on the nativist/empiricist debate: that concepts—as specified above—are constructed from experience, but that this process is only enabled through (and mediated by) innate, species-specific embodied (cognitive, perceptual, affective, sensorimotor, etc.) mechanisms and abilities. By identifying the nature of such general, underlying embodied factors, I hope to clarify what sort of key mechanisms should be accounted for if one wishes to construct an artificial agent approaching human-like command of natural language.

---

[37] Although, if the abstract view makes no metaphysical claims and is taken merely as an understanding of concepts as abstractions from use, then it is not necessarily in conflict with the view I propose.

In what follows, I evaluate major contemporary accounts regarding the acquisition and structural nature of concepts—understood as mental representations in a weaker sense, as specified above. Before doing so, it is worth mentioning a major traditional approach that has mostly fallen out of favour, as it is precisely the problems it presented that motivated the more recent approaches (and a lot of their strengths are best revealed in contrast) (Margolis & Laurence, 2019). Another big source of inspiration/turning point in the field was Eleanor Rosch's studies on typicality effects, which I also discuss briefly. Both of these discussions will also be useful for my evaluation of NLU approaches in AI, in chapter III.

### 1.1.2. The classical theory

The standard traditional account of conceptual structure is known as the classical (or definitional) theory. Apart from the imagistic account, this view dominated since antiquity, so much so that serious alternatives were only starting to be developed in the 1970s (Margolis & Laurence, 2003:191). From an ontological perspective, this approach fits best with the concepts-as-abstract-entities view given above. Students of philosophy are probably familiar with the activity of conceptual analysis; when prompted to define a common term, a lecturer keeps giving counterexamples that prove a given definition needs further clarification[38]. This process of testing a given definition against new examples, and consequently rejecting and/or updating it if the result does not fit common judgment, is common in philosophical argumentation (Murphy, 2002:10-11). In fact, this distinctively *a priori* activity is something that is often taken as the essence of analytic philosophy[39], which further motivated the prominence of this view (Margolis & Laurence, 2019).

Implicit in this activity is the classical view of concepts, according to which, for each class/set (i.e. the word), there are certain necessary and sufficient conditions that can be used to determine membership (i.e. the definition). Whereas *necessary* conditions refer to distinctive attributes that are essential for class membership, even if they exclude other requirements, *sufficient* conditions are conditions that, if all are met, are enough to determine class membership (Murphy, 2002:11). These conditions take the form of simpler concepts that constitute the type, each of which is assumed to be similarly definable (Margolis & Laurence, 2003:191). Some key claims of the (traditional) definitional view include, firstly, that concepts are psychologically represented as definitions. Secondly, it used to be assumed that there are no in-between cases for objects of different types: every object either falls under a given category or does not (where a definition *attempts* to describe such a category in terms of its necessary and sufficient conditions). This was a significant aspect of the classical theory's philosophical background, following from the *Law of Excluded Middle*[40] in logic. Thirdly, the classical view does not distinguish between class members: anything that satisfies the definition (i.e. falls under the same class) is equally considered a member (Murphy, 2002:15).

---

[38] A popular example from ancient Greek philosophy is Plato defining man as a "featherless biped", to which Diogenes responds by presenting a plucked chicken and declaring, "Behold! I've brought you a man" (Laërtius, 1925, VI:40).
[39] Moreover, traditional logic relies on statements of the sort 'All horses are animals', which necessitates clear specification of membership to sets and subsets (Murphy, 2002:15).

This intuitively appealing approach has a number of advantages. Much of its power is due to the fact that it seems to account for a host of key mental phenomena. Categorisation, for instance, is often considered one of the most fundamental parts of our conceptual ability, as most of our higher cognitive functions—and survival—depend on our ability to swiftly and reliably classify things in our environment (Valera et al., 1991). The straight-forward account of categorisation that the classical theory offers, is to analyse the concept and check whether it applies to the object in question (Margolis & Laurence, 2003:191). On the other hand, its account of concept acquisition is like the categorisation process in reverse: concepts are acquired by learning their necessary and sufficient constituents in light of one's experience (Murphy, 2002:15). In brief, the classical theory offers a simple unified account of concept acquisition, categorisation, and reference determination (Margolis & Laurence, 2003:191).

Despite its straightforward elegance (or perhaps because of it), the classical theory has few adherents today (Margolis & Laurence, 2003:192; Murphy, 2002:38). As discussed earlier, although it may be compelling to assume that our abilities of categorisation, which we tend to employ with relative ease, can be reduced to simple, clear-cut rules, it is usually the case that our conceptual abilities and conventions of language, as well as the world itself, prove too complex. Moreover, perhaps the most pressing objection is the sheer lack of uncontroversial definition examples. A popular example comes from Wittgenstein (1953) who shows the difficulty of finding a comprehensive definition for the class GAME. Thereby, he argues:

> And one has to say this in many cases where the question arises 'Is this an appropriate description or not?' The answer is: 'Yes, it is appropriate, but only for this narrowly circumscribed region, not for the whole of what you were claiming to describe.' (Wittgenstein, 1953:3).

Even the stock example of BACHELOR defined as an 'unmarried man', is controversial; for instance, is the Pope a bachelor? (Margolis & Laurence, 2019). This difficulty even arises in more technical domains; for example, in disputes over species membership in biology; in cosmological controversies, like whether or not

---

[40] That is, the rule that every unambiguous statement is either true or false.

Pluto is a planet; and in law when a judge has to decide whether a borderline case counts as a legal transgression (Murphy, 2002:17-18).

Proponents might respond that the fact that comprehensive definitions are hard to formulate, does not necessarily mean that none exist (Margolis & Laurence, 2003:194). However, if it is the case that we tend to be unable to think of precise definitions that hold, it questions both the usefulness and accuracy of this view as a theory of our conceptual abilities. Arguably, a part of the problem could also be attributed to the natural development of human language conventions. As we become familiar with a word, like *game,* we may start using it to refer to a host of things that seem loosely related, without ever thoroughly evaluating the specifics. Thus, our intuitive judgment might include more members to a category than we could coherently support with a definition (perhaps purely because of a lack of a better word) as we draw from a limited set of conventional linguistic constructs to describe patterns (we are able to perceive) in complex real-world phenomena.

According to Murphy (2002:21), psychologists now largely consider it a necessary feature of our mental categories that they are 'fuzzy' (or vague); that is, most properties we use to distinguish objects from each other either fall on a spectrum (e.g. hot versus cold), are relative (e.g. big versus small), or consist of complex combinations of features. If our aim were complete accuracy, each object/boundary case may end up requiring its own category (*ad infinitum*), but that will cease to be useful—not to mention highly inefficient. Instead, we want a small number of sufficiently informative categories that we can use to distinguish and make inferences about phenomena, imprecise as they may be.

### 1.1.3. Typicality effects

One of the main critics of the definitional account was Eleanor Rosch, who provided a lot of the key evidence for its shortcomings that helped spur on the development of most contemporary accounts. Rosch (1977) provided empirical evidence of *typicality* effects in people's concepts. That is, we often view some objects as more typical, or representative, of a given category than other of its members[41]. For instance, an eel is not considered as typical a fish as a goldfish; it seems people are able to rate different members of the class fish in terms of how fish-like they are[42] (Murphy, 2002:22).

Based on their studies, Rosch and Mervis (1975) argue that items come to be viewed as typical in proportion to their family resemblances/overlapping attributes with other members, as well as their lack of resemblances with members of other categories. That is, typical category members are not necessarily those that are the most common, but those that tend to have most of the properties common in their category, and tend not to have those of non-members (Murphy, 2002:31-32). For instance, even though chickens, which fall under the category BIRD, are quite common, a chicken is not considered a highly typical member of the category as several of its properties (i.e. can't fly, are butchered) are atypical.

---

[41] This is similar to Putnam's (1975) notion of a *paradigm case*.
[42] This was also influenced by Wittgenstein's (1953) notion of family resemblance.

Rosch, however, remained ambivalent regarding how exactly typicality structure is represented, and several theories emerged to interpret her findings further (e.g. Barsalou, 1985). A popular interpretation is the *prototype theory*—a bunch of related approaches that employ some notion of a 'prototype' to explain typicality effects and other psychological data.

## 1.2. The prototype theory

A *prototype* is a complex mental representation which, rather than specifying necessary and sufficient conditions, specifies properties that a member of a certain category *tends* to have (Cain, 2016:101). The general formulation of the theory takes categorisation to be a feature-matching process. On one interpretation, every category is represented by such a single prototypical example—an ideal member encapsulating all the attributes most likely found in the category—and typicality is measured in terms of similarity with this prototype (Murphy, 2002:41; Margolis & Laurence, 2003:197). However, as in the earlier imagistic theory, some are sceptical that a single example could represent an entire category:

> For example, is there really an 'ideal bird' that could represent all birds, large and small; white, blue, and spotted; flightless and flying; singing, cackling, and silent; carnivorous and herbivorous? What single item could pick out penguins, ostriches, pelicans, hummingbirds, turkeys, parrots, and sparrows? It seems unlikely that a single representation could encompass all of these different possibilities (Murphy, 2002:42).

On a more popular interpretation, a prototype is a *summary representation* describing the common features of the category as a whole, with some being more important than others (Murphy, 2002:42). This representation unifies all the common features in a category as a weighted list of properties, which means that even contradictory features can be included. For example, for a certain class of animal (say, BIRD), even if the property 'herbivorous' is weighted most heavily, 'carnivorous' could still have a significant, yet smaller, weighting. The 'representation' itself could be described in terms of Rosch and Mervis' (1975) account above, where the weights are understood as family-resemblance scores. Under this interpretation, new items are categorised by essentially calculating the similarity of the item's features to those in the category representation: for each it has in common, it gets credit; for each feature it lacks, it loses credit. After all of the observed features have been accounted for, the sum of the negative credits is subtracted from the sum of the positive credits. If that number exceeds a certain threshold (*the categorisation criterion*), then the item is considered to fit the category (Murphy, 2002:42). Thereby, the higher an item is scored for its features, the greater the chance that it will be considered a category member. This contrasts with the definitional account in that class membership is more seen as a matter of probability: an object need not satisfy every property of a concept's structure, just a comparatively large number of them (Margolis & Laurence, 2003:196).

The primary concern of the cognitive scientists who developed the prototype theory was with categorisation, particularly to account for two phenomena that emerged from empirical studies, namely (i) typicality judgments (described above), and (ii) the fact that people are quicker to categorise certain items belonging to a given class/concept than others (Cain, 2016:103). Another strength of the theory relates to concept acquisition. Most cognitive scientists agree that, even if some of our concepts are innate[43], most are learned on the basis of experience (Cain, 2016:103). Under the prototype view, your acquisition of the concept TREE, for example, is based on your varied experiences of trees, from which you build a relevant prototype consisting of salient properties you perceived (Cain, 2016:104).

Moreover, Murphy (2002:44) lists some failures of the definitional account that the prototype theory sought to correct on. Firstly, categorisation does not depend on satisfying any particular defining features; as long as an object has enough features in common with those of the concept's representation, it can become a member of that class. Secondly, this family-resemblance approach offers a good explanation of controversial borderline cases, as it is possible for an item to have features that overlap with the criteria of more than one category. Third, addressing this borderline phenomenon, it makes sense that typical items will be easier (and hence faster) to categorise than less typical ones, as they reach the categorisation criteria for a particular class faster. Fourthly, this theory accounts for empirical phenomena that seem to contradict the logical foundations of the definitional approach: for some concepts (as mediated by language conventions), it is possible for some concept A to be similar to B, and concept B to be similar to C, but A to not be similar to C (because the features that A and B share are different than those shared by B and C). For example, Big Ben is considered a clock, and clocks are typically considered furniture, but Big Ben is not typically classified as furniture. The prototype view can explain this in terms of typical features of FURNITURE that Big Ben lacks, despite satisfying typical features of CLOCK (Murphy, 2002:44-45). According to Murphy (2002:45), this kind of intransitivity is not possible in the classical (set-theoretic[44]) view, because any set would 'inherit' all the conditions of the definition of its superset. As such, there is no clear way to state how a definition that includes CLOCK would not also include FURNITURE.

Despite such relative improvements, the prototype theory presents its own difficulties. Margolis and Laurence (2003:197) maintain that both the definitional and prototype approach suffer from problems of error or ignorance: the fact that people seem able to possess a concept even if they have wrong ideas about the items it applies to, or do not have sufficient information to pick it out uniquely. Moreover, prototypes based on mere lists of features are notoriously bad at reference determination. For example, although prototypical grandmothers have features like 'has grey hair', 'has wrinkles', 'wears glasses', 'is a good cook', etc., there are many people that satisfy most of these features that are not grandmothers, and people who are grandmothers who lack most of them (Margolis & Laurence, 2003:197). This seems an advantage of the definitional account—the fact that it includes notions of causal connections rather than mere surface-level correlations, identifying what is essential and not just salient.

---

[43] That is, at least in the sense that the same concepts reliably develop in humans (see Samuels, 2007)—although it remains to be proven that the concepts themselves are innate and are not merely due to the same kinds of experiences and innate perceptual mechanisms.

[44] That is, the presumption (which is often, but perhaps not necessarily, held by proponents of the definitional account) that concepts are understood in terms of nested classes with superordinate (i.e. more general) and subordinate (i.e. more specific) classifications/definitions. For instance, the concept BIRD is a category belonging to the more general ANIMAL class, and encapsulating the more specific RAVEN class.

Another objection given by Margolis and Laurence (2003:197), is that many concepts simply lack prototypes. Fodor uses the example that, even if there are prototypical grandmothers, there is no prototype of CHAUCER'S GRANDMOTHER (Fodor, 1981:297). Arguably, however, a defender of the theory might respond that, if there is only one item in a category, the prototype would contain all the features that uniquely pick out the single item. Fodor also objects that prototypes have trouble composing into coherent complex concepts (Fodor 1998, Fodor & Lepore, 1996). Fodor and Lepore illustrate this with the concept PET FISH: whereas the prototype of PET should encode properties associated with more typical pets like cats and dogs, the prototypical FISH might be a trout or herring. From these, Fodor and Lepore find it difficult to see how their combination can produce the concept PET FISH, which they maintain is something like a goldfish (Fodor & Lepore, 1996:263). Firstly, I would object that, perhaps if a particular complex concept is common enough, as PET FISH is, it might be treated as its own concept with its own relevant prototype[45]. For more general cases[46], however, it seems what is needed is a notion of context-sensitivity in composing concepts, where one only selects those features that seem relevant from their respective representations (like 'is domesticated' or 'is owned by a person'). How one goes about judging this relevance remains a problem for a weighted feature-list approach—and, arguably, likewise for Fodor's atomist approach (see §3.2.2 of ch.II).

As a final critical point, I would contend that we are quicker to mention certain concepts than others without necessarily considering their features as representative of their class, but also on the basis of other heuristics. For example, some concepts like APPLES and BACON are often used as examples of the classes they fall into (i.e. FRUIT and MEAT), which, arguably, seem more a matter of (contingent) sociocultural convention. With apples, their features are actually not as typical in the FRUIT category: they tend to be less sweet, and their more solid texture is quite uncommon amongst fruit (which tend to be juicier). Likewise, the features of bacon are also quite special amongst meat (e.g. 'is cured', 'is eaten at breakfast'). Rather than being describable in terms of feature commonality or category exemplification, such examples beg further explanation.

The next contemporary account I discuss is the *exemplar theory*. As another attempt to explain typicality effects, the exemplar view resembles the prototype one in some respects, and departs quite radically in others.

---

[45] This also only really seems like an issue for an approach like Fodor's that assumes a language-like sentential conceptual structure, or at least a very close link between linguistic and conceptual structure.
[46] Cain (2016:105) maintains that compositionality is particularly problematic in cases where the features of the combined concepts stand in conflict. He uses the example of ANGRY COW, where the feature 'docile' is weighted heavily in COW and so conflicts with ANGRY.

### 1.3. The exemplar theory

A distinctive factor of the exemplar theory is that it rejects the notion that people have a single mental representation that somehow encompasses a whole concept. On this view, a class/concept is neither a definition that applies to all members, nor a list of attributes members tend to have. Rather, one's concept, say, of CAT, consists of all cats that one remembers. Along with the prototype view, the exemplar view is one of the most popular current theories and also uses notions of resemblance to account for the same kinds of psychological phenomena (Murphy, 2002:49).

According to the exemplar theory, a concept is the *set of memories* one has of encounters with things belonging to that category—some more salient and vivid, some more fuzzy and incomplete. Thereby, this approach aims to account for flaws in memory. Although a concept does not accurately represent all potential members, it can still be used for making decisions regarding objects of that type in general, as they bear a certain similarity to past encounters (Murphy, 2002:49; Cain, 2016:104). A novel object may be very similar to a handful of past encounters, and vaguely similar to a few hundred. If, broadly, most things to which it bears strong similarity were classified as dogs, then one would conclude it is a dog. For example, in seeing a golden retriever, it might remind you of other golden retrievers you have seen, which you consider dogs, and hence you would conclude that it is also a dog (Murphy, 2002:49). Thus, as in the prototype theory, categorisation is based on similarity: concepts are acquired through perceptual encounters with similar objects, constituting categories (Cain, 2016:104). Then, on Medin and Schaffer's (1978) account, similarities with members of various categories (as particular objects of memory) are weighted, and the category of the exemplar (i.e. category member) with the highest weighting (closest match) will be one that is decided on.

Murphy (2002:50) lists some common aims of the prototype and exemplar theories. Firstly, the exemplar theory also aims to avoid the problems of the definitional approach by not saying anything about specific defining characteristics. Secondly, it aims to explain typicality effects, although slightly differently: the most typical items are the ones judged to closely resemble many members of a category. Basically, the more similarities an object bears to recollected members of a given concept, and the fewer it has to recollected non-members, the more typical it will be. Thereby, thirdly, it aims to account for borderline cases, i.e. as those that are equally similar to remembered members of different categories (Murphy, 2002:50).

Given its similarities to the prototype approach, the exemplar theory is subject to similar objections, particularly regarding the issue of concept composition (Cain, 2016:104). Here the objection seems especially pertinent, where there is allegedly a lack of a coherent, unified notion of a concept (as categorisation depends on feature-similarity with individual members, rather than a single definition/list of features). Moreover, Murphy (2002:50) maintains that many people find this view very counterintuitive. Concerning the phenomenology of thought, many people do not experience consciously recalling exemplars of a given concept in order to categorise something new as a member—although, arguably, people do not typically consciously experience a definition or a list of features in such instances either. This also presents the opposite problem, in that both theories fail to explain how people do sometimes have a conscious experience of making reflective judgments which seem to go beyond similarity comparisons (Margolis & Laurence, 2019). Another point on phenomenology is that people feel they have knowledge about, say, *birds* in general, rather than just knowing things about individual exemplars (Murphy, 2002:50). Not only are we able to make inferences regarding causal factors that underlie membership to categories (e.g. functional attributes like judging a lamp as such even if it resembles a space ship, or guessing that an animal can fly even if you have never seen it fly), we can also incorporate explicit definitions we heard or read to update our understanding; for instance, learning that whales are mammals, even if they have unusual features for mammals.

This relates to another point, which is that this approach requires one to have specifically categorised memories (Murphy, 2002:50), and whilst it makes an attempt at explaining how new exemplars are added to an existing concept, it fails to explain how the earliest exemplars were categorised. For instance, if your basis for categorising a golden retriever as DOG depends on having seen a golden retriever and knowing it was a dog, how that knowledge originally came about still begs for an explanation. Moreover, concept acquisition (and, hence, learning the contents of words) is arguably more a process of trial-and-error, of sculpting at a concept, rather than assuming all former exemplars were reliable as the exemplar theory seems to do. It may have been that I miscategorised a given exemplar, only to update it later (for example, thinking dolphins are fish rather than mammals). What the theory lacks is an explanation of how exemplars are updated throughout the learning process, and the non-perceptual factors that may be involved (such as language conventions, non-linguistic background knowledge, etc.).

The prototype and exemplar accounts arose in response to the definitional approach, and specifically aimed to account for the data that it found problematic. Although neither seems complete, these empirically-grounded accounts at least help to illustrate the role of memory and perception in concept construction. In turn, the theory theory, which I discuss next, rose in response to their respective shortcomings and, in some sense, builds on them (Murphy, 2002:108).

### 1.4. The theory theory

The theory theory draws from developmental psychology to characterise the nature of infants' concept acquisition and development (Cain, 2016:109). Also called the *knowledge approach*, it considers concepts as part of our general knowledge framework—that is, that we do not learn concepts in isolation, but as part of our broader understanding of the world. In learning the concept, say, ANIMAL, we integrate our general (albeit flawed or incomplete) knowledge of relevant domains such as biology, genetics, etc. and may make inferences about properties that are not readily observable (Murphy, 2002:60); for instance, that it needs food and sleep, reproduces, has organs, will have similar-looking offspring, etc. This touches on an earlier point, that concept learning seems a more integrated process than the prototype and exemplar theories lead on.

This relation between concepts and knowledge acquisition goes both ways: our general knowledge influences our learning of new concepts, and vice versa (Murphy, 2002:60-61). For instance, if you learn some fact regarding the cognitive abilities of octopuses, this could affect your more general understanding regarding animal intelligence, and if your experience of a particular object seems to conflict with your general knowledge you may call it into question. In general, the knowledge approach emphasises that concepts should be consistent with the rest of your knowledge base (Murphy & Medin, 1985; Keil, 1989). To maintain this consistency, categorisation and other conceptual processes may involve processes of reasoning and inference, drawing from a range of both formal and informal information sources (Murphy, 2002:61). For instance, in seeing a brightly coloured bird with a similar-looking duller bird, you may reasonably infer that the brighter one is a male, given your general knowledge regarding birds (and basic biological principles)—even children are able to infer that the small fuzzy birds following a larger bird is its babies (see Gelman & Wellman 1991; Keil, 1989).

The reason for the name 'theory theory' is the conviction that concepts are embedded in psychological structures resembling scientific theories. That is, that they embody an explanatory schema (i.e. a set of rules or principles) that is used in the categorisation process (Margolis & Laurence, 2003:200). On this view, children are portrayed as "little scientists" who construct theories that develop and change over time, based on experience (Cain, 2016:109). Moreover, although experience plays a key role, Cain (2016:110) maintains that proponents of this approach tend to credit children with a substantial innate endowment regarding theory construction. Firstly, children are assumed to instinctively carve up the world into distinct domains (e.g. the physical, the mental, etc.), and construct distinct theories to deal with each; for instance, using folk-psychological theories to presume the existence of (unobservable) mental states that cause people's behaviour (as discussed earlier). Secondly, the cognitive mechanisms that children use to construct such theories are considered innate (Cain, 2016:110-111). Thirdly, some theories are themselves considered innate, and subsequently updated/replaced through further development (Carey, 2009).

In contrast to the prototype and exemplar accounts, the theory theory maintains that people do not rely on simple feature observation for constructing concepts; rather, they specifically select those features that their general knowledge indicates as important. This process includes making inferences and postulating information that is not readily observable (Murphy, 2002:63). Moreover, much weight is placed on the use of prior knowledge to reason about more fundamental relations between items in a category, such as their functional purpose and/or causal relationships between properties. This idea was partly inspired by Barsalou's (1985) insights regarding the relation between ideals and typicality. His studies suggested that, even if most of the items in a category people had experienced were only mildly effective at their intended purpose, items closer to the ideal would still be judged as more typical (Murphy, 2002:62-63). For instance, even if most pencil sharpeners one has used before have been faulty—breaking more pencil tips off than sharpening them—the (proto)typical concept of a pencil sharpener would still be an 'ideal' one that sharpens as intended.

A special strength of the knowledge approach is its ability to explain people's tendency towards essentialist thinking; that is, the psychological phenomenon that people are prone to understand category membership less as a matter of readily observable features in an instance, and more a matter of it containing some 'essential' internal quality, based on relational and historic factors (Margolis & Laurence, 2019; Medin & Ortony, 1989). For instance, subjects that were shown a picture of a skunk, but were told that it was historically a raccoon that underwent a series of surgical changes to resemble a skunk, still considered the animal a raccoon (Keil, 1989). Thus, rather than comparing the immediately observable features of an object with a previously acquired check-list of properties, the knowledge approach takes into consideration that categorisation may depend on notions of supposed *essences*, or underlying hidden causes of the readily perceivable features (Margolis & Laurence, 2003:199; Cain, 2016:110). If a child does not already have a clearly-developed understanding of any such hidden property, Medin and Ortony (1989:184) maintain that they might just start off with an 'essence placeholder'.

According to Murphy (2002:63-64), a clear limitation of this approach is that most of what constitutes a concept cannot depend on prior knowledge. That is, even if one has previous knowledge that helps one make sense of an object, it is only through observing the object that one learns its peculiar features, as most are unpredictable. Many contend that this key aspect of concept acquisition—an empirical learning mechanism, foundational to the former two accounts—is neglected in the knowledge approach, and a complete account would have to include an integrated explanation of all these aspects of concept learning (Murphy, 2002:64; Wisniewski & Medin, 1994). Likewise, some argue that the knowledge approach does not clearly account for more rapid categorisation judgments and typicality effects, as the empirical models could, and hence we may require some version of a "dual theory" that combines the strengths of the knowledge and prototype models (Margolis & Laurence, 2003:200).

Other points of criticism against the knowledge approach concern the issues of reference determination and concept stability. Margolis and Laurence (2003:200) argue that, as proponents of the knowledge approach typically allow for people having rather anaemic or flawed theories—especially considering the aforementioned 'essence placeholder' contains relatively little information—implying that concepts will probably not contain adequate information to pick out a correct referent. Moreover, if people have different theoretical frameworks through which they interpret their concepts, and a concept is determined by its role in a theory, rather than by its composed constituents, Margolis and Laurence (2003:200) contend it is unclear how people can talk about a particular concept as if it is even the same thing. To illustrate, they use the example of two people having conflicting ideas regarding the plague, one considering it caused by divine retribution, the other assuming a more scientific account. In order to dispute about the nature of the PLAGUE, they argue that the theory theory should be able to explain how it is possible for them to talk about *the same* thing despite their divergent theories (Margolis & Laurence, 2003:200; Cain, 2016:112). Fodor, as a strong believer in 'publicity' as a necessary requirement for concepts, makes a similar argument: that the theory theory does not allow different minds to have the same concepts, or one mind to have *the same* concepts over time (Fodor & Lepore, 1992; Fodor 1987).

Firstly, I would respond that the burden of proof is rather on the proponents of Fodor's position to prove that concepts are, in fact, *the same* to everyone, and that they do not change and develop over time. As suggested earlier, I believe that having sufficiently similar concepts associated with words are enough for communication to succeed; there is nothing independently (outside of a CCTM) that requires people to have the exact *same* concepts. However, having sufficiently similar concepts despite having different theoretical understandings does require having similar experiences in other regards—such as perceptual and linguistic-conventional[47]—which the knowledge approach does not clearly acknowledge. Secondly, Margolis and Laurence's critique regarding the difficulties for incomplete or flawed personal theories of concepts to uniquely pick out referents is not necessarily such a big issue; it seems to underestimate the rate at which people are able to acquire adequate interpretations of concepts in social settings. Following Wittgenstein's (1953) account of language games, we also employ the ability to infer reference determination and meaning from context, regardless of established senses of words—which, again, arguably requires some integration of salient features, understandings of prototypes (which may perhaps vary in different settings[48]), and broader non-linguistic knowledge.

For an approach that could potentially accommodate all these constraints, I next look at associationist theories of thought, which fits in well with the connectionist paradigm. Rather than necessarily being a new alternative to the aforementioned theories, it can be taken either as an extension or further explanation of how concept structures are formed empirically and implemented in the mind/brain, although different versions of associationism carry more or less stringent requirements and assumptions. For my purposes, I mainly try to explore an account of mental structure that is both cognitively plausible and can accommodate the desiderata for concepts discussed in this section, as well as the kind of empirical, neural-network learning approaches of AI, discussed in §1 of ch.III. As there are many relevant factors that remain to be discussed, particularly regarding the general (embodied) cognitive mechanisms, biases and processes that allow us to acquire concepts (and language), the following section is merely a preliminary framework for a more thorough theoretical discussion at the end of this chapter (in §4).

---

[47] Arguably, restrictions imposed by language conventions make it easier to speak about different experiences under some unified label—one that may apply to a broad or loose range of corresponding phenomena, purely because there is no better name available (whereas, if there were names for nuanced distinctions, like God-given-plague and flea-given-plague, people might use those instead).
[48] For instance, the typical fish that may come to mind in a pet store (as a pet) or at a restaurant (as a meal) may differ.

## 1.5. Associationist theories of thought

Associationism, in some form or other, is one of the oldest and most widely held approaches to understanding the mind (Mandelbaum, 2017). Rather than a particular theory of cognition, it refers to a collection of different, but related accounts, generally ascribing to a view that different experiences become associated in the mind as a result of their co-occurrence in experience (which, for instance, leads certain concepts to elicit certain others). Hence, the history of an organism's experience is considered a crucial factor in defining its conceptual structure. An idea tracing back to Plato and Aristotle, associationism became the backbone of empiricist accounts of human thought and behaviour: from the British empiricists, through to the behaviourists and modern-day connectionists (Encyclopaedia Britannica, 2020). It also features prominently in psychological doctrines such as anti-representationalism and domain-general learning (Mandelbaum, 2017). Part of its popularity has to do with the sheer breadth of its application to explain different aspects of cognition: as a theory of learning (as in behaviourism), a theory of thinking (e.g. *System 1* in Kahneman's Dual-Systems theory, 2011), a theory of conceptual structure, or a theory of how thought is implemented in the brain (e.g. connectionism). However, these are largely independent, and no particular theorist has quite ascribed to all of them simultaneously (Mandelbaum, 2017).

In Empiricism, most of thought is assumed to be learnt through experience (apart, usually, from the mental processes that enable such learning). Hence, a limited amount of innate machinery should be able to account for a wide variety of mental phenomena (Mandelbaum, 2017). In its early forms in Locke ([1689] 1975) and Hume (1738), associationism was put forward as a theory of such a domain-general mental function: that our ability to associate ideas is our most basic and general mental capacity, accounting for both learning and thought (Mandelbaum, 2019; Encyclopaedia Britannica, 2020). Hume's account is, at base, a description of how perceptions ('Impressions') determined thought (as successions of 'Ideas'); predicated on the assumption that there are no *Ideas* that were not originally given in experience. Hume's analysis consists of three kinds of associative relations that are formed through experience: cause and effect (e.g. repeatedly seeing one event follow another); contiguity in time or space; and resemblance.

Moreover, Hume uses associations to account for the transition from one thought to another—rather than using reason or logical inference, as proponents of LOTH/CCTM maintain—an idea which William James (1890) also described in his *Stream of Thought* approach. Hume writes:

> When the mind, therefore, passes from the idea or impression of one object to the idea or belief of another, it is not determined by reason, but by certain principles, which associate together the ideas of these objects, and unite them in the imagination (Hume, 1738:148).

There are obvious empirical differences between the views of thought transitioning as a result of associative, and of logical/inferential relations. Whereas logical transitions would mostly stay within a focused set of relevant contents, associative transitions can better account for the tendency of thought to move across different content domains, based on different experiences/ideas being 'called to mind' due to their (not necessarily logical) associations with one another (Mandelbaum, 2017).

After the British Empiricists' introduction of the view into philosophy, associative learning gained empirical support when the work of Ivan Pavlov motivated the behaviourist movement in psychology. Pavlov (1906) introduced the notion of *classical conditioning* as a modern version of associative learning, whereby animals learn (or are 'conditioned') to associate a given stimulus (e.g. the smell of meat) with another (e.g. the ring of a bell), such that the substitution of the former with the latter elicits the same response (e.g. salivating). In his account, the (instinctual) responses that are elicited remain the same; the only change is in the sensory stimulus that it gets associated with. As such, classical conditioning was considered too restrictive to account for all the novel behaviour organisms seem capable of (Mandelbaum, 2017). A broader account was offered by Thorndike (1911) who introduced the notion of *operant conditioning* as a general, active theory of learning by which an organism learns how to 'operate' in novel situations through creating associations between new behaviours and certain consequences. That is, learning to execute/avoid certain actions in given circumstances based on their historical positive/negative effects in similar situations. For instance, in his initial experiments, cats learned the novel behaviour of lifting levels in order to open doors that gave them treats (Thorndike, 1911).

Thorndike's work was popularised and extended by Skinner (1953), who highlighted the notion of *reinforcement*, and not just outcomes, as the foundation for learning new associations: the strength of the reinforcement associated with a given action, as well as the frequency of association between certain actions in certain situations, together help determine the likelihood those actions would be repeated in similar situations.[49] As the empiricists, the early behaviourists posited associative learning as a domain-general cognitive function that can apply equally to any contents (e.g. any sensory stimuli) (Mandelbaum, 2017).

Concerning conceptual structure, a simple associationist view is that information is stored in the mind in direct associative links between distinct mental states (e.g. entertaining the concept SALT and entertaining the concept PEPPER). According to Mandelbaum (2017), the statement that two concepts are associated means that there is a "reliable, psychologically basic causal relation" that holds between them, by which the activation of the one leads to the activation of the other, without any mediating psychological states (such as explicit rules, etc.). Such relations are strengthened through repeated experiences of the concepts together and cannot be broken through reason alone—only through *extinction* (i.e. no longer encountering the concepts together) or *counterconditioning* (i.e. forming a contrasting association). A consequence of this strong/pure view of associationism (which Fodor, 2003, calls 'Bare-Boned Association'), is that, all else being equal, associations between concepts are symmetric: the thought of SALT should activate the thought of PEPPER just as reliably as the other way around. However, even if this were the case, all else is rarely equal: not only does the order of learning seem to affect the causal sequence of activation (e.g. Thorndike, 1911; Skinner, 1953), but it is also affected by differences in the number of further associative links that were formed with the respective associated elements (Mandelbaum, 2017). For instance, if one associates RED with TOMATO, one may easier recall RED as a consequence of TOMATO (being one of its distinctive features) than the other way around, as there are many concepts that have RED has a salient property (e.g. BLOOD, APPLE, LAVA, etc.).

---

[49] Thorndike and Skinner's operant conditioning has been very influential in AI, especially in the development of aforementioned *reinforcement learning* algorithms, by which agents learn to execute certain actions based on positive feedback (reinforcement) and a reward function (see Russell & Norvig, 2010, Ch.18).

To some, this pure associationist view seems to contrast with propositional structures, which appear to have structures beyond the mere associative links between concepts but tries to indicate an *actual* relation between things in the world (see Mandelbaum, 2017). However, some have argued that associative links may also hold between more complex concepts like propositional elements (e.g. Mitchell et al., 2009) or valences (i.e. positive or negative emotions/attitudes), and that *all* conceptual structures need not be associative (or propositional) (Mandelbaum, 2017). Moreover, it would make sense that associations reflect actual relations between things in the world as that is where they are learnt from.

The successes (and broad application) of connectionist architectures for empirical learning methods in AI offers a strong case for the strengths of a form of pure associationist/empirical learning, albeit on a lower (sub-concept) level of abstraction. Connectionist networks are essentially an implementational interpretation of associationism, and, as described earlier, is considered by many as a realistic model of human thought (e.g. Smolensky, 1988; Elman et al., 1996), perceptive mechanisms[50] (see Buckner, 2019, for a review), and how our mind works on a neural level. On the psychological reading, these associations are defined functionally (as structured connections between different distributed features); on the neurological reading, they can be understood as the physical synapses connecting neurons. Note that here, in either case, the nodes themselves are not typically equated with discrete, high-level representational states like concepts (see Gallistel & King, 2009), but correspond to basic multimodal, sub-symbolic elements/features that may be meaningless on their own (e.g. low-level sounds, colours, shapes, sensations, feelings, etc.). Rather, a concept might be distributed over the network, through its relevant activation patterns through different associated features. However, this setup could arguably also allow for the instantiation of 'concept'-level associations (as originally understood), as well as more complex/systematic combinations of concepts, and as such is not necessarily in conflict with any of the aforementioned accounts.

What associationism does, arguably, offer, at least in the connectionist sense, is the ability to combine different experiences (whether simple or complex) of different modalities or abstraction levels (e.g. images, sounds, 'concepts', or more complex bodies of knowledge) in a semi-unifying way—that is, as a complex network of associations, with different relative weights between them—even if it does not amount to a coherent picture that can be described fully in terms of discrete constituents (such as clear-cut features, images, etc.). Nevertheless, there might be emergent structure in the network, corresponding to higher-lever features[51] like coherent images, sounds, or even sentences, which may present themselves as such to the cogniser. Thereby, I would argue it offers a (cognitively) plausible account of the needed 'dual theory' that combines the empirical 'list-of-features' or 'resemblance' accounts of the prototype and exemplar theories, with the broader background knowledge associated with concepts that the theory theory emphasises—an argument that will become clearer near the end of the chapter. This approach avoids the popular assumption that concepts, on a structural level, should be discrete/coherent entities (as in the definitional approach) or lists of/calculations over discrete entities (as in the weighted feature-matching in the prototype approach)— which, as I argued earlier, may either have been the result of the way in which we talk about concepts, or still symptomatic of earlier formal/computational approaches to language and mind.

---

[50] Currently, deep learning networks are considered the best models of perceptual similarity judgments in primates (Yamins & DiCarlo, 2016).

Moreover, what this approach acknowledges, and the prototype and exemplar accounts apparently fail to, is that the social/physical contexts in which concepts are learnt may be a part of what is associated with a given 'concept' (as a structured distribution of low-level features). As such, it may be misleading to offer an account of conceptual structure focusing merely on the (visual) features of the referents, just as focusing merely on abstract necessary and sufficient conditions, as either assume that the entire 'concept' should be summarised in a single, unifying way. By taking contextual learning into account, it might more easily account for our ability to infer the sense of ambiguous utterances in context, as the context itself may prompt for the activation certain patterns of associations rather than others. For instance, the concept of 'fish' that comes to mind may be different when one is in a restaurant, at the pet store, or at an aquarium[52]—something that the theory theory might approach an explanation for, but the prototype and exemplar theories fail to.

More than external context, however, what is also needed is an understanding of how our particular embodied experiences of the world factor into our acquisition and structuring of concepts. Similar sentiments are expressed by Dreyfus (1992:xxi), who argues that, given a mere list of facts (whether the features in the prototype theory, the exemplars in the exemplar theory, or the background knowledge in the theory theory) it is unclear how one might determine which information is relevant to retrieve for interpreting an utterance in a specific context. For that, rather than consulting lists of facts about entities or situations in the world, Dreyfus (1992:xxi-xxiii) argues that it seems we tend to rely on a sort of intuitive body-based 'know-how' for understanding language. That is, we need to imagine *feeling* and *doing* things in order to organise the knowledge required for understanding typical sentences—which are some of the factors I explore in the rest of this chapter.

---

[51] This has also been illustrated by neural network literature, such as in Buckner (2019), where a deep learning algorithm is explained in terms of combinations of increasingly higher-level features being extracted (e.g. lines, eyes, faces, etc.) using merely weighted patterns of interconnected nodes.

[52] Prinz (2004) also points out this limitation, and aims to account for it with his *proxytype theory*. A summary of his view is beyond the scope of this paper.

As I am not committing to a strong associationist account on a purely inter-conceptual level, I do not exclude the possibility for other psychological processes like performing arithmetic or logical deductions, given that, as mentioned earlier, classical computations can be implemented on a connectionist system (as seems indeed the case, given the physical constitution of the human brain). Hence, the view of conceptual structure that such an associationist/connectionist account offers, may be something like a distributed network of simple multimodal states (some of which may constitute their own complex 'concepts', multimodal scenes, or even sentences, but need not), that are mainly related through experience—although, arguably, some associations may be more or less instinctive, like associating the sound of screaming with a feeling of distress. These associations have different relative weights, but may (to some extent) actively depend on which parts of the associational map is prompted for in context (e.g. fish may be strongly associated with saltwater at the beach, with tanks at a pet store, or tartar sauce at a restaurant, etc.). 'Contexts' can also extend beyond the physical to, for instance, events (e.g. seeing a bloody knife at a Halloween party), linguistic context (e.g. being in an argument or not), one's mood (e.g. seeing a romantic film in a happy or sad mood) or current task (e.g. seeing a bottle of wine as a weapon when attacked), etc.—this point will be elaborated on the following sections.

This rough initial picture serves not only as a means to combine the strengths of different approaches to conceptual structure, but also as a (loose) interpretation of how they may be realistically implemented in our neural architecture, assuming that a connectionist view of the mind/brain is somewhat accurate (and judging by the surprising strengths of associative learning in AI, discussed in chapter III). This is by no means a complete account yet, but merely provides the groundwork for the rest of the chapter where I discuss different relevant factors more in depth. Amongst what remains to be described, is how exactly these concepts (as associative networks) are acquired through experience and used: what may be the relevant (internal and external) factors that are involved in the process, and how conceptual structure relates to language. In the following two sections, I address these in turn, starting with an investigation of the role of certain biological and situational factors involved in grounding human cognition. On that basis, in the section that follows, I discuss theories in cognitive linguistics for explaining how conceptual structure forms on the basis of these 'embodied' factors and general cognitive mechanisms, and how it informs our structuring/use of language.

## 2. Grounded theories of cognition

The debate about the role of the body in cognition has, in some form or other, been ongoing nearly since the beginning of philosophy. Ancient Greek philosophers like Anaxagoras and Aristotle considered the extent to which the body might play a role in human rationality, and such inquiries can also be traced through medieval texts (e.g. Aquinas and the Neoplatonists), to modern investigations such as those by La Mettrie, Condillac, pragmatists, phenomenologists, and many others (Newen et al., 2018:1). However, it was only during the 1990s, provoked by the prevailing internalist, cognitivist picture and its limitations, that the mutual interaction/influence between mind, body, and world became a serious area of multidisciplinary empirical study[53].

As discussed in §2 of ch.I, the multiple disciplines within cognitive science traditionally understood the mind as an abstract information processor, whose connections to the particularities of the human body, as well its particular (goal-driven) interaction with the external world, were of minimal theoretical importance. This coincided with approaches in AI that tried to model human behaviour using abstract symbol processing, and many were optimistic about its initial successes. Philosophy of mind likewise followed the zeitgeist, with theories like Fodor's LOTH offering compelling arguments for vindicating our intuitions about the formalisability of human reason, grounded in what was considered our best understanding of the mind at the time (Wilson, 2002:625). During the past couple of decades, this slowly started to change with further developments in those respective disciplines, especially in areas of *embodied cognition*. Contra the classical approach, this broad research programme emphasises the central role that an organism's particular embodiment and/or its situatedness in a given (physical/social) environment plays in the determination of its cognitive processes.

The general enterprise takes the view that cognitive mechanisms and capacities (including the formation of concepts) develop on the basis of real-time, goal-directed interactions between a biological organism and its surroundings (Cowart, 2019; Clark, 1999). Hence, the particular way in which an organism has evolved to perceive, and navigate in, its environment is considered fundamentally important for understanding its cognitive capacities and (social) behaviour. However, this area of enquiry is still in its early stages, and comprises a host of different, and not equally agreed upon, perspectives on how cognition can be understood as grounded (Cowart, 2019). These are broadly categorised under the banner of '4E Cognition', in reference to four of the main claims in the area: that cognition is Embodied, Extended, Embedded, and/or Enacted, respectively. In what follows, I briefly discuss these in turn to evaluate their significance for theories of language and concept acquisition in subsequent sections.

## 2.1. Embodied, embedded and extended cognition

Although the different strands of 4E cognition are united in their opposition to the internalist, abstract, brain-centred views of cognitivism/classicism, there are lingering controversies over various issues within these approaches. These include the nature of embodiment; the way in which extracranial factors are causally implicated in (or perhaps constitutive of) cognition; as well as the extent to which we can generalise from observing embodiment effects in one kind of cognitive process to others (Newen et al., 2018:2). 'Extracranial' here, can be understood in two ways; that is, bodily (involving mind and body) and extrabodily (involving

---

[53] That is, particularly in each of the subfields comprising cognitive science, such as developmental psychology, neuroscience, AI/robotics, and linguistics

mind, body and environment) (Newen et al., 2018:3). Moreover, there is disagreement between a strong and weak interpretation of the involvement of such extracranial factors in cognition: on a strong reading, cognitive processes are partly based on/constituted by extracranial processes, whereas a weak reading takes cognitive processes as *causally dependent* on extracranial processes, but not *constituted* by them (Newen et al., 2018:3).

Whereas all of the views in 4E see cognition as, at least, embodied (in the bodily sense), *embedded* (and situated) cognition is the claim that cognition is also partially dependent on extrabodily (physical/social environmental) processes (i.e. the weaker reading), where *extended* cognition is the similar, but stronger claim that cognition is at least partially constituted by things outside of the body; i.e. as if 'extending' into extrabodily components or tools (Newen et al., 2018:3). Embodiment, in the mere bodily sense, can be understood as the way in which our cognitive capacities are defined by our physical constitution, how the "anatomic make-up, physiological capacities, and sensorimotor capabilities of the human body directly inform our experience of negotiating the natural and sociomaterial environment" (Overmann & Malafouris, 2018:2). Such considerations were also encouraged by various lines of research questioning whether human cognition might work in a more distributed, less orderly way than was typically assumed: in ways that rely minimally on representations of the environment (Churchland et al., 1994) or that exploit 'irrational' (but perhaps evolutionarily useful) heuristics/shortcuts (Kahneman et al., 1982). As mentioned in §2.2 of ch.I, insights in neuroscience regarding distributed neural processing and neural plasticity also pointed to the significance of experience for shaping dynamical cognitive structures (Fox & Friston, 2012; Cech & Martin, 2012).

Within these distinct strands, Cowart (2019) points out some commonly shared theoretical assumptions. One is the primacy of goal-directed actions in real-time: the view that cognition is founded on an organism's capacity to act in its environment. That is, the view that an organism's development of basic motor-based and perceptual abilities is a crucial initial step towards developing the capacity for more complex cognitive processes, like using language. Hence, goal-directed actions are considered primary, and the development of higher cognitive capacities are considered to depend on the initial performance of these low-level actions (Cowart, 2019). Influential evidence for this position was offered by Thelen and Smith (1994) in their investigation of how infants learn to reach. They found that, although infants take a similar amount of time overcome this developmental obstacle (leading psychologists to traditionally consider the process innate), the kind of strategies they employed were rather unique given variations in their particular embodiment, such as their body mass index, energy level, and different initial approaches to reaching, etc. They concluded that the solutions found by individual infants were "discovered in relation to their own situations, carved out of their individual landscapes, and not prefigured by a synergy known ahead by the brain or the genes" (Thelen & Smith, 1994:260). Change, in their account, is explained in terms of a dynamical-systems framework, wherein the challenge is to understand how a given system can "generate its own change, through its own activity, and within its own continuing dynamics" (Thelen, 1995:91) by tracking how new behaviour is generated from the current resources, and "environmental scaffolding" of the system (Cowart, 2019).

Thelen and Smith's grounded, emergent conception of change contrasts with other leading developmental theories, in which change is typically described in terms of innate mechanisms such as genes, brain maturation, or 'shifting to a new stage'. Thelen (1995) contends that, given the unique problems faced by particular infants, it is implausible that the solutions were innately specified, as no internal mechanism could anticipate the specific, contingent parameters involved beforehand. According to Cowart (2019) such a real-time analysis of how behaviours evolve over time, as provided by a dynamic systems approach, is completely overlooked by classicist/cognitivist accounts of developmental change.

Another common theoretical assumption in embodied cognition is the conviction that the particular kind of embodiment of an organism (e.g. whether it has eyes, wings, hands, a tail, etc.) determines (and limits) the kinds of cognitive processing it is capable of (Cowart, 2019). That is, an organism's particular embodiment determines the way in which it executes goal-directed actions in the world, and the specific sensorimotor experiences to which such actions correspond help determine its category and concept formation (Cowart, 2019). An influential paper by Nagel (1974), *What is it like to be a bat?*, tries to illustrate the deep differences between the worlds of subjective experiences of different (even closely related) organisms, in this case between humans and bats:

> Bats, although more closely related to us than those other species, nevertheless present a range of activity and a sensory apparatus so different from ours that the problem I want to pose is exceptionally vivid (though it certainly could be raised with others species)…Now we know that most bats…perceive the external world primarily by sonar, or echolocation, detecting the reflections, from objects within range, of their own rapid, subtly modulated, high-frequency shrieks. Their brains are designed to correlate the outgoing impulses with the subsequent echoes, and the information thus acquired enables bats to make precise discriminations of distance, size, shape, motion, and texture comparable to those we make by vision. But bat sonar, though clearly a form of perception, is not similar in its operation to any sense that we possess, and there is no reason to suppose that it is subjectively like anything we can experience *or imagine* (Nagel, 1974:438, own emphasis)

That is, Nagel argues that, although bats are mammals (like us) that perceive the same external world, the differences between our means of perception (and arguably other physical differences, like our differences in size, shape, etc.), is so profound that our imagination (which is structured in terms of our own kind of embodied experiences) cannot even fathom what it is really like 'to be' (to inhibit the phenomenal world of) such a creature[54].

---

[54] This is despite the fact that some people have learnt how to similarly detect objects in their environment ('echolocate') by actively creating sounds (e.g. tapping their canes, snapping their fingers, making clicking noises with their mouths) and listening for their echoes (see Kolarik et al., 2014, for a summary).

To further illustrate the effects of physical differences, Cowart (2019) uses the example of a child and a dog playing with a ball. Whilst the dog is likely to use its mouth to interact with the ball, the child would rather use her hands or feet than her mouth, given not only social constraints, but also the kind of options that their particular kinds of embodiment make more readily available. In either case, every kind of interaction (kicking, grabbing with the hand, clutching in the mouth, etc.) also has its own set of associated sensorimotor experiences that directly influence their means of interacting with the object (Cowart, 2019). Likewise, the particular experience of another animal of the same event may differ substantially from that of a child or dog—even different children may have disparities in their experiences based on different senses (loss of sight or colour-blindness), abilities (hand-eye coordination, disabilities), etc., which might encourage them to approach tasks differently.

Hence, embodiment theorists maintain that the particular way in which an organism is embodied not only constrains the way in which it interacts with the world, but also partly determines how the world appears to it. This is not to deny the existence of an observer-independent world, but merely that the world does not present itself to all organisms in exactly the same way. Rather, it is partly actively constructed, depending on a number of dynamic, directly relevant factors, such as its vantage point, its form of embodiment, the task at hand, etc. (Cowan, 2019). Consequently—even to the same organism—a given environmental space is represented differently depending on the task being performed, as different environmental features are relevant to different goal-directed activities. For example, different aspects of common household objects might have more or less salience to different goal-directed agents, such as children playing a fantasy game (looking for parts of their spaceship); or someone cleaning the house (looking for dirt on the objects), or to someone when they hear a burglar and perceive anything as a potential weapon.

This view of active construction in goal-directed action contrasts directly with the cognitivist/classicist assumption, i.e. that our surroundings exhibit sets of pre-given features that our minds passively receive as sensory data and model as representations that directly reflect the state of the world. Instead, theorists in embodied cognition contend it is unnecessary and inefficient for an organism to form representations that fully represent all the features in its environment (even as mediated through their particular perceptual mechanisms) when trying to perform a goal-directed activity efficiently (Cowart, 2019). This idea is expanded on by the *enactive* view, discussed below.

The idea that at least some forms of cognition are constructive is backed up by an increasing amount of research output in a variety of disciplines (e.g. Valera et al., 1993; Damasio, 1994; Glenberg, 1997; Lakoff and Johnson, 1999; Fauconnier & Turner, 2002). On this approach, there is no proper or 'correct' way to view the world, as that merely amounts to using whichever sensorimotor/perceptive mechanisms have developed through evolution to help a particular organism act successfully in its environmental niche: "The point is that an organism's knowledge of the world is primarily through its experiences within the world and these experiences are constrained by the types of functioning sensorimotor modalities it as" (Cowart, 2019). Moreover, when one of its modalities is compromised, its experience of the world may be affected on multiple levels, as these modalities are considered to mutually influence one another. This is illustrated by Valera et al. (1991:164-165) who use the example of a painter who became colour-blind due to an accident. His loss of the ability to see colour seemed to affect the way in which he experienced other modalities (e.g. sound or taste), to the extent that his enjoyment of partaking in various activities suddenly decreased.

Embodied theorists typically support this sort of view of multimodal processing: that the brain uses resources from various modalities to execute tasks like interpreting language. That is, the 'semantic substrate' of concepts is directly grounded in, and emerges from, the kinds of modalities represented by the concept (e.g. Damasio, 1994; Barsalou, 1999; Lakoff & Johnson, 1999; Zwaan, 2004). As a neo-empirical enterprise, the embodied account assumes that concepts emerge from perceptual experiences: when we perceive and interact with dogs, for instance, this leads us to extract certain functional and perceptual attributes of dogs, which are stored in an *analogue* (graded) manner. When we then imagine (or talk about) dogs, this involves reactivating (or *simulating*) the different perceptual an interoceptive experiences we came to associate with the concept DOG (Evans, 2019:209). The multimodal view of conceptual structure is further encouraged by research on synaesthesia[55], which, some have suggested, is not as rare as previously thought: "Everybody potentially starts off as a synaesthete …. As the brain develops or responds to the input it receives, we lose this kind of connectivity and we lose these superfluous connections between neurons, and different areas of our brains become specialised to respond to different stimuli" (Duncan Carmichael, as cited by Bilby, 2015). Again, this directly contrasts with the classical modular view of the mind that assumes different modalities are processed separately.

As an expansion of the embodied view, the embedded view of cognition understands human cognitive processes as influenced by aspects of the natural and sociomaterial environment that enable, shape, and constrain the range of possible behavioural and psychological responses an individual can make in a given society[56] (Overmann & Malafouris. 2018:3). According to (Fenici, 2012:279), this view encapsulates two close principles: 'situatedness' and 'embeddedness'. On the one hand, cognitive *situatedness* claims that the body, and hence cognitive activity, cannot be artificially separated from the environment in which it is situated—emphasising the central role of background information to the processing of any stimulus (Fenici, 2012:279). In general, environmental factors are considered crucial as they can determine not only the set of available options for a particular organism, but also its preference among those options when engaged in a given goal-directed activity (Cowart, 2019). For instance, the way in which one goes about playing a game outside may be influenced by factors such as the rules of the game, the weather, the number of players, etc. Moreover, one's past experiences with similar activities and objects may also contribute to one's current understanding of the activity (Cowart, 2012).

---

[55] That is, a neurological phenomenon where one sensory modality crosses over into another, e.g. to 'see sounds', 'taste colours', etc. Abstract concepts may also evoke sensory experiences, like numbers being associated with different colours.

[56] Franz Boas conducted early work on embedded cognition in anthropology, exploring how different environments affect mental capacities like how aspects (like colour) or entities in the world are categorised and expressed in language (Overmann & Malafouris, 2018:3).

On the other hand, *embeddedness* emphasises the role of external structures in supporting and scaffolding cognitive activity (Fenici, 2012:279). Language, for instance, can be seen as a useful tool for alleviating the computational complexity of a task; for instance, writing down a shopping list to help one remember the items, or rehearsing the words in one's head as a rhyme. However, although language is learnt in interaction with one's (social) environment, it then becomes internalised and can be used irrespective of where the cogniser is (immediately) situated (Clark & Karmiloff-Smith, 1993; Clark, 1998). Hence, not all forms of embedded cognition are necessarily considered situated, and *vice versa* (see Fenici, 2012:280).

The stronger view of cognition as *extended* takes the resources and processes of the physical world, including the body, as central components of cognition. It sees the mind as externalising many of its functions by incorporating and recruiting extracranial tools[57] and processes—'extending' cognitive processes into them. General motivations for this approach include observations that (i) a complex combination of neural and additional resources cooperate, often in ad hoc ways, suited to a specific context; and (ii) neural processes are scaffolded by extracranial structures (which may include social scaffolding like social institutions and interpersonal interactions) to such an extent that the former cannot perform the same feats without them (Rupert, 2019). Moreover, if one accepts that the particular physical constitution of the body is implicated in cognition (by defining the way in which we experience/perceive the world), then it seems likely that artificial devices that modify the constitution of our bodies—anything from improving/modifying our senses, to adding new ones[58]—would likewise affect our cognitive processing.

A classic example of referred sensations is the blind man's stick (e.g. Lotze, 1888; Merleau-Ponty, 1962), which many consider to function as an integral part of the cogniser's perceptual system (and thus cognition), as extends the user's sense of touch to its tip. Some maintain that the blind man primarily (directly) feels the resistance of the floor rather than the resistance of the stick in his hand (Martin, 1993; O'Shaughnessy, 2003; De Vignemont, 2018). In a related seminal study, Iriki et al. (1996) found that using a rake to reach food affected the neural responses in monkeys: neurons that displayed no visual response to food at the far location began to display visual responses whilst the tool was used. This seemed to suggest that the use of the tool allowed things that were perceived as far from the body to be perceived as close, i.e. by extending their *peripersonal space*[59] (De Vignemont, 2018). Hence, some contend that, conceptually, there is no principled way to distinguish between our own bodies, bodily extensions such as artificial limbs, and some other tools that we wield at will—these are seen as extending our motor, sensory, and spatial abilities (e.g. Beck 1980; De Vignemont, 2018).

---

[57] Tool, here, is meant specifically in the sense of any unattached external objects that one actively manipulates for functional purposes (Beck, 1980; De Vignemont, 2018)

[58] For instance, there have been some experiments with 'cyborg antennae' integrated into the skulls of blind people to help them get a sense of colour (see Ramoğlu, 2019, for a review of such modifications).

[59] That is, the space within which the body can act (Maravita et al. 2003:531).

In a foundational work on extended cognition, Clark and Chalmers (1998) propose an *active externalism*, emphasising the active role that an organism's environment plays in its cognitive functioning. To illustrate, they evaluate various cases of problem-solving in which human reasoners tend to rely heavily on environmental supports: e.g. the use of a pencil and paper to carry out long division, the reconfiguration of Scrabble tiles to stimulate word recall, and the general use of diagrams, language, books, and other cultural resources. In such cases, Clark and Chalmers argue that the subject and the external entity form a *coupled* cognitive system in its own right, with each component playing an active causal role in governing behaviour, just as internal cognition normally does (Clark & Chalmers, 1998:8). Considering the previous examples of using language to support cognition (Clark & Chalmers, 1998:12) argue that there is no principled distinction between human memory and using external cues as a means to remember things: for a person with Alzheimer's carrying around a notebook to write down and look up things they need to remember, the notebook plays the same functional role as biological memory would, which they consider sufficient for it to count as a part of cognition. The distinction is blurred further given that some people are able to recall image-like representations of texts (Clark, 2005), or think 'in a language' to scaffold thought (Menary, 2010)—just as the process of writing may 'help' one think.

On this matter, however, there is much controversy. Extended cognition theorists have been criticised of the so-called coupling/constitution fallacy: that the strong coupling between neural and non-neural processes (i.e. their causal or enabling role in thought) is not sufficient to make those non-neural processes constituents of thought (e.g. Adams & Aizawa, 2008; Rupert, 2009). Nevertheless, it does give compelling examples to further support the claim that extracranial processes, at least, have some role in cognition and hence the mind cannot be treated as wholly implementation-neutral.

## 2.2. Enactivism and affordances

Enactivism is perhaps the most complex of the 4Es; it entails that cognition is both embodied and embedded, as well as *affective*; that is, made meaningful by the goal-directed perspective of the organism. Whether it also entails that cognition is 'extended' remains a matter of debate (Colombetti, 2018:1). As seen above, the view of embodied cognition generally takes a middle-ground between what Valera, Thompson and Rosch (1991) call the *chicken* and *egg* positions:

> *Chicken position*: The world out there has pregiven properties. These exist prior to the image that is cast on the cognitive system, whose task is to recover them appropriately (Valera et al., 1991: 172).

> *Egg position*: The cognitive system projects its own world, and the apparent reality of this world is merely a reflection of internal laws of the system (Valera et al., 1991:172).

Instead, the embodied understanding is that the world and perceiver mutually specify each other: categories, such as colours, are experiential, informed by our perceptual and cognitive capacities, yet they belong to our shared biological and cultural world (Valera et al., 1991:172). In other words, they are universal to the extent that they reveal how our shared biology as a species enable us to perceive the external (and our own internal) world in sufficiently similar ways, as mediated through a shared physical and socio-cultural context.

The particular take on this view by enactivism, is that cognition—rather than a mere recovery or projection— is "embodied, embedded action": *embodied*, in the sense that it is determined by a particular body with certain sensorimotor capacities and *embedded*, in the sense that these various sensorimotor capacities are themselves embedded in a larger overarching biological, psychological, and cultural context (Valera et al., 1991:172). Moreover, *action* emphasises that sensory (perception) and motor (action) processes (including eye movement) are considered inseparable in the lived cognition of an organism. The basic approach is described by Valera et al. (1991) who popularised the term 'enactive' in cognitive science:

> We propose as a name the term *enactive* to emphasise the growing conviction that cognition is not the representation of a given world by a pregiven mind but is rather the enactment of a world and a mind on the basis of a history of the variety of action that a being in the world performs (Valera et al., 1991:9).

Their formulation involves two central ideas. The first is that perception consists in perceptually guided action: the world we perceive is 'enacted' in the sense that our perception and action are interdependent processes (Gangopadhyay & Kiverstein, 2009:64; Colombetti, 2018, 1). Secondly, cognitive structures emerge from the organism's perceptually-guided interaction with its environment (Valera et al., 1991:173). These interactions bear on the particular way in which the organism is embodied, as its particular embodiment determines which properties in the environment are perceived as meaningful/significant to it (Ward et al, 2017:368). Thus cognition, in an enactivist sense, is at least partly specified by the cogniser's own activity, interests, and capacities, as well as the entities, threats and opportunities in its immediate environment.

A primary motivation for the move towards enactivism is the conviction that classical computational theories fail to recognise the role of agency in perception—the importance of goal-directed behaviour, as discussed earlier. Classicists typically describe the visual system as functioning almost like a camera taking snapshots of a pregiven, observer-independent world, which are then saved in the form of detailed internal representations (Gangopadhyay & Kiverstein, 2009:64). For enactivists, in contrast, the reference point for understanding perception is to understand how the sensorimotor structure of the perceiver (i.e. how the sensory and motor surfaces are coupled) determines the interplay between its action and perception in a local environment. That is, how the organism's perception changes as a result of its perceptually-guided activity (Valera et al., 1991:173). Thus, enactive theories of perception emphasise that experience must be understood as it unfolds in *embodied action* embedded in, and constrained by, its surrounding world (inspired by work in the phenomenological tradition[60]) (Gangopadhyay & Kiverstein, 2009:63).

Another formulation of enactivism focuses on the ability of a cognitive system to infer systematic interrelations between current (*actual*) and future (*possible*) perceptions, sensations and actions (e.g. Hurley, 1998; Noë, 2004). One example is Noë's (2004) sensorimotor theory, through which he aims to account for our ability to perceive objects as whole and three-dimensional, given that, at any point in time, we are only able to perceive narrow subjective perspectives/selected aspects of them. For example, when we see a bed from the front, we normally understand that it is likely not just a cardboard cut-out of a bed and more of the object extends beyond it. According to Noë (2004), this is enabled by our ability to grasp how our activity can bring different aspects of entities in our field of vision into view. Thus, our perceptual experience is dependent on grasping how *what we can do* affects *what we see*.

One of the foundations of enactivism, and for embodied theories of cognition in general, was James Gibson's (1979) ecological psychology. For Gibson, perception is not the passive reception and representation of environmental information. Rather, perception is seen as *active* and *direct* in at least two ways: firstly, our perception of the environment is modulated by our own exploratory activity (e.g. moving around, squinting, scanning, etc.) and unfolds over time (e.g. we generally move our eyes to parts of our surroundings that are most relevant for our current task) (Ward et al., 2017:366). Secondly, not only can we perceive the layout of things before us in space, but we can also *directly* perceive the possibilities for interaction that they afford us (Gangopadhyay & Kiverstein, 2009:64). Gibson (1979) termed these *affordances*: perceived opportunities to engage with things in the world, in ways that reflect our capacities and purposes. For instance, we can see a chair affords sitting, a roof affords shelter from rain, a ladder affords climbing, a button affords pushing, etc. The characteristics of an object that make it look both positively (beneficially) or negatively (harmfully) *interact-with-able* in specific ways—for a given agent/species in a given environmental niche—together form its affordances. Hence, the information the observer uses to specify the utilities in her environment involves information about the observer herself—her body, legs, hands, and mouth. Thus, exteroception is accompanied by *proprioception*: "to perceive the world is to co-perceive oneself" (Gibson, 1979:141).

According to Gibson (1979), the direct perception of affordances and their affective allure is precisely what makes objects meaningful or significant to us: "Objects of perception are meaningful to us just to the extent that [the] perceiver has a grasp of the possibilities for action the object affords" (Gangopadhyay & Kiverstein, 2009:65). That is, in given contexts, relevant affordances move us as they get us ready to act (Rietveld et al., 2018:14). This idea is derived from Gestalt psychologists like Koffka (1935), who likewise acknowledged that the meaning/value of a thing seems to be perceived just as immediately as its physical properties. In *Principles of Gestalt psychology*, Koffka noted:

> Each thing says what it is… a fruit says 'Eat me'; water says 'Drink me'; thunder says 'Fear me'; and woman says 'Love me' (Koffka, 1935:7).

---

[60] The phenomenological work of Heidegger and Merleau-Ponty particularly served as great inspirations for enactivism. Merleau-Ponty (1962), for instance, acknowledges that the perceiver both initiates and is shaped by its environment, and should thus be seen as bound together in a reciprocal specification and selection (Valera et al., 1991:173).

These values have what Koffka calls a 'demand character' that is relative to the immediate needs of the observer. On Gibson's (1979) account, however, affordances differ in this crucial respect: although the meaning of an object may change with the changing needs of the observer, the affordances are considered potential possibilities for use, regardless of the perceiver's immediate needs. Hence, affordances are neither physical nor purely phenomenal, but an emergent quality in reference to an agent's abilities for engagement: "The object offers what it does because of what it is" (Gibson, 1979:139)—and because of what *we* are.

Researchers in enactivism have since expanded on the idea of affordances to emphasise a notion of context-sensitivity; for instance, in certain social/cultural contexts, a lifted hand can invite a *high-five*, or an extended hand can invite a handshake, whilst not in others (Rietveld et al., 2018:2). Moreover, the abilities that allow us to respond to affordances do not have to be pregiven, but can be acquired (Rietveld et al., 2018:2). For instance, a history of social practices has allowed us to learn to read, which enables us to grasp the affordance of readability. Affordances, and their embeddedness in sociocultural practices and a particular sociomaterial environment, feature prominently in recent work in enactivism (e.g. Colombetti, 2018; Rietveld et al., 2018).

According to the view of enactivism sketched thus far, one's cognition fundamentally involves grasping the impediments and affordances in one's environment, with respect to one's particular embodiment, aims, and sociocultural context. It thus follows that cognition is essentially *affective*, as it depends on the perceiver taking an "evaluative stance" towards objects in the world and their relation to her interests and embodied capacities (Ward & Stapleton, 2012:99). According to Gallager (2005), any account of embodied cognition focusing merely on the sensorimotor aspects of action and perception, without including affect, provides an incomplete story, as one needs to account for 'motivational pull' in one direction or another. Affect significantly contributes to our attentive outlook, even if we are not consciously aware of its effect in motivating our (perceptually-guided) actions—moods such as happiness, sadness, or anger modulates one's viewing or listening behaviour.

These affective states are also revealed in subtle behavioural expressions we might be unaware of (Colombetti, 2018). Moreover, this influence can go both ways: an ample body of research has centred on the way in which our body is positioned/facial expression we have, affects our mood (e.g. Soussignan, 2002; Briñol et al., 2009; Veenstra et al., 2017). For instance, studies have suggested that holding a pencil in one's mouth may trick the body into thinking that one is happy, as the face is positioned in the way it would if one were smiling (Soussignan, 2002). Likewise, assuming physical stances that exhibit confidence (like puffing out one's chest or lifting one's chin) has been found to raise self-confidence (Briñol et al., 2009). Such studies offer further evidence for the (at least causal) role of embodiment on cognition.

Based on these considerations, the final part of this section looks at the role of affective states in social perception/cognition; that is, how our ability to pick up aforementioned behavioural cues, as well as our capacity to feel emotions ourselves, may aid in our ability to successfully interpret the actions and intentions of those around us.

## 2.3. Embodied social cognition and affect

Concerning the role of emotion, the emerging research programme of embodied social cognition takes it as a central concern to understand the way in which our social behaviour is influenced by affect and empathy. This research programme falls within the 4E enterprise as an investigation of our ability to 'mindread'; that is, in the weak sense of our ability to (often successfully) interpret the actions and intentions of others. Traditional accounts of social cognition typically explain this in terms of the attribution of folk-psychological mental states (beliefs, desires, etc.) to others. This either relied on classical computational accounts of theory-of-mind acquisition using sentence-like LOT representations to describe the processing of theory-of-mind abilities (e.g. Baron-Cohen, 1995; Scholl & Leslie, 1999); or in terms of child-as-scientist accounts (e.g. Gopnik, 1996) which considered children's ability to form theories about others' folk-psychological states, often supposing the use of innate inferential rules and/or domain-specific social cognition modules. Focusing mainly on processes 'in the mind', neither of these approaches considered that embodied capacities may play a central role. More recent research, however, suggests that our knack for inferring the affective states of others can largely be attributed to our shared biological constitution and situatedness in a broader sociocultural context. In this subsection, I discuss various such theories and findings that oppose traditional body-independent approaches.

One embodied approach is simulation theory, which explains mindreading in terms of modelling the behaviour of others as a *motoric simulation* (see Wolpert et al., 2003). Rather than drawing on a separate body of stored information about human behaviour, the contention is that we 'reuse' our own action mechanisms 'offline', as a model for inferring information about the behaviour of those around us. As such, simulation theory is generally seen as coinciding with embodied cognition. It rejects the idea of a distinct innate, domain-specific mindreading module that traditional theories often postulate (e.g. Leslie, 2005). Rather, social cognition is considered to rely on our own systems for perception, action, belief-formation, decision-making, etc. Not to be confused with imagination, simulation is the rapid (and not necessarily conscious) attempt to see things from another's perspective (see Hurley, 2008, for an evaluation of different approaches).

What seems like compelling scientific evidence for a simulation approach is the discovery of 'mirror neurons' in macaque monkeys, and, subsequently, of a mirror system in the human brain (Rizzolatti & Craighero, 2004). These are neurons in the sensorimotor system that activate both when an animal is performing an action and when one observes another carrying out the same action (Jacob, 2008:190). It has been argued that the activity of such neurons underlies a variety of (social) cognitive abilities like mindreading (Gallese & Goldman, 1998), imitation learning (Rizzolatti et al., 2002), and even language comprehension[61] (Rizzolatti & Arbib, 1998).

Another embodied approach to social cognition, closely aligned with enactivism, is interaction theory. Whilst simulation theory focuses on the mental mechanisms *in a subject's head* when they engage in mindreading, interaction theory claims that social cognition is essentially *interactive*, and thus not limited to the mind of an interpreter (Gallagher, 2001; Fuchs & De Jaegher, 2009). Gallagher (2008, 2011) argues that interpreting the mental states of other people relies on *direct perceptual* (as opposed to inferential) capacities, as agents' intentions are explicitly expressed by their bodily actions. On his account, as we develop, we learn more about the kinds of actions that humans engage in, which affects both the way in which we perceive social behaviour and the social actions we ourselves engage in. Moreover, these actions are situated in social interactions, which employ low-level sensory-motor associations that are developed since early childhood. Thus, interactive (e.g. gaze coordination), embodied (e.g. sensorimotor processes), and environmental factors (e.g. social context) all need to be taken into account to understand social mindreading. Similarly, a related theory, *situationism*, emphasises the role of social and environmental contexts in guiding human behaviour. According to Barker (1968), a situation is often a better predictor of an individual's behaviour than any amount of past information about them—a popular experiment supporting this view is the *Stanford Prison Experiment*[62] (Haney et al., 1973). A more nuanced position is advocated by Fenici (2012) who claims that this influence is more gradual: whilst there is ample evidence that children's earliest forms of social cognition mainly rely on capacities to process goal-directed behaviour and motor intentions, these become more embedded in social and dialogical practices (present in the context in which they are situated) later in life.

Overall, this subsection illustrates another important factor in cognition (and, hence, in communication): that it is not merely an organism's own biology and environment that is relevant to understand its cognitive processes, but also its interaction with other agents and interpretation of their embodied states. As a whole, this section offered a theoretical foundation by which to explain some of the key aspects of conceptual structure that I argued for in the previous section: firstly, that concepts are not merely abstract mental calculations, but have to be understood in terms of the particular embodied (perceptual, sensorimotor, affective) capacities of the individual. Secondly, that different kinds of context (e.g. goals, moods, conversation types, cultural events, social and physical environments, interactive cues from other agents, etc.) are involved and play a role in how we make sense of things in the world. Thirdly, that these multimodal inputs are not clearly separable in terms of discrete inputs/elements (like images, sounds, emotions, etc.) but are experienced as a complex integration of different associated modalities, and as such, if the embodied view is accurate, our concepts will also reflect this complex, integrated structure.

---

[61] However, there remains some controversy regarding the causal efficacy and explanatory power of these neurons: whilst some argue that they make it possible for us to enter into the same mental states that we observe in another person (e.g. Goldman, 2006), or interpret their intentions (e.g. Gallese, 2001), others, like Jacob (2008), have questioned whether they are used to infer other's intentions at all, or perhaps merely to predict their actions.

[62] To investigate the psychological effects of perceived power, twenty-one participants were randomly assigned to either the role of guard or prisoner in a fake prison environment. Participants were found to quickly embrace their assigned roles, with some 'guards' enforcing authoritarian practices and even subjecting some 'prisoners' to psychological torture, whilst many 'prisoners' passively accepted such abuse and even harassed others who objected.

Having provided a broad overview of 4E paradigm and its various claims about the nature of (human) cognition, the next section explores specifically how our linguistic knowledge and communicative abilities emerge from such general (embodied, embedded, and/or enacted) cognitive processes. This is guided mainly by Evans' (2019) comprehensive overview of the *cognitive linguistics* movement, which aims to give a naturalised understanding of acquisition and use of concepts and language, on the basis of these and other recent scientific findings about human cognitive/psychological development.

## 3. Cognitive linguistics

Cognitive linguistics emerged in the 1970s as a broad research enterprise responding to formal and cognitivist approaches to language. It is firmly rooted in the development of (embodied) cognitive science, as well as earlier traditions like Gestalt psychology emphasising the particularities of embodied human experience (Evans, 2019:1). Early research in the field was spearheaded by the work of Ronald Langacker (2000), George Lakoff (1987) and Leonard Talmy (2000), and also drew from Eleanor Rosch's work on typicality effects, to form a broad research programme[63] that adopted a broadly empiricist and non-modular approach to the study of language and mind (Evans, 2019:2-3).

More than studying language for its own sake, cognitive linguistics is unique in its approach to describe language as reflective of more fundamental evolutionary patterns of human cognition; that is, species-specific characteristics that are revealed in the nature, structure and organisation of thought (Evans, 2019:5). According to Evans (2019:25) this constitutes one of the key commitments of the enterprise: to characterise linguistic principles that coincide with that which other scientific disciplines have revealed about our cognitive/perceptive functions (i.e. the *Cognitive Commitment*). As argued earlier, rather than consulting lists of facts about entities or situations in the world, it seems we tend to rely on a sort of intuitive body-based 'know-how' for interpreting utterances in context (Dreyfus, 1992:xxi). Based on the insights from embodied cognition above, cognitive linguists seek to explain our context-sensitive (i.e. pragmatic) interpretative abilities in terms of general facts about our cognitive/bodily functions, like how particular lexical constructions and contexts prompt the reactivation of associated stored sensorimotor states in the brain (see §3.2 of ch.II).

---

[63] Cognitive linguistics is often referred to as an 'enterprise' or 'movement', as it is not any specific theory, but an approach guided by a shared set of principles, perspectives and assumptions (Evans, 2019:2).

A second key commitment is to the articulation of general principles that hold for all aspects of our language use, i.e. the *Generalisation Commitment* (Evans, 2019:25). Whereas formal linguistics tends to take different components of language—such as syntax, semantics, pragmatics, phonology and phonetics—as distinct and incommensurable areas of linguistic study, based on unique structuring principles (and perhaps mind modules[64]), cognitive linguists typically consider this distinction more a matter of practicality as they reject the modular view of the mind (Evans, 2019:26). Instead, they consider all these components to share certain fundamental organisational properties of human cognition. One of these is the 'non-criterial' nature of categorisation: as mentioned earlier, cognitive psychologists found that the boundaries of human categories are usually fuzzy, and membership is typically more a matter of centrality than sharing a defining trait(s)[65]. Moreover, how we classify objects is also often determined by our contextual interaction with them. For instance, my perception of a cup as a bowl may depend on degrees of overlapping features, such as its width, handle size, etc, but also as my particular use of the object, e.g. eating soup out of it with a spoon or drinking tea from it. According to Evans (2019:27) these category structuring principles of fuzziness and family resemblance are exhibited in linguistic categories across the board, including phonology[66], syntax[67] and morphology[68].

Another fundamental cognitive structuring property is embodiment (and embedded/situatedness), following the research in 4E cognition discussed above. In the following subsection, I explore some cognitive linguistic theories regarding how various aspects of our particular 'embodiment' factor into our conceptual (and linguistic) structure.

## 3.1. Conceptual structure

As noted earlier, the folk understanding that the aim of language is to describes states of affairs obtaining in the world, rests on the assumption that our constructions of language merely reflect the external world as is. However, cognitive linguists agree with embodied cognition theorists that such an objectivist approach is fundamentally misguided, as our world of basic experiences are constrained by our species-specific evolutionary modes of perception,[69] and may even differ across individuals within the same species (see Cassanto, 2014). Thus, it follows that the human mind—and thus how we structure/comprehend language—cannot be understood in isolation from our particular embodiment (Evans, 2019:203-213). As argued, one reason for this is that the way in which we make sense of the world constrained by the ecological niche to which our particular perceptual and experiential mechanisms have adapted[70] (recall Nagel's example of *What is it like to be a bat?*). On this approach, language reflects our particular human construal of the world—our *projected reality* (Jackendoff, 1983).

---

[64] However, most formal linguists may remain indifferent as to the physical instantiation of thought in the mind.
[65] Again, recall Wittgenstein's (1953) notion of family resemblance.
[66] For instance, research presented by Jeager and Ohala (1984) suggests that the phonological category of *voiced sounds* behaves as a graded continuum rather than a binary feature as it is typically described in formal linguistics.
[67] For instance, lexical classes like 'noun' and 'verb' also constitute fuzzy categories—verbs exhibit common variable *verbish* behaviour, but there are no fixed set of grammatical criteria (see Evans, 2019:29).
[68] For instance, affixes like diminutives in Italian (*-ino, -etto, -ello*) do not have single meanings attached to them, but share a related set of meanings like a reduction in size, strength, quality, etc.

Given this view, and following from the Generalisation Commitment, a key assumption of cognitive linguistics is that semantic structure (i.e. the meanings encoded in language) reflects conceptual structure, which ultimately reflects our embodied experience. This section reviews evidence for this. To do so, I first elaborate on some important foundations of human embodied experience that are taken to inform our use of language.

### 3.1.1. Experience and concept acquisition

Given the Cognitive Commitment, a core concern for cognitive linguists is to investigate the relationship between our conceptual structure and our particular phenomenal world (Evans, 2019:203). In this subsection, building on the points on 4E cognition above, I discuss findings from perceptual psychology to investigate embodied foundations of human concept acquisition, focusing on the topics of perception, categorisation, and conceptual metaphor theory.

As noted above, cognitive linguistics typically rejects the view that our three-dimensional world of spatial experience exists objectively *as we perceive it*. Instead, Evans (2019:57) contends that our particular psychological representations of our surroundings are formed through a process of perception that consists of three stages: the first is *sensation*, which concerns the way in which external energy (e.g. light, heat or vibrations) are picked up by the human body's modal systems and converted into recognisable neural codes. In the second stage, *perception*, this multimodal sensory information is organised and converted into a perceptual object, i.e. percept. The third stage is *identification and recognition* (i.e. categorisation) whereby percepts are interpreted on the basis of previous experiences and conceptual knowledge. As such, previously formed concepts are employed in order to categorise new precepts (Evans, 2019:57).

According to Bregman (1990), the formation of coherent percepts is determined by a process of 'scene analysis', which consists of a combination of *bottom-up* and *top-down* processing of perceptual information—as well as higher-level cognitive abilities (such as those of categorisation, described below), which seem to be innate (Evans, 2019:59). Bottom-up processing refers to the processing and integration of perceptual details[71] that constitute, for example, object percepts, such as a dog. On the other hand, top-down processing uses global principles to integrate perceptual information into coherent percepts within a coherent perceptual scene (Tyler & Evans, 2003). Principles of this kind have already been proposed in the 19[th] century by aforementioned Gestalt psychologists, who sought to characterise fundamental human perceptual mechanisms that allow us to construct wholes (*gestalts*) out of incomplete perceptual information. These can broadly be summarised in terms of five principles, i.e. the *principle of proximity* (that we generally group visual elements that are closer together); *similarity* (that we tend to group entities that share visual characteristics); *closure* (that we tend to complete incomplete images[72]); *continuity* (that we tend to perceive objects as continuous rather than segmented[73]); and *smallness* (that we tend to perceive smaller entities before larger ones) (Evans, 2019:63-64)—see Koffka (1935). Moreover, according to Gestalt psychology, our perception of certain kinds of sophisticated distinctions in images (like that of matter and form, figure and ground, essential and accidental aspects), is due to certain species-specific neurological biases/perceptual mechanisms that structure (and thereby mediate) our experience (Dreyfus, 1992:19).

---

[69] For example, the human visual system is incapable of observing colours in the infrared spectrum, which other organisms—like rattlesnakes—can access (Peter & Hartline, 1982).
[70] This view has been termed experiential realism (or experientialism) by Lakoff and Johnson (1980).

The bases for such basic perceptual distinctions have also been explored by neuroscience. Neuroscientists have discovered that when information is converted into electrical signals by the human visual system, different types of visual stimuli travel along two separate pathways (Ungerleider & Mishkin, 1982). This corresponds to two different areas of visual processing: the primary (focal) visual system, which provides information regarding form recognition and object representation (based on details of the object); and the secondary (ambient) system, which provides information about where an object is located with respect to body-centred space. These amount to distinct 'what' and 'where' information regarding objects in our external environment (Evans, 2019:65). More recently, Milner and Goodale (2006) have shown that the ambient stream also provides functional information (affordances) that facilitates our 'readiness for action' to interact with objects in our environment. Hence, the focal stream provides perceptual information of entities in our environment, whilst the ambient system facilitates our ability to engage with them.

Apart from our perception of the external worlds (*exteroception*), other types of perception include *interoception*, whereby internal sensory systems provide information on the body's inner state (e.g. hunger, affect, pain, temperature, etc.) and *proprioception*, relating to a sense of the body's parts in relation to each other through sense receptors in muscles, tendons and joints (Evans, 2019:58-59). A fourth type that is sometimes distinguished, is *kinaesthesia*, i.e. the felt sense of movement, which derives from a multimodal integration of sensory information. According to Evans (2019:59), contemporary cognitive linguistics assumes that embodied experience arises from a combination of all these different kinds of perceptual experience, as constrained by the nature of our embodiment (which includes the size, shape and orientation of our bodies in space, as well as our perceptual and other biological mechanisms).

---

[71] For visual perception, for instance, objects may be identified though visual patterns like colour and light intensity.
[72] For instance, perceiving a square out of four unconnected corners.
[73] For instance, if a car is on the other side of a fence, we would perceive the car as if it continues behind the areas that are obstructed by the fence, rather than seeing it as independent segments.

Whilst percepts arise as a result of immediate (on-line) perceptual processes, concepts—on an embodied account—emerge as stored schematisations (as mental representations, weakly defined) that are constructed/updated through an ongoing process of abstracting commonalities from different particular percepts and can be recalled off-line (Evans, 2019:58). Moreover, whilst percepts primarily relate to (sensory) details that are directly perceivable, concepts incorporate a much wider variety of information types, including abstract/relational properties like the sort of entity, its functional purpose, and how it relates to other concepts (Evans, 2019:58). Thus, a concept is seen as a 'theory' for identifying/categorising a particular percept, and thereby for making it meaningful (see Mandler, 2004).

According to Valera et al. (1991), for all organisms, categorisation is among the most fundamental cognitive processes as it is thereby that "the uniqueness of each experience is transformed into the more limited set of learned, meaningful categories to which humans and other organisms respond" (1991:176). As a part of her work on typicality, Rosch (1977) argued that the basic-level construction[74] of categories for concrete objects is by no means arbitrary, but is determined by a combination of (i) biological factors (i.e. being used/interacted with by similar sensorimotor systems), (ii) being perceived by similar sense modalities (i.e. serving similar needs/desires), (iii) cultural factors (e.g. having culturally meaningful attributes and functions), and (iv) cognitive needs for informativeness and economy—interacting with an external world that is itself structured in a particular way.

A revolutionary line of experientialist[75] research on conceptual structure was inspired by Lakoff and Johnson's (1980) conceptual metaphor theory, one of the first theoretical frameworks in the cognitive linguistics programme (Evans, 2019:300). Drawing from a broad range of interdisciplinary research[76], Lakoff and Johnson found that the prominent linguistic/philosophical distinction between literal and figurative language is fundamentally flawed. Rather, they argue that language—and, in fact, our thought—is fundamentally metaphoric in nature: our conceptual structure (and hence, the ways in which we act and perceive, think and talk about the world) is organised by virtue of cross-domain mappings (i.e. supposed analogies) between different conceptual domains. According to this approach, the complexity of our conceptual representations should be understood in terms of the kinds of concepts we are capable of forming, as determined by our particular embodiment (Evans, 2019:202). Some of these mappings result from pre-conceptual embodied experiences[77], whilst others build on those experiences to construct more complex conceptual structures. For example, to understand abstract concepts like communication, we rely on the parallels that we can form between it and our understanding of (more easily comprehensible) physical processes, i.e. the transference of material from one container to another—treating words as containers of ideas (objects)[78]. Thus, our mental representation of even very abstract concepts might be grounded, albeit indirectly, in our knowledge of basic sensory and motor processes (Wilson, 2002:10). Lakoff and Johnson use the term *structural metaphors* to refer to such cases where the structure of one domain is transferred to another.

---

[74] That is, on some standard medium level of specificity; e.g. categorising something as a *chair*, rather than a leather sofa (more specific) or piece of furniture (more general) (Rosch, 1977).

[75] That is, views that take experience (of an external world) to be the principle basis of human knowledge.

[76] This includes Charles Fillmore's work on frame semantics; Wittgenstein, Rosch, and Joseph Goguen's respective work on language and categories; the Sapir-Whorf hypothesis, amongst others (Lakoff & Johnson, 1980:xi-xiii).

[77] These have also been called 'primary scenes (Grady, 1997) or grounding scenarios (Moore, 2000).

According to Lakoff and Johnson (1980:7) such metaphors proceed systematically: particular mappings between different conceptual domains lead to further analogies/metaphorical treatments in language and general behavioural dispositions when dealing with those domains. For instance, in the West, where labour tends to be associated with (quantified) time, and people are typically paid for each hours/months/year of work, the conceptualisation of time as a commodity has become prevalent (Lakoff & Johnson, 1980:7). In keeping with this conceptualisation, we also conceive/talk of time as 'money' or as a valuable resource. That is, these systematic mappings constitute basic ways in which we *think* about time, and hence, are reflected in many of the ways in which we speak about time—e.g. 'How did you spend your time?'; 'We're running out of time'; 'How much time do I have left?'; 'He's not worth your while'; 'I've invested a lot of time into this', etc., as well as our treatment of time (Lakoff & Johnson, 1980:8-9).

Whereas some of these 'metaphorical concepts' are culture-specific (e.g. *time as a commodity*), others are based in more basic human perceptual biases (although the particular choice in experiential basis may also be mediated by culture). According to Lakoff and Johnson (1980:14), we structure many of our concepts in terms of our understanding of our physical bodies as oriented in space (e.g. UP-DOWN, FRONT-BACK, CENTRAL-PERIPHERAL, IN-OUT, etc.). Rather than the structural metaphors (given above) which map the particular structure of one concept onto another, *orientational metaphors* organise a whole system of concepts with respect to another: i.e. in terms of our understanding of different spatial orientations. For instance, 'positive' concepts (e.g. HAPPINESS, GOODNESS, MORE, HIGH STATUS, POWER) are typically oriented as UP, whilst 'negative' concepts (e.g. SADNESS, BADNESS, LESS, HIGH STATUS, WEAKNESS) typically oriented as DOWN. For example, note expressions like 'He cheered me *up*', 'The news is *uplifting*', 'She has *high* standards', versus 'It's bringing me *down*', 'I won't *stoop to your level*', 'He has a *low* status'. According to Lakoff and Johnson (1980:14), these metaphorical orientations are not arbitrary, but are grounded in our common physical and cultural experiences. For example, the difference in orientation for HAPPINESS/HEALTH (UP) and SADNESS/ILLNESS (DOWN) corresponds to the differences between posture (i.e. standing upright or drooping) that correlates with positive or negative emotional states (or good versus ill health). Likewise, the difference between DOMINANCE/POWER (UP) and SUBMISSION/WEAKNESS (DOWN)—e.g. 'I have the *upper* hand'; 'He has power *over* them'; 'She ranks *above* me', etc.—has a physical basis in the fact that power often correlates with physical strength, and winner of a physical fight is usually on top (Lakoff & Johnson, 1980:15).

---

[78] For example, 'I couldn't *get* my idea *across*'; 'It's hard to *put it into* words'; 'I enjoyed the *content*'; 'What *gave* you that idea?', etc.

Another kind of experiential basis is used for the orientational metaphor for UNDERSTANDING, where UNKNOWN is typically treated as UP (e.g. 'It is up in the air') and KNOWN as DOWN (e.g. 'The matter is settled'). According to Lakoff and Johnson (1980:18), this is due to the metaphorical structuring of UNDERSTANDING as GRASPING—which we find hard to do when something is out of reach, as we cannot fly. They also provide examples for orientation metaphors that (more obviously) have a cultural basis, e.g. viewing RATIONALITY as UP and EMOTIONALITY as DOWN (for instance, 'He couldn't rise above his emotions/basic urges'), which they consider to be based in a particularly Western cultural bias that *places* culture *above* nature, so to speak (Lakoff & Johnson, 1980:18).

On the basis of various such examples, Lakoff and Johnson (1980:17) conclude that most of our fundamental concepts are structured in terms of one or more spatialisation metaphors, drawn from our (physical and social) experiences, which form a coherent system (in a given culture). However, they admit that their descriptions of the respective physical bases are mere speculation, and noticing/unpacking metaphorical concepts is difficult as "[i]n actuality, we feel that no metaphor can ever be comprehended or even adequately represented independently of its experiential basis" (Lakoff & Johnson, 1980:17). Yet, their examples serve to suggest that different orientation concepts, though relying on a similar notion of verticality or *up-ness*, may have markedly different sorts of experiential bases (e.g. in MORE IS UP versus HAPPY IS UP or RATIONAL IS UP), as verticality enters into our experience in different ways (Lakoff & Johnson, 1980:17). They maintain that the major spatial orientations (front-back, up-down, near-far, on-off, central-peripheral, etc.) seem to cut across all cultures, although the orientations that are used for different concepts largely vary[79] (Lakoff & Johnson, 1980:24). Although there are, objectively speaking, many other possible frameworks for structuring space, Lakoff and Johnson (1980:56-57) argue that our particular orientational concepts just so happen to take priority for us, as they are most relevant to our normal bodily functioning and emerge from our continual experience of objects (and our own movement) in space::

> Almost every movement we make involves a motor program that either changes our up-down orientation, maintains it, presupposes it, or takes it into account in some way. Our constant physical activity in the world, even when we sleep, makes an up-down orientation not merely relevant to our physical activity, but centrally relevant (Lakoff & Johnson, 1980:56).

Beyond spatial orientation, Lakoff and Johnson (1980:25) argue that our experience of physical entities and substances provide a further basis for structuring our concepts—forming so-called *ontological metaphors*. Understanding our experiences in terms of objects with clear boundaries help us pick out parts of our experience as if they were discrete and uniform, which helps us to categorise, group, quantify, and thereby reason and speak about them. For instance, physical phenomena (e.g. street corners, mountains); events (e.g. hand-waving, a party); times of year (e.g. Christmas, summer); ideas (e.g. strategies or intentional states); and emotions (e.g. happiness or love), are all instances where we conceptually impose boundaries on phenomena that are not by nature discrete or bounded. According to Lakoff and Johnson (1980:25), just as our basic experiences of spatial orientations inform our orientational metaphors, so do our experiences involving physical objects (particularly our own bodies) form the basis for a great variety of ontological metaphors. As in the case of orientational concepts, most ontological concept metaphors are not commonly recognised as being metaphorical, as not only are they so natural and pervasive that they are assumed to be direct, self-evident descriptions of phenomena, but it is hard for us to refer to them otherwise (Lakoff & Johnson, 1980:25). That is, it is hard to refer to unbounded regions like, say, 'the corner' without treating it as a coherent entity.

---

[79] For instance, not all cultures give equal priority to up-down orientation—some structure their concepts more on balance or centrality (Lakoff & Johnson, 1980:18).

According to Lakoff and Johnson (1980:29), *container metaphors* are a common kind of ontological metaphor, which is a projection of in-out orientation (possibly of that of our own bodies) onto other physical objects that are bounded by some kind of surface, line or plane. For instance, we may conceive anything on the one side of a fence is seen as being 'inside', and beyond it is 'outside'; or conceive ourselves as moving *out of* one room and *into* another. We may also impose such conceptions on our natural environment, like being *in* or *out of* the woods, Cape Town, the harbour, etc. To justify this move, Lakoff and Johnson (1980:29) argue that territoriality is one of our most basic human instincts, and as such we naturally tend to quantify or put (imaginary) boundaries around territories. Apart from land areas, we also tend to conceive of our visual field as a container, with all that we see as being inside of it; for instance, 'The train is *coming into* view'; 'I have it *in* sight', 'The tree is *in the way'*, etc. Likewise, events like a race can be conceived as containers: a race, for instance, is typically conceived of as a container with the participants as objects (e.g. 'Sally was *in* the race') (Lakoff and Johnson, 1980:31).

Apart from protecting our own body-based understanding of space and bounded-ness onto concepts, there are many ways in which we personify non-human entities to make them more comprehensible to us. According to Lakoff and Johnson (1980:33-34), different personification metaphors (as extensions of ontological metaphors) pick out different aspects/characteristics of people, such as our goals, actions, intentions, and motivations, to make sense of phenomena in familiar terms; for example, 'This fact *argues* against the standard position'; 'Inflation is *eating up* our profits', 'This book *explains* neuroscience really well', etc. Related to these examples is another kind of associative process, *conceptual metonymy*, which relies on an actual relation between two components within the same domain (rather than structural analogies between different domains) (Lakoff & Johnson, 1980:35). For example, when one speaks about a book as if it 'explains' something, one is actually referring to the author's words. Other examples include using the part to refer to a whole (e.g. 'Get *your butt* out of my office'); whole for the part (e.g. 'I filled *the car* with petrol');

producer for the product (e.g. 'This painting is *a Picasso*'); object for the user (e.g. '*The car in front of me* isn't paying attention'); and the place for the institution/event (e.g. '*The White House* denies the allegations').

Although metonymy is primarily a referential device (i.e. using one thing to refer to another), Lakoff and Johnson (1980:36) argue that, like metaphors, metonymic concepts also facilitate comprehension. For instance, in the case of the part-for-whole metonymy, the part that is picked out signifies which particular aspect of the whole is focused on: e.g. 'We need more *muscles* to work on this', versus 'We need more *heads* to work on this', the speaker reveals the different aspects of people that are required for each given task, i.e. physical strength and intelligence respectively. Likewise, in saying '*Nixon* bombed Hanoi'), the metonymy reveals who one considers responsible for the action. As metaphoric concepts, Lakoff and Johnson (1980:37) argue that metonymic concepts are also pervasive in our ordinary ways of thinking/talking about the world, and not arbitrary, isolated occurrences. Rather, there are general kinds of metonymic concepts (like those given above) that are also used systematically to organise thought, and may reveal certain cultural biases. For instance, using 'head' to refer to intelligence/thought, reveals a cultural bias that considers the head to be the centre of human intelligence/cognition, whereas referring to someone as 'a good soul' reveals a cultural tendency to posit a metaphysical soul as the locus of a person's personality/essence.

More than merely structuring concepts in terms of others, another effect of grounding concepts in sensorimotor experiences is that it helps constrain the kinds of concepts an organism is able to acquire. For example, it is likely that an infant would have a simple understanding of concepts of macroscopic objects, like trees or bushes, whilst not understanding/possessing concepts of microscopic objects, like bacteria, despite both being within the same environment. This is unsurprising, as macroscopic objects are more readily perceived through her direct sensorimotor experiences—although she may likely learn about, or even perceive microscopic objects (with the aid of artefacts like a microscope or television) later in life (Cowart, 2019).

As such accounts suggest, the embodied picture of concept acquisition takes concepts to be (i) enabled and constrained by the kinds of categories our particular embodiment (and embeddedness/situatedness) allows us to form, and (ii) actively constructed (and not merely apprehended) through an organism's ongoing interaction with its (physical and social) environment. Apart from determining the kinds of concepts we are most likely to form, our fundamental sensorimotor experiences also facilitate concept formation by helping us understand inanimate/abstract concepts in terms of our basic understanding of our familiar bodily characteristics. Thus, if we were embodied quite differently, we would see the world in a different way: in terms of another set of defining bodily characteristics that enable the formation of new spatial schemas that we can project on a scene to facilitate our understanding (Cowart, 2019).

In this subsection, I evaluated some (innate and cultural) foundations of human perception and cognitive abilities that allow us to build up concepts through experience. The next subsection expands on this by evaluating how aspects of our human-specific embodiment are (further) evidenced in linguistic structure.

### 3.1.2. Embodiment effects in language

Earlier, I mentioned some of the species-specific perceptual biases we use to coherently process perceptual information, such as the Gestalt principles. Naturally, these basic perceptual biases are also reflected in the ways in which we structure our language, as a means by which we communicate our experiences (Evans, 2019:64-65). Our basic-level perceptual distinction between figure and ground, or *what* and *where*, is, for instance, reflected in the distinction between the lexical classes *noun* (i.e. the 'what') and *preposition* (i.e. the 'where'), as well as the grammatical distinction between *subject* (i.e. the primary entity) and *object* (i.e. the reference entity) in scene analysis (see Landau & Jackendoff, 1993; Talmy, 2000). According to (Evans, 2019:65), the lexical system (based on *what*-information) is further enriched by words distinguishing different object parts, forms, and other attributes, based on other perceptual distinctions.

Moreover, language mirrors the order in which we observe different entities relative to each other, and where the visual focus is. In English, this occurs in our syntax (Evans, 2019:65). For instance, in 'The wheelbarrow is next to the house', the figure (*wheelbarrow*) is primary, and the ground (*the house*) is a reference object. Though not incomprehensible, it would be semantically odd if these were reversed: 'The house is next to the wheelbarrow'. Talmy (2000) considers this an effect of the fact that the figure is more focal for us as a perceptual element, as determined by a combination of a range of human visual criteria that he specifies[80]. One such criterion is the property of being of greater relevance, as determined by the contextual/immediate focus of the speaker. For example, one might say 'The stone is in the wheelbarrow' or 'Look at the dewdrop on the wheelbarrow', which likewise reflect figure-ground distinctions, but picks our different 'figures' based on the speaker's current focus.

Recent evidence from cognitive neuroscience and psychology also a host of embodiment effects in (our use and comprehension of) natural language. Although neural processing has been revealed to be largely distributed (as described in §2.2 of ch.I), the human brain is (loosely) divisible into different functional regions; for example, some parts of the cortex[81] especially process and store sensory (e.g. tactile, visual, and auditory) information; others process motor information (e.g. information related to body movement); and some sub-cortical structures process and store emotional experience. One important finding is that multimodal brain states in such functional regions are instantaneously activated when we use corresponding body-based language[82]—that is, they are in a *somatopic arrangement* (Evans, 2019:214). In measuring subjects' brain activity, researchers found that those multimodal brain regions that are active when we perform certain actions (e.g. using certain implements like hammers, swords or pencils) automatically and immediately activate when we use/interpret utterances that refer to such tools (Pulvermüller et al., 2005; Buccino et al., 2005). Similarly, specific regions of the brain that process certain types of information (e.g. visual information), activate when we interpret language associated with processes that occur in that region (e.g. visual details such as orientation and shape) (Martin & Chao, 2001; Zwaan & Yaxley, 2003). Martin & Chao, for instance, found that regions that process animal recognition, such as shape/colour detection, activate when we read or hear corresponding animal words. Language involving emotions (e.g. rage, excitement) have likewise been found to activate the relevant brain regions (Isenberg et al., 1999).

---

[80] Talmy (2000) proposes a list of linguistic criteria for 'grammatical' figure and ground description, which closely mirrors the perceptual criteria of Julesz (1981) and Biederman (1987). Some of the criteria for 'figure' objects include being smaller, more movable, being less immediately perceivable but more salient once perceived, etc.

[81] That is, the outer layer of the brain.

Apart from activating the relevant sensorimotor brain regions, language also primes sensorimotor activity. This is revealed in the fact that we engage in a range of subtle behaviours that correspond to the sensorimotor activities that are implied by the language we use or interpret (as mentioned in the previous section)—particularly through our eye and hand movement (e.g. Klatzky et al., 1989; Glenberg & Kacshak, 2002). For instance, when we read about someone shooting a bow and arrow, we automatically activate muscle systems that ready the hands for the kind of motions required for drawing and shooting arrows (Evans, 2019:214). Moreover, when an experiment requires an action that is at odds with the kind of movement primed for by the sentence, the subject's processing time is slower than when the action is in line with the primed action[83]. To explain such phenomena, psycholinguist, Rolf Zwaan, describes users of a language as *immersed experiencers*: that language provides the interpreter with a set of clues for constructing and experiential ("perception-action") simulation of the situation described (Zwaan, 2004:36). Such discoveries fit in well with an embodied view of cognition (and language acquisition), but not a disembodied view.

## 3.2. Semantic structure

In this section, I look at the ways in which cognitive linguists approach the study of semantic structure. These are led by two guiding principles: that semantic structure is encyclopaedic, and that it contributes to meaning construction through the facilitation of so-called *simulations* (Evans, 2019:351).

### 3.2.1. The encyclopaedic view

The traditional view of meaning, the so-called dictionary/classical view described earlier, took it for granted that linguistic units represent neatly packaged bundles of meaning, which are distinguishable from other non-essential (background) knowledge that a speaker associates with them. That is to say, my knowledge that a bachelor is an unmarried man, for instance, has a different nature than my historic/social associations with (stereotypical) bachelors (e.g. that they live alone, are young men, are womanisers, etc.). Likewise, my knowledge of what hammers are, is distinct from my experiential knowledge of how it feels to use a hammer, the settings in which they are typically used, the kinds of hammers I am most familiar with, etc., as none of those are necessary nor sufficient for understanding what *hammer* means. Cognitive linguists, in contrast, take the view that that meaning is *encyclopaedic*; that is, that words and other linguistic units serve as 'points of access' to large repositories of structured knowledge associated with a particular concept or conceptual domain[84] (see Langacker, 1987). Thus, there is no fundamental distinction between dictionary (semantic/lexical) knowledge and encyclopaedic (non-linguistic/'world') knowledge: the former is merely a subset of the latter (Evans, 2019:377).

---

[82] See Taylor and Zwaan (2009) for a review.
[83] For instance, Glenberg and Kaschak (2002) gave subjects buttons that were placed at certain orientations, the pressing of which would either mimic or be at odds with the action described (e.g. closing a drawer). Subjects were asked whether a given sentence was meaningful or not, and responded quicker when the action of pressing a button was in line (directionally) with the activity primed for by the sentence.

In §1 of ch.II, I sketched out some of the issues with the traditional view of concepts/word meaning: that our inference of what a word means very much depends on contextual matters, and cannot neatly be spelt out in terms of strict necessary or sufficient conditions. The encyclopaedic approach aims to account for this variation by viewing word meanings as drawing from a broad range of associations and more complex/sophisticated bodies of use-knowledge (Evans, 2019:352). This is not to suggest that words have no conventional (typical) meanings associated with them, but that a conventional meaning serves as a 'prompt' for constructing a meaning that seems most appropriate in the given context of utterance (Evans, 2019:353). For instance, the word *safe* has a range of possible (though related) senses, which are evident in uses such as 'The baby is *safe*'; 'The ladder is *safe*'; 'Your secret is *safe*', etc. The meaning here ranges between 'unlikely to come to harm', 'unlikely to cause harm', and 'secured in one place'. To find the correct interpretation in each context, we draw from our encyclopaedic knowledge relating to babies, ladders, and secrets, as well as our knowledge regarding what 'safe' means, from which we then *construct* a meaning that seems most appropriate (Evans, 2019:353). This also helps us to disambiguate words that have distinct and unrelated meanings in context, like homonyms such as *bank*, e.g. 'I have some money in the *bank*' or 'My boat has not left the *bank*'.

This likewise applies to the figurative use of some words. For instance, in the example of *bachelor,* my encyclopaedic knowledge includes my knowledge that that certain types of unmarried men would not conventionally be considered bachelors, as well as my knowledge of cultural stereotypes associated with the word. Therefore, I can understand utterances such as 'Be careful Alice, your husband is a right *bachelor*!' despite the apparent contradiction, as the appropriate sense draws from my cultural knowledge of bachelors as womanisers, which the word also prompts for (Evans, 2019:352).

---

[84] This idea is somewhat similar to Quine's (1976) semantic holism.

To further illustrate the effects of encyclopaedic knowledge, Evans (2019:383) points to the distinction between a *bucket* and a *pail*: whilst these words are considered synonymous, and thus have the same denotations (the set of entities that they refer to, i.e. roughly cylindrical open containers with handles), their connotations (the associations evoked by the word) diverge (Evans, 2019:383). Where *pail* is typically used to describe containers of a certain size and material (wooden or metal) used to carry liquids like milk or water, *bucket* is applied more generally to containers of all sizes and materials to carry just about anything. Hence, it may sound a bit odd to say 'I built a sandcastle with a pail and a spade', albeit not technically wrong. Rather than being differences in 'core' meaning, these are merely differences in (dynamic) colloquial uses of the words and the associations that a given community of speakers form. Therefore, cognitive linguists argue that the decision of which information to include and exclude from the core meaning of a word is largely arbitrary, and is determined by the interrelation of a host of dynamic, contextual factors (Evans, 2019:384)— some of which I discuss below.

According to Evans (2019:387), rather than an unstructured mess of information, the network of encyclopaedic knowledge associated with a word has some form of a hierarchal structure, as different aspects have varying degrees of centrality (or, in a connectionist network, this hierarchy may be represented as the differing strengths of the respective weights and biases of connections/units). Langacker (1987) distinguishes four types (or dimensions) of knowledge that make up this network; i.e. *conventional, generic*, *intrinsic*, and *characteristic* knowledge—each of which contributes to the relative centrality/strength of particular aspects.

*Conventional knowledge* concerns the extent to which a particular aspect of cultural knowledge/meaning is agreed upon within a linguistic community; for instance, knowing that cereal is conventionally consumed at breakfast. *Generic knowledge* concerns the degree of generality of a certain facet of knowledge of a particular category; for instance, it is general knowledge that Einstein developed the theory of relativity. *Intrinsic knowledge* is that facet of meaning that does not refer to external influence; for instance, knowing that mammals have brains (which may be conventional but need not be). Finally, *characteristic knowledge* concerns the degree to which aspects of the encyclopaedic information are typical or peculiar to a particular class of entities. For example, female marsupials characteristically have pouches, whilst males and non-marsupials tend not to—though there are exceptions, like male seahorses. As illustrated, whilst, in principle, these kinds of knowledge are distinct, they often overlap. Moreover, they operate on continua rather than being matters of binary classification (Evans, 2019:387). According to (Evans, 2019:390), the extent to which these kinds of knowledge overlap, and/or how strongly they each meet their criterion—including other factors like how salient the aspect of knowledge is in memory—together determine how central that information is to the meaning of the lexical expression.

Apart from these general determining factors, as I argued in §1.5 of ch.II, meaning is also influenced by the context in which the expression is used, as different aspects of knowledge are selected based on the kinds of contexts in which different aspects are embedded (Evans, 2019:391). This phenomenon, which Cruse (1986) calls *contextual modulation*, occurs when a specific facet of the encyclopaedic knowledge associated with a term is privileged due to the discourse context, which leads to the on-line construction of an appropriate content. This entails considering a combination of different kinds of context, such as sentential context, (physical) situational context, interpersonal context (i.e. the relationship between the interlocutors), prosodic context (i.e. the intonation pattern of the utterance), etc. (Evans, 2019:391-392).

From this perspective, meaning is necessarily domain or *frame*[85]-dependent, and as such, words (and 'concepts') do not typically have a fully-specified, coherent meaning (Fillmore, 1982; Langacker, 2000). These 'frames' relate aspects of an individual's knowledge with a specific culturally embedded scene, event or situation from their experience. Importantly, frames are understood as schematisations of concepts—that is, we have a sense of how different features or attributes are (causally) related to each other[86], rather than merely knowing them as an ordered list (as assumed in the prototype theory earlier) (see Fillmore, 1982; Barsalou, 1992). According to Fillmore (1982) and Langacker (2000), lexical expressions cannot be grasped independently from the frame with which they are associated, and as such language is situated (contextualised) by definition (this follows from the usage-based approach that cognitive linguists typically take to language, described in §3.4 of ch.II).

Whilst the central meaning associated with a lexical item is comparatively stable, the particular associated network of encyclopaedic knowledge that each linguistic form gives access to is dynamic, updating continually on the basis of new experiences, further acquisition of knowledge, etc. (Evans, 2019:392). For instance, one may only learn that whales are mammals by reading about it, which might affect one's former perception of what *mammals* are, albeit not radically. Furthermore, new scientific discoveries might lead to the invention of new classes of (observable or theoretical) entities, like photons. From experience, one might also learn what certain emotive words like *depression* or *anxiety*, or even *love*, refer to, (arguably) better than one would by just reading a definition—see Jackson's (1982) *Mary's room* thought experiment.

This grounded, experience-based approach goes against a number of assumptions in formal approaches to semantics. One of these is the semantic/pragmatic distinction: where *semantics* may take a definitional or perhaps compositional approach (like that of Fodor), that takes word meaning as a coherent, context-independent core consisting of necessary and/or sufficient semantic features/concepts (e.g. MALE, ADULT, UNMARRIED), all the questions concerning how linguistic meaning interacts with the external world and how it is used in context would fall within the domain of *pragmatics* (Evans, 2019:378). Where proponents of formal semantics may maintain that a word's meaning determines how it is used, cognitive linguists take the opposite (Wittgensteinian) approach: the use, and consequent real-world associations, determines the meaning (Evans, 2019:385).

---

[85] That is, a *schematisation* (i.e. knowledge structure) of contextual experience, represented at the conceptual level and kept in long-term memory (Fillmore, 1982).

[86] For instance, we know the ability of flight in birds is related to factors such as their size, shape, wingspan, etc., as well as the fact that they are often seen perched on high tree branches or telephone wires.

The encyclopaedic approach gives a partial picture of how meaning is grounded in individual experience. Another aspect of it relates to how meaning is *embodied*. This is accounted for by the second principle, which is that semantic structure facilitates the construction of body-based *simulations*, discussed below.

### 3.2.2. Meaning and simulations

As discussed earlier, language has been found to activate certain multimodal brain states and prompt subtle motor behaviours. Barsalou (1999) integrates such findings into a theory of meaning: that (contextual) utterances elicit certain *perceptual symbols*, in a process he calls *simulation*. These perceptual symbols are understood as memories of multimodal (interoceptive and exteroceptive) perceptual states, stored in sensorimotor areas of the brain (Evans, 2019:355). To illustrate, if someone asks you to imagine using a pencil, Barsalou (1999:582) argues that we draw upon all of our discrete embodied/past perceptual experiences of pencils—their weight, colour, texture, shape, etc—which consist as distinct perceptual symbols. The (contextual) integration of such perceptual symbols forms a complex representational structure for a given utterance, which he calls a *simulator*.

According to Barsalou (1999), such simulators allow for the extraction of two different types of information. The first is a *frame*, understood here as a relatively stable schematic concept (representation) that abstracts across a range of different encounters (as perceptual symbols). The second is a *simulation*, a (subdued) 'enactment' of a sequence of perceptual experiences. Hence, one's knowledge of the concept *pencil* may be partially constituted by a schematic frame associated with the kinds of contextual perceptual information one associates with pencils, as well as reactivations (simulations) of one's embodied experiences with pencils (Evans, 2019:355).

The simulation view also gives a compelling account of how ideas are communicated: cognitive linguists typically maintain that language offers a framework enabling that facilitates the construction of these simulations—reactivations of perceptual symbols—in the minds of others (e.g. Zwaan, 2004; Barsalou, 2005; Bergen, 2012). Moreover, lexical units and grammatical constructions cue the reactivations of particular body-based (not exclusively sensorimotor) states in the brain in specified combinations, which give rise to complex simulations (which can account for the compositionality of language) (Evans, 2019:356). In fact, I would argue that this approach gives a more convincing explanation of the semantics of compositional statements than any other theory, as it offers a compelling explanation of how complex representations can be formed in context-specific ways. To illustrate, consider the different uses of the word *red* in the following examples: (i) 'The *red* fox is a sneaky animal', (ii) 'We have some *red* wine in the cupboard', (iii) 'Her face turned *red* when she saw him', (iv) 'I like the *red* jellybeans'. In each of these cases, a different hue of red is called to mind, activated on the bases on our past experiences of different types of red, as well as encyclopaedic knowledge of the different objects referred to—allowing for the construction of an appropriate meaning in context.

Again, this points to a limitation of Fodor's LOT account (as well as the prototype and exemplar theories) I mentioned earlier: if atomic concepts, e.g. *red* and *fox* are atomistic and got their contents causally from 'the right kinds of experiences', what one would end up with might be a simple combination of the most brilliant primary red and most typical fox. That is, such an approach fails to account for the appropriate emergent qualities that come in the particular combination of concepts, based on a contextual evaluation of our knowledge and experience with each; nor does it account for the imagination of unique and novel object-property combinations[87]. On this account, the conventional (prototypical) shade of red is used merely as a prompt for this process of meaning-construction in context—or offers access to its 'meaning potential' (Allwood, 2003). This dynamicity of meaning, based on an integration and weighting of different relevant factors in context, fits in much better with a connectionist approach to conceptual structure, as I described earlier.

Having described basic cognitive linguistic accounts of both semantic and conceptual structure, in the next subsection, I turn to the relationship between them: to explore how, from a cognitive linguistics point of view, conceptual and semantic structure maps onto each other.

### 3.3. Semantic versus conceptual structure

Whilst cognitive linguists (and other embodied cognition researchers) generally agree that language facilitates the construction of simulations, there are two different understandings of how this occurs—forming part of a greater debate in linguistics more generally. The one view is that semantic structure (i.e. the semantic content encoded by language) equals conceptual structure; that is to say, there is a direct link between the semantic content associated with linguistic forms and the concepts they elicit (Evans, 2019:358). On this understanding, the linguistic system has no inherent semantic content, but merely activates the aforementioned 'meaning potential' that resides entirely in the conceptual system (e.g. Glenberg & Robertson, 1999; Barsalou et al., 2008). One argument to support this view is that meaning and its expression in specific linguistic terms seem largely independent; for instance, in a study conducted by Bransford and Franks (1971), subjects were able to recount the gist of the narrative (i.e. the conceptual content) of a story they heard a few seconds prior using quite different words.

---

[87] For instance, forming a representation of, say, a robotic giraffe wearing trousers, cannot be a straightforward combination of the concepts, as not only would one have to decide which legs the trousers go on—evaluating how exactly the concepts should be combined based on one's knowledge and experience of both objects—but one would also have to use some subjective imagination to interpret what a robotic giraffe would look like.

The other view is that both semantic and conceptual structures have roles in meaning construction, but that the two are qualitatively distinct: the former serves as a guide for the activation of the latter, and it is this interaction that results in simulations (Evans, 2019:359). One argument for this more nuanced view, is that language seems to encode much more limited, specific types of semantic content that the kinds of rich, multimodal representations that reside in the conceptual system. Moreover, there are words in the linguistic system—particularly grammatical terms—that have no clear corresponding conceptual representation. For instance, even though we can understand the distinction between the definite article 'the' and the indefinite article 'a', they do not have conceptual representations that we can call to mind (like to distinguish between the concepts BIRD and FROG). Rather, their semantic distinction depends on the different functional roles they play in a specific linguistic system. According to Evans (2019:360), this seems to suggest that there is a level of semantic structure, indigenous to language, that is distinct from conceptual structure.

One might also argue that languages enable unique conceptual constructions, as, whilst it seems the same idea can be expressed in various languages (which seems to support the first view) there is rarely a one-to-one correspondence between languages, and sometimes there are simply no equivalent words by which a term in one language can be expressed in another. This ties into another argument by Evans (2019:360), to the effect that the way in which the language we use carves up the world appears to influence the way we see it. For instance, in Greek, there are different words for distinctions between lighter shades of blue (*ghalazio*) and darker shades (*ble*). Neuroscientists have shown that the brain activity of a native Greek speaker diverges when they perceive various shades of blue, where that of native English speakers (who typically categorise both merely as *blue*) exhibits no such divergence (Thierry et al., 2009). Many more such examples have been presented (e.g. Clifford et al., 2010; Liu et al., 2010; Boutonnet et al., 2012).

The fact that distinctions imposed by language seem to modulate non-linguistic perceptual categorisation—that semantic structure can influence conceptual structure—seems to suggest that the two are somewhat different (Evans, 2019:360). This points to a related argument, that language seems to induce the illusion of semantic unity (Evans, 2009). Considering the encyclopaedic view sketched above, what natural language seems to do is unify a range of distinct, albeit related, concepts by classifying it beneath a single term—whilst other languages may have distinct terms for those concepts. Hence, when a given word is used in one language, it may encapsulate/call to mind a broader landscape of conceptual meaning than a word in another language that can point to the same referent. For instance, in French, the word *mouton* refers both to a sheep (the animal), mutton (the food), and sheepskin, where England has different words for each meaning. Hence, *mouton* denotes more entities than *sheep* does, although they may sometimes pick out the same referent.

This is not to advocate linguistic relativity of the strong kind[88], i.e. that our language entirely determines (and limits) the way in which we categorise and perceive objects in the world (and, hence, that thought is impossible in the absence of language[89]), but rather a weak version that acknowledges an interaction between existing patterns in the environment, the kinds of bodies we have, and (dynamic) linguistic conventions that standardise the drawing of certain categorical boundaries (Evans, 2019:189) and use of certain conceptual metaphors (Lakoff & Johnson, 1980).

---

[88] This was inspired by Whorf's (1956) work on how language shapes ones world-view.

Having discussed the grounding of concepts; the effects of this embodied grounding in language; the nature of semantic structure; and how semantic and conceptual structure relates; the final part of this section concerns the way in which we acquire natural languages (and learn their grammars). For this, I discuss the prominent view in cognitive linguistics, i.e. the usage-based approach.

## 3.4. Usage-based theories of language acquisition

As noted above, one of the guiding principles of cognitive linguistics is that our knowledge of language emerges from (experience and context of) use. This is seen as playing a central role not only in how we acquire language, but also how language evolves over time. This empiricist/externalist approach contrasts with the internalist perspective of mentalists like Chomsky, who argued that language *competence* (knowledge) should be separated from its *performance* (use), prioritising the former (Evans, 2019:126). In rejecting this distinction, cognitive linguists contend that language knowledge is inferred from patterns of language use in the context of communication (Goldberg, 2006; Tomasello, 2008).

An important notion underlying this approach is the 'usage event' or (culturally and contextually embedded) utterance (Evans, 2019:128). According to Evans (2019:128), an utterance is *unit-like* in that it refers to the expression of a coherent idea. However, other than a sentence, it involves a combined range of linguistic[90] and non-linguistic strategies[91], which often makes it easy to omit some of the grammaticality requirements of a well-formed sentence; rather, utterances exhibit what he calls *graded grammaticality*. This is performed in context by a member of a particular linguistic community aimed at achieving a particular interactional (set of) goal(s) through making, at least partial, use of existing linguistic conventions (Evans, 2019:128). Thus, the interactional and goal-directed nature of language use, as well as its situated context—which interacts with the speaker's intentions and scaffolds interpretation—are all considered central to a usage-based account (Evans, 2019:130-131). This draws upon (but is not limited to) the conventional (coded) meaning of a particular word/construction, which, in cognitive linguistics, is understood as an idealised abstraction from a range of contextualised (pragmatic) uses. Moreover, its degree of relative frequency of use affects its level of entrenchment in the linguistic system (Evans, 2019:131). Hence, language use is often partly innovative—with new constructions/uses becoming more entrenched as they propagate through a given linguistic community (Evans, 2019:134). In this section, I discuss the use-based approach to language acquisition, which is one of the key approaches in the cognitive linguistics enterprise.

---

[89] Findings from ethology have demonstrated that many species that lack our kind of linguistic ability are, in fact, capable of cognitive processes like rudimentary categorisation (e.g. Hurford, 2007), and that pre-linguistic infants are able of conceptual processes like basic arithmetic operations (Wynn, 1992).

[90] Linguistic strategies include intonation, choices over whether or not to conform to discourse conventions (e.g. taking turns), choices over words and grammatical constructions, etc. (Evans, 2019:128)

[91] Non-linguistic strategies include *kinesics* (i.e. movement-based strategies like gestures, facial expressions, eye movements, bodily orientation, etc.), *haptics* (i.e. the use of touch), *proxemics* (i.e. space-based strategies like proximity), *vocalics* (i.e. the use of non-prosodic auditory features like laughter, sighing, etc.), and *chronemics* (i.e. the use of timing in exchanges) (Evans, 2019:128).

In opposition to innateness theories like that of Chomsky,[92] language acquisition is understood not as the activation of a system of innate linguistic knowledge, but as the derivation of linguistic units or constructions from patterns in the usage events that the user experiences.[93] This process does, however, rely on general (innate) cognitive mechanisms/abilities, and follows from the Cognitive and Generalisation Commitments described earlier. Langacker (2000) expresses this view in his *Cognitive Grammar* model, whereby he argues that language follows general principles that are common to other aspects of the human cognitive system. 'Grammar', here is not limited to the syntactic and/or morphological knowledge of language, but refers to the entire language system, incorporating meaning, morphosyntax, and sound (Evans, 2019:132). This rejects the modular view of language in favour of a *constructionist* view in which there is no fundamental distinction between lexicon and syntax; rather, grammar is understood as an inventory of units (symbolic assemblies) that are form-meaning pairings (words, morphemes, and grammatical constructions), which unify properties of meaning, grammar and sound in a singular representation (Evans, 2019:132-133). These are derived from language use through a process of abstraction and schematisation. Abstraction refers to the process by which structure emerges from generalising over patterns of instances. On this view, a speaker of a given natural language can, as the consequence of regular exposure, acquire recurring words and longer constructions in terms of the various (contextual) meanings with which they are associated. Schematisation refers to a special kind of abstraction which results in *schemas*, i.e. less detailed representations of structural commonalities between instances (Evans, 2019:133)—as seen in the range of associated meanings of the word 'safe' above.

According to Langacker, a usage-based account of language learning postulates that acquisition involves "a prodigious amount of actual learning, and tries to minimise the postulation of innate structures specific to language" (Langacker, 2000:2). Whilst cognitive linguists acknowledge that humans are, to some extent, biologically equipped to acquire language, they deny the existence of some innate cognitive module specifying grammatical knowledge. Instead, cognitive linguists maintain that we also employ species-specific sociocognitive capacities to acquire language (Evans, 2019:153-154). Since early studies in developmental psycholinguistics, one of the key cross-linguistic findings was that infants' first use of language appears to be item- rather than rule-based: they first acquire specific units (words), followed by their more complex combinations (pairs and sentences), before being able to develop (increasingly sophisticated) knowledge of grammar (Braine, 1976; Bowerman, 1973). Cognitive linguists maintain that this supports a usage-based theory of language acquisition[94] (Evans, 2019:154).

---

[92] That is, that all humans are born with a hard-wired infrastructure or pre-specification for a *Universal Grammar* that allows them to learn the grammar of any natural language with relative ease (Chomsky, 1965).

[93] Whilst Chomsky rejected such a position with his assumption that there is too little in children's experiences from which they can learn all they need about the complexities of language (his famous *poverty of stimulus* argument, Chomsky, 1959), Evans (2019:180) argues that Chomsky had little evidence to back up this assumption, and preceded the field of *developmental psycholinguistics* that investigated such issues more scientifically—and now largely disagrees with his view.

As the abstractness and complexity of the lexical units increase, linguistic creativity starts to develop (Tomasello, 2003). Hence, on this view, a lot of learning is involved in language acquisition, which are facilitated by domain-general cognitive and sociocognitive abilities. According to Tomasello (2003), these can be divided into two sorts of cognitive capacities that facilitate language acquisition: (i) pattern-finding abilities, and (ii) intention-reading abilities. Pattern-finding abilities refer to those general cognitive mechanisms that enable us to recognise patterns and statistically 'analyse' sequences of perceptual (visual, audio, etc.) patterns. Tomasello (2003) describes four capacities that make up this category, i.e. the capacity to associate similar objects and events (leading to their perceptual categories); the capacity to form sensorimotor schemas based on the recurrent perception of action sequences; the capacity to recognise and identify recurrent perceptual and behavioural sequences (as combinations of elements); and the capacity to create analogies based on functional similarities between some aspects in different wholes (as summarised by Evans, 2019:158).

To Tomasello (2003), such kind of pattern-finding abilities are what allows pre-linguistic children (infants under a year old) to abstract across utterances and construct their understandings of linguistic units. However, research demonstrates that such abilities are neither limited to language (see Tomasello, 2008, for a review), nor to humans (Evans, 2019:157). Thus, Tomasello contends that pattern-finding abilities are necessary, although not sufficient, to enable language use (as there are animals that have such abilities but do not use language). For that, he maintains, one also needs intention-reading abilities: where pattern-recognising skills enable infants to identify linguistic units, intention-reading skills are needed to imbue them with meaning. According to Evans (2019:158), this process occurs when, around one year of age, children start to understand other people as *intentional agents* with deliberate actions and changeable mental states. Again, Tomasello (2003) describes four capacities that comprise this category: the capacity to coordinate or share attention to similar entities in the world; the capacity to follow attention and gaze/gesturing; the capacity to actively direct the attention of others (e.g. through gesturing); and the capacity to learn how to (culturally) imitate the intentional actions of others.

---

[94] This is merely a brief overview of a cognitive linguistic approach to grammar learning, as a comprehensive discussion is beyond the scope of this paper. See Goldberg (2006) and Langacker (2000) for more thorough introductions by cognitive grammarians.

As pattern recognition abilities, these intention-reading abilities are domain-general. However, they are considered more species-specific—only humans seem to possess a full set of them[95]. According to Evans (2019:159), something that seems to distinguish humans from other species is our ability, even from infancy, to adopt a cooperative cognitive stance towards others. Language exemplifies this fundamental social part of our nature as it requires both an awareness of the communicative intentions of others and an awareness of the appropriate strategies for interacting with each other, in goal-directed communication (see Evans, 2015).

This section offered an introduction to some of the key approaches to understanding concept acquisition, conceptual structure, language acquisition, semantic structure, and the relationship between concepts and language in the cognitive linguistics discipline. Based on all the theoretical insights I have gathered, and arguments I have made, regarding human cognition and language use, I can now turn to the practical side to see how these various aspects have been, and are yet to be, implemented in artificially intelligent systems. Before doing so, I conclude this chapter with a brief summary of what I have gathered thus far.

## 4. Chapter summary and reflection

The first section of this chapter critically evaluated some of the major current theories of concepts: how they are structured in the mind, how they are acquired, and the relative strengths and weaknesses of each. Starting with a clarification of the term concept, I discussed how different theorists in various disciplines have defined *concepts*; i.e. as (some form of) mental representations, abilities, and abstract entities respectively. Thereby, I also clarified how I use the term for my purposes: as complex and distributed mental abstractions, associations and inferences from individual experience, that are involved in such abilities as recognition, categorisation, and language comprehension, and which tend to be, at least partly, directly accessible to memory.

From there, I critically evaluated the prototype, exemplar and theory theory, each within the context of what they were opposing, and what psychological phenomena they aimed to account for. Whilst the prototype and exemplar theories offered different accounts to explain typicality effects in human category explanation, and to account for fuzzy concept boundaries and borderline-cases, they both lacked a strength of the definitional account that they were opposing, i.e. our ability to identify features that are essential and not just salient, based on notions of causation rather than mere surface-level correlations. This tied into another point, which is that a theory should ideally be able to explain our context-sensitive, common-sensical judgement in selecting the relevant features for composing concepts in novel ways, which neither the weighted feature-list prototype theory, nor exemplar-based approach can clearly account for. Whilst the theory theory sought to improve on some of these failings, it lacked a crucial feature of the prototype and exemplar accounts, i.e. to explain how, despite having different theoretical frameworks by which to interpret concepts, different language users still have similar experiences in other regards (e.g. perceptual, affective and linguistic-conventional) that allow them to communicate their ideas/speak about similar concepts successfully.

---

[95] Although there is evidence to suggest that primates, like chimpanzees, are able to recognise members of their species as intentional agents, what they appear to be missing is a fully-fledged cooperative intelligence—what Tomasello calls *shared intentionality* (Evans, 2019:158-159).

To account for these key shortcomings—mainly, our ability to draw from complex sources of (sufficiently similar) subjective experiences and relevant background knowledge to interpret utterances, and novel compositions, in context—I drew from my discussion of connectionist theories of mind in §2.2 of ch.I, to offer an interpretation of an associationist account of conceptual structure. That is, that 'concepts' are not quite the discrete, coherent 'entities' that some theories (e.g. the classical theory/LOTH) make them out to be, neither are they fully describable in terms of (calculations over) discrete constituents (such as clear-cut properties, images, etc.). Rather, concepts, although we may talk about them as specific entities, might better be understood as complex (and weighted) distributions of multimodal sensory data (which may, themselves, have emergent structure[96] as coherent scenes, sounds, or sentences, but need not). Thereby, this approach is able to account for the integration of various inter-sensory modalities, including subjective associations with concepts such as affective states (e.g. associating CLOWN with *fear* based on bad experiences) or synesthetic associations (e.g. associating the number 5 with the colour red), or other inter-conceptual/conditioned associations (e.g. associating SALT with PEPPER, or the ringing of the doorbell with food, based on subjective encounters), etc. Such a connectionist view can also account for context-specific combinations of activations (i.e. context-specific interpretations of the senses of words), as different patterns of 'features' will (dynamically) be activated by different relevant contextual factors.

I also acknowledged that some of these associations may not be empirically acquired, but may be based on instinctual responses (e.g. evolved responses to certain behaviours/sounds that signal trouble), or may be, if not innate, common amongst people—given our specific embodiment, such as the basic experiential (e.g. spatial, orientational, etc.) metaphors that we use to structure our understanding of the world (as discussed in §3.1.1 of this chapter).

To explain the relevant components more in detail, §2 of ch.II turned to 4E theories of cognition. Through my discussion of embodied, embedded and extended cognition, I highlighted some major ways in which our cognitive function is determined by the kinds of bodies we have, as well as how our external (physical/sociocultural) environment plays an important (at least causal) role in shaping our cognitive behaviour, and drew some important contrast between such an approach and cognitivist/functionalist approaches to the study of mind. With my discussion of enactivism, I also highlighted the important role of goal-directed action, and embodied affect, on our perception of the world; most notably, the active perception of relevant affordances in objects, rather than the passive observance of objective surface-level features. From there, I also discussed the role of affect and embodiment on our social ability to 'mindread' or interpret the mental states/communicative intentions of others.

---

[96] This model seems to resemble some of those in complexity theory. For future research, it may be worth investigating this link in more detail.

Finally, I explored key findings in the developing cognitive linguistics programme that draws from various sciences to construct a naturalist, cognitively plausible understanding of human concept and language acquisition. That is, by characterising (i) linguistic principles that coincide with what other scientific disciplines have revealed about human cognitive/perceptive functions; as well as (ii) general principles that hold for all aspects of our language use. Thereby, I found that this neo-empiricist/embodied approach contrasts directly with some common underlying assumptions of formal semantic approaches, which includes: that semantic (context-independent) knowledge can be separated from pragmatic (context-dependent) and encyclopaedic (non-linguistic) knowledge; that sentence meaning relies on its correspondence to facts in an observer-independent world 'out there', which can be expressed in terms of a logical metalanguage; and that the meaning of complex expressions is compositional, whilst figurative language (being non-compositional[97] and non-literal) is exceptional (Evans, 2019:371).

In opposition, the experientialist perspective of cognitive linguistics places a bigger emphasis on subjective experience: that linguistically mediated meaning derives not *directly* from an observer-independent world 'out there', or abstract entities, but from structured representations in the mind that mirror our experiences in the world as embodied human beings (Evans, 2019:372). This is not to say that work in formal semantics should be discredited; it is probably the case that formal linguists tend to have different aims than those of cognitive scientists and psychologists. Rather, the argument is that formal semantics, although useful for some purposes, is not a good ground for a science of the mind and human language acquisition/use in general; for that we need a thorough engagement with the relevant (biological, psychological, etc.) sciences of human cognition and social behaviour. Formal semantics, on the other hand, typically takes a truth-conditional approach to meaning, by which meaning is understood as the conditions for a proposition[98] to be true (given a certain model of the world[99]). For this, one requires a clear-cut understanding of reference determination, knowing which names pick out which entities[100], as well as a clear understanding of how entities fit into specified sets (as discussed in §1.1 of this chapter) in the world-model.

Although this approach may be useful for a logical analysis into the formal representation[101] of *sentential* meaning (i.e. the propositional content of sentences), it is not always well suited for an analysis of *utterance* (pragmatic) meaning where information outside of the linguistic utterance itself (e.g. background knowledge, gestures, and other behavioural or contextual cues) need to be taken into account (see Chierchia & McConnell, 2000). As my purpose is to consider practical (i.e. cognitively feasible and computationally tractable) approaches for *grounded* language learning AI, to facilitate natural language understanding for artificial agents in real-world interactions, it is only within my scope to evaluate the suitability of models for the task at hand.

---

[97] That is, that the meaning of a complex expression does not amount to the combined meanings of its parts.
[98] That is, the semantic 'value' of a sentence, understood as a statement about the state of affairs in the world which can be either true or false.
[99] This world-model is a mathematical abstraction made up of sets.
[100] This is a topic of much debate in philosophy of language, e.g. Donellan (1966), Kripke (1972).
[101] That is, the expression of natural language statements in a logical language like first-order logic, propositional logic, or lambda calculus.

At the beginning of the chapter, I laid out some key aims in putting together a cognitively realistic, grounded theory of concept (and language) acquisition, given my ultimate aim of determining necessary requirements for enabling machines to do the same. To reiterate, that is, to account for (i) how we are able to learn language and acquire concepts that are sufficiently similar for communication to succeed; (ii) how that is possible given our unique set of subjective real-world experiences; and (iii) how that is explicable in terms of general cognitive mechanisms and abilities that evolved naturally. Through my research into embodied cognitive scientific/linguistic theories, I was able to offer biologically plausible accounts for (iii). Thereby, the answer to the remaining two points may be something like the following: that we, as similarly embodied organisms, are able to experience the world (and ourselves/our emotions) in largely similar ways, situated in an external physical world that is also structured in a certain way, as well as a specific socio-cultural environment that mediates conventions of language to help us express our (sufficiently similar) experiences in ways that are similarly carved up/structured and labelled, through currently existing linguistic conventions.

Whilst connectionists (as well as embodied theorists) typically reject the (classic computational) need for mental representations, I offered a looser understanding of concepts that accounts for the complex network of associations of connectionists, as well as the phenomenon of mental representations, in some form, that we are often able to call to mind when prompted by language—as emergent structures within the distribution[102]. Drawing from the cognitive linguistics literature, it seems these subjective 'representations' also consist in *simulations* or subtle sensorimotor behaviours in the rest of our bodies, as cognition is, evidently, not limited merely to the mind. Using a connectionist interpretation of conceptual structure, and understanding the role of our particular embodiment (as organisms of a specific size, shape, orientation, with certain emotional and cognitive capacities) in our language-use abilities, it is easier to see the possibilities (and difficulties) for modelling these relevant aspects in social machines using statistically-trained artificial neural networks, 'embodied' in robots. This is what I explore in the following chapter.

---

[102] As mentioned, these have also been illustrated by neural network literature (e.g. Buckner, 2019), where a deep learning algorithm is explained in terms of combinations of increasingly higher-level features being extracted (e.g. lines, eyes, faces, etc.) using merely weighted patterns of interconnected nodes.

Chapter III: Towards grounded language learning in AI

This chapter offers a critical investigation into the major current approaches for natural language understanding in AI. Particularly, I discuss and compare the relative strengths and limitations of different approaches for representing/grounding semantics in AI systems, guided by the aim of ultimately evaluating their relative suitability for addressing the grounding problem introduced in the first chapter, as well as their ability to account for (as many as possible) of the theoretical aspects of our embodied language learning/comprehension discussed in chapter II. I first offer a critical comparative evaluation of the main approaches (and individual models within them), followed by a summary of my findings in which I compare and contextualise the approaches in terms of the theoretical convictions they seem to reflect.

# 1. Current approaches in Natural Language Understanding

Research in the field of natural language processing can be divided into five main areas: Machine Translation; Spelling and Grammar Correction; Speech or Voice Recognition; Natural Language Generation; and Natural Language Understanding (Bose, 2004:2). Whilst all of these tasks are difficult for a machine to perform, the subfield of Natural Language Understanding (NLU) is particularly challenging, as it involves accounting for (a combination of) various complex semantic, syntactic and contextual factors that impact meaning. In finding ways to tackle such complexities, NLU aims to enable machines to interpret text appropriately, and is expected to ultimately overcome one of the longstanding goals of AI: machine reading (Etzioni et al., 2007).

To actualise this dream, as well as the aforementioned aim of introducing communicative robots into our homes and workspaces, much effort and resources have gone into developing means for enabling machines to construct semantic representations from natural language input[103] and/or execute spoken commands. In this section, I evaluate some of the main approaches in NLU to semantic interpretation; namely, semantic parsing, vector-space (distributional) semantics, and different varieties (and integrations) thereof. I also discuss some recent approaches that aim to 'ground' such models in multimodal information about the real world. By exploring and contrasting the various approaches, capabilities and strengths of current systems— and the sorts of computational methods they use—I aim to offer a general overview of the current state of the field to inform my later discussion of our relative progress in tackling the grounding problem, and to illuminate some of the various technical challenges involved in this task.

## 1.1. Semantic parsing

Central to the objective of NLU is semantic parsing, a method that aims to map natural language input to semantic/meaning representations. These representations can take various forms, such as logical formalisms (e.g. statements in first-order logic or lambda calculus), relational graphs[104], or perhaps commands in a programming language (Kamath & Das, 2019:3). *Parsing*, here, entails analysing a string of words to determine its underlying (syntactic) structure, whereas *semantic* relates to its interpretation, which is typically generated on the basis of an 'environment' (e.g. a knowledge base containing complex structured and unstructured data like relations between words[105], metadata of certain entities[106], or other bases designed for executing specific tasks[107]). Through such methods, meaning representations can allow systems to perform certain tasks involving automated reasoning, such as parsing queries for digital conversational assistants (e.g. Siri and Alexa) or interpreting and executing commands for robotic navigation (Kamath & Das, 2019:1).

---

[103] Thus far, the field has been dominated by resources for processing English, but the hope is that, eventually, systems would be able to deal with any natural language (Ananiadou et al. 2012:2).

Some basic elements of contemporary semantic parsing systems include a *grammar*, i.e. a set of rules that defines possible valid logical forms in the language. The kind of grammar that is used determines the expressivity and computational complexity of the model (Kamath & Das, 2019:3-4). Contemporary systems also typically include a *language model* that generates a probability distribution over the valid logical forms, and a *parser* that uses this distribution to predict which form (e.g. which syntactic analysis[108]) is most probable for the given input. Finally, models often use supervised, corpus-based learning algorithms to update their parameters towards better fitting the given (annotated) examples (Kamath & Das, 2019:3-4). Through using a combination of these, a semantic parsing system can try to establish which meaning representation is most appropriate for the given input.

As mentioned, one approach is to use models based on predicate logic, by which expressions are formalised in terms of objects (the heads of noun phrases), properties (e.g. *blue* or *round*) and relations between objects, based on their determined syntactic roles. The semantic parser then finds an interpretation (i.e. appropriate output) based on the logical representation and its match in the knowledge base (Bose, 2004:6). To analyse longer clauses, the rules follow the principle of compositional semantics, i.e. that the meaning of a complex phrase is a function of the meanings of the subphrases (Russell & Norvig, 2010:901). This classical, *logicist* view, follows aforementioned theoretical commitments in philosophy of mind (like Fodor's LOTH) and approaches in formal linguistics, in understanding language in terms of a formal metalanguage with a compositional, indexical semantics (Schubert, 2020).

Traditionally, semantic parsing systems were typically rule-based (and thus, highly task-specific), mapping parse trees to an underlying database query language using pre-defined rules (Kamath & Das, 2019:4). For instance, in syntax-based models (e.g. LUNAR[109] and ELIZA[110]), a syntactic parser analyses linguistic input in terms of its phrase structure (as a *parse tree[111]*) which is then mapped to an underlying database using rules. Such hard-coded approaches allow systems to respond appropriately to very basic kinds of (domain-specific) input. However, given how difficult and time-consuming it is to construct rules for semantic parsing by hand, especially for broader domains, over the past couple of years the field has seen a steady divergence from purely procedural, classicalist approaches to more flexible statistical methods aimed at allowing systems to learn the majority of linguistic and world knowledge autonomously (Kamath & Das, 2019:7-10; Schubert, 2020).

---

[104] That is, labelled graphs (or *semantic networks*) that illustrate the semantic relations between different entities/events (e.g. Banarescu et al., 2013; Oepen et al., 2015).
[105] For example, big hand-crafted bases like WordNet or Freebase (Bollacker et al., 2008).
[106] For instance, tables from Wikipedia (Pasupat & Liang, 2015).
[107] For example, booking flights (Hemphill et al., 1990) or answering geography queries (Zelle & Mooney, 1996).
[108] This can be determined by estimating the relative likelihood of different syntactic structures, and of different words/phrases to fulfil certain lexical/syntactic roles (Russell & Norvig, 2010:895).

This new statistical paradigm was enabled by the increasing availability of vast amounts of machine-readable text and speech data, increased computational power, and the hope that the new statistical machine learning methods would overcome the problems of scalability that had always troubled NLP, and AI in general (Schubert, 2020). The most popular statistical approach involves the use of supervised learning methods whereby algorithms are trained on given examples of desired input-output pairs. For semantic parsing, one such approach is to train systems on input-meaning representation pairs (e.g Zettlemoyer & Collins, 2005; Kwiatkowski et al., 2010). In Zettlemoyer and Collins (2005), for instance, the algorithm learns from examples of sentences paired with their correct lambda-calculus expressions[112]. To deal with ambiguity, a probabilistic grammar model may be used to rank possible parses for a given sentence in terms of their relative likelihood (i.e. commonality) (Kamath & Das, 2019:5).

Another approach trains systems solely on the intended *result* of the execution (also called the *denotation[113]*) of the formal meaning representation (i.e. program/instructions for mapping the input to output), absolving the need for intermediate representations altogether. These 'denotations' can take many forms, such as the answer to a given query, the execution of robotic commands, etc. (Kamath & Das, 2019:6). To illustrate, Berant et al. (2013) had a semantic parser learn how to answer queries from question-answer pairs on Freebase[114]. However, training models with such a weak form of supervision poses various challenges, such as requiring many varied examples of input-output pairs (to make sure the data does not overfit, as described earlier), and dealing with inevitable noise and variations in phrasing (Kamath & Das, 2019:6).

Another common approach to semantic parsing involves the use of knowledge bases that indicate some semantic or real-world relation between different words/concepts. For instance, rather than using the semantic structure of a particular language, a semantic parser can draw from a knowledge base that describes either semantic relations between the concepts involved (e.g. as sets and subsets) (Lehman, 1992:2) or other real-world relations (e.g. Donald Trump is *president of* the US, *husband of* Melania Trump, *owner of* The Trump Organisation, etc). Relational knowledge bases are also often combined with others—like the denotation-based approach above—to improve performance when dealing with unfamiliar formulations (e.g. Pasupat & Liang, 2015). Although there exist multiple hand-crafted relational knowledge bases (such as WordNet[115]) such bases can also be created automatically using *relation extraction* methods that infer relations between words from their use in unstructured text (using machine learning). In such methods, a relation is commonly defined in the form of a tuple $t = (e_1, e_2 ..., e_n)$ with $e_i$ representing entities in a predefined relation $r$ within a given document (Bach & Badaskar, 2007:1). Most relation extraction systems focus on extracting binary relations, e.g. *located-in*(*Stellenbosch University*, *Stellenbosch*), *mother-of*(*Judie Garland*, *Liza Minnelli*) (Bach & Badaskar, 2007:2). Another approach is to extract these relations as relational triples (De Marneffe et al., 2006), for example:

> (*founder*, *SpaceX*, *Elon_Musk*)
> (*worked-at*, *Tesla_Motors*, *Elon_Musk*)
> (*father-of*, *X_Æ_A-Xii*, *Elon_Musk*)

---

[109] Developed by NASA, the LUNAR prototype answers simple geology questions (Woods, 1973).
[110] ELIZA (Weizenbaum) is a semantic parsing system that manipulates input to ask questions (see Bose, 2004:6).
[111] That is, a representation of the grammatical structure of a sentence, expressed in terms of its constituent branches (lexical and syntactic categories) and leaves (lexical units) (Russell & Norvig, 2009:895).
[112] For instance, the query 'What states border Texas' will be matched to the formalism $\lambda x:state(x)borders(x;texas)$.
[113] Note that this is a different use in the term as in logic.
[114] Freebase was a large community-authored knowledge base consisting of relational triples, which became the foundation for Google's knowledge graph.

In these examples, the first component expresses a relation between two entities (the remaining components), and the underscores suggest that the names are some unique identifier of those entities. The relations are typically one of a small number of relations of interest that are predefined for a given task (Bach & Badaskar, 2007; Kamath & Das, 2019). Accumulating a large number of such relational triples creates a valuable knowledge base of facts about the world that may be used for knowledge-related tasks, like answering queries (as in the case of Freebase, mentioned earlier).

Traditionally, statistical algorithms were trained with a lot of supervised data, wherein the relations (if any) between different entities in a sentence are specified with annotations (e.g. 'owner of' or 'no relation'). Difficulties for both rule-based and hand-annotated approaches to semantic parsing include issues of reliability (in dealing with novel input), and, especially, moving from microdomains to broader, more varied domains (Schubert, 2020). A big reason for the trouble with scaling up is what has become known as the 'knowledge acquisition bottleneck': the challenge of specifying all the various, complex rules and facts required for more general (human-like) understanding (Schubert, 2020). More recent relation extraction models have tried to address such issues by relying on what is called *distant supervision*, which involves using relational triples from an existing, external knowledge base to automatically generate new annotations, thereby vastly extending the training data (Mintz et al., 2009). However, even then, the amount of background data required for disambiguating utterances in context is surprisingly large (see Jusoh, 2018)—especially without a grounded/common-sense understanding of words in goal-directed communication.

---

[115] WordNet is a large English lexical database that categorises nouns, verbs, adjectives and adverbs into sets of synonyms (*synsets*), each corresponding to a different concept. These synsets are connected in terms of certain semantic relations; e.g. antonyms, super-subordinate relations (i.e. subsets) and other hierarchies expressing differing degrees of specificity (Fellbaum, 2005:665).

This refers back to Dreyfus' (1992) aforementioned point, that we tend to rely on a sort of intuitive body-based 'know-how' for understanding language rather than lists of facts about entities or situations in the world. Such issues were also already anticipated by phenomenologists like Husserl (1960), who attempted to 'explicate' the noema for all types of everyday objects, but found that he had to include more and more of what he called the 'outer horizon,' in a subject's total knowledge of the world: "To be sure, even the tasks that present themselves when we take single types of objects as restricted clues prove to be extremely complicated and always lead to extensive disciplines when we penetrate more deeply" (Husserl, 1960:54-55). Marvin Minsky, cognitive scientist and AI researcher, ran into a similar problem:

> Just constructing a knowledge base is a major intellectual research problem …. We still know far too little about the contents and structure of common-sense knowledge. A 'minimal' common-sense system must 'know' something about cause-effect, time, purpose, locality, process, and types of knowledge …. We need a serious epistemological research effort in this area (Minsky, 1975:68).

As such, statistical NLU remains very task and domain-specific, and is mostly restricted to rather shallow aspects of language processing, such as extracting specific data regarding particular kinds of events from text (e.g. location, date, attendees, etc. of notable events), clusters of different argument types, task-relevant relational tuples, etc.) (Kamath & Das, 2019).

All things considered, the respective approaches to semantic parsing have different relative strengths and weaknesses. Whilst rule-based systems may be more robust (in highly limited areas), they are tedious to construct and extremely brittle, as rules more easily map on to certain phrasings of a message than others (Kamath & Das, 2019:1-2). On the other hand, statistical methods are more flexible and re-usable, yet, without predefined rules are shallower (being merely correlation-based) and, hence, may be less reliable, picking up incorrect relations or inferring the wrong syntactic structures (Marcus, 2018). Moreover, they require vast amounts of (sufficiently varied) training data to work effectively.

The main drawback for both approaches remains the pervasive ambiguity and underspecified context-dependency of natural language use; generalising to other linguistic tasks or domains, and dealing with novel input that does not resemble the given system's predefined rules/training data (although statistical methods are relatively better). However, despite these drawbacks, approaches in semantic parsing can be sufficiently effective at dealing with semantics as an issue of producing a suitable output (e.g. written, verbal or behavioural response) for a given input in natural language—*in specific tasks or domains*. Other tasks or domains may require different approaches. In the next subsection, I look at another big area in NLU that builds on a different theoretical foundation, modelling word meaning not as a bunch of rules or list of facts, but as a function of its contextual use.

## 1.2. Vector-space semantics

Whilst the idea behind semantic parsing was (at least initially) largely inspired by rationalist ideas about formalising natural language expressions via rules (e.g. as expressions in predicate calculus), and definitional approaches to meaning (e.g. defining words through hand-crafted semantic networks), another strand of research in semantic analysis takes a fundamentally empiricist approach. That is, understanding the meaning of a word purely in terms of its contextual use. Although nearly all contemporary statistical approaches have started to rely on contextual evidence in some way or other, vector space semantic models give it its most systematic and extensive application (Bruni et al., 2014:1). Given the growing availability of digital texts on the internet, recent research in NLU has increasingly been employing variations of vector-space semantics as a simple and practical method to derive word meaning representations from large sets of language data: as a vector representation (or *word embedding*) of its characteristics across different dimensions (called *features*) (Brychcín, 2015:21; Smith, 2020:3). The dimensionalities in vector space can be used for different purposes, sometimes remaining similar to traditional approaches that treat different words types[116] as discrete. For example:

- Each word type may be given its own dimension, and assigned 1 in that dimension (while all other words get 0 in that dimension).
- For a collection of word types that belong to a known class (e.g., days of the week), we can use a dimension that is given binary values. Word types that are members of the class get assigned 1 in this dimension, and other words get 0.
- For word types that are variants of the same underlying root, we can similarly use a dimension to place them in a class. For example, in this dimension, know, known, knew, and knows would all get assigned 1, and words that are not forms of know get 0 (Smith, 2020:3).

There are many such examples in NLP where vectors are used to represent word types, for features like syntactic classes (e.g. *verb* or *noun*), grammatical features (such as gender or plurality), or semantic characteristics (e.g. *animate* or *furry*). Such features can be designed by experts or derived automatically through learning algorithms (Smith, 2020:4).

An influential idea from structural linguistics is that words (or phrases) that are used in similar linguistic contexts are likely to have similar meanings (Harris, 1954:156)—called the *distributional hypothesis*. This idea has been exploited by a number of vector space models that take contextual similarity as a proxy for semantic relatedness. Drawing from large corpora, *distributional semantic models* (DSMs) record the functional (i.e. usage) characteristics of a word type by capturing the number of times it appears near every other word type, and mapping it as a vector (i.e. numerical list) representing its 'meaning' as a function of those instances of use (Baroni, 2016:3).

---

[116] *Type*, here, refers to a distinct word, in the abstract, as opposed to *token*, which is a particular instance of use of that *type*.

'Context', in distributional semantics, is defined in various ways by different models: Brychcín (2015:22) broadly distinguishes between a local and global context. Models that use a *local context* are most common; these collect short-range usage contexts (i.e. sequences of words around the target word) using a moving *window* (of a specified size) to capture the features of its immediate neighbours. To illustrate, given sentences such as 'A cow is an animal', 'The old lady milked a cow', 'I was chased by the cow', etc., a distributional model might start to infer the contexts in which the word *cow* typically appears (i.e., the words it is typically surrounded by), represented as a continuous-valued vector in a space where each dimension corresponds to a different feature (i.e. functional characteristic) (Bengio, 2008:3881). Being sensitive to word order, such models can also be used to model specific syntactic relations among words (Bruni et al., 2014:5; Boleda & Herbelot, 2016:624).

The main advantage of the local-context approach is that it can efficiently capture the semantic similarity between words. Given that vectors can be understood as points in a geometric space (with their values as coordinates), the set of vectors produced by a distributional semantic vector is called a 'semantic space', and semantic similarity is often quantified geometrically, e.g. as the proximity between vectors in that space (Baroni, 2016:3). That is, functionally (and thus possibly semantically) similar words will, at least in some directions, be closer to each other in the semantic space, which helps the system predict the probability distribution over the next word in a sequence, as well as which words can replace each other in similar contexts (Bengio, 2008:3881).

On the other hand, models that use a *global context* are usually based on the 'bag-of-words' hypothesis, which assumes that words have related meanings if they occur in similar documents (i.e. a sentence, passage, paragraph, or text)—irrespective of their order (Brychcín, 2015:23). For instance, if the document is about *golf*, it is likely to contain words like 'Green' or 'Birdie', which are then taken as semantically related. An example of this is *probabilistic topic models*, which assume that words in a document exhibit some probabilistic structure connected to their topics (as specified in advance) (e.g. Blei et al., 2003; Griffiths et al., 2007). Thereby, words are either represented as a vector of topics, or each topic is defined as a probability distribution over different words (Bruni et al., 2014:6).

Underlying all word vector models is the notion that "the value placed in each dimension of each word type's vector is a parameter that will be optimized, alongside all the other parameters, to best fit the observed patterns of the words in the data" (Smith, 2020:7). This operational approach to dealing with word meaning has become increasingly popular in computational semantics—in fact, Brychcín, (2015:23) calls it the "backbone research area in NLP"—as well as more broadly in AI (Boleda & Herbelot, 2016:623). More than a practical method for semantic representation, vector-space semantics is seen as providing a theoretical framework for explaining how the meanings of words are constructed in memory, that assumes a "formal cognitive mechanism to learn semantics from repeated episodic experience in the linguistic environment" (Jones et al., 2015:239). The idea has multiple theoretical roots in psychology, structural linguistics, lexicography and the contextual view of meaning advocated by Wittgenstein (Wittgenstein, 1953; Harris, 1954; Firth, 1957; Miller & Charles, 1991).

Whereas symbolic semantic representations (like those in first-order logic or semantic networks) are discrete and categorical, distributed representations are graded and distributed[117], given the continuous values in vector spaces (Lenci, 2018:153). Hence, by investigating the interplay between meaning and contexts, these models exhibit a gradient approach to semantic representation, consistent with psychological views of 'fuzzy' conceptual boundaries (Murphy, 2002). As such, distributional models are better suited to tackle the dynamicity and plasticity of meaning (Bruni et al., 2014:4), and many have started to take DSMs seriously as plausible models of how we acquire and use semantic/conceptual knowledge ourselves (e.g. Lenci, 2008; Baroni, 2016; Bryson, 2001).

One drawback of models that assign a global embedding to each word is that they are unable to account for cases of polysemy.[118] As such, deeper networks have started providing different representations for various senses of the same word (Young et al. 2018:59). Rather than representing a word type as a fixed data object, trying to encapsulate all its features at once, *contextual* or *distributional* word vectors are created from a combination of type-level vectors and neural network parameters that 'contextualise' each word (Smith, 2020:10). This means that every instance (i.e. token) of a word, e.g. *plant*, will have a different embedding; vectors with a context that contains more references to flora should be closer to each other, whilst those that seem more likely to refer to manufacturing centres will cluster elsewhere in the semantic space (Smith, 2020:10).

Another common limitation of word embeddings is their inability to represent compositional phrases of which the meanings are not reducible to the combined meanings of their parts.[119] In recent years, some models have been extended to deal with some versions of compositional semantics (e.g. Baroni & Zamparelli, 2010; Socher et al., 2012), although they are yet unable to account for the wide range of composition phenomena examined in formal semantics (Boleda & Herbelot, 2016:626). Another limitation is introduced when embeddings are learnt based only on a small window of surrounding words, and so semantically-similar words that express opposing sentiments may be clustered together (e.g., words like 'good' and 'bad' can share almost the same embedding)—which is particularly problematic for tasks that require sentiment analysis (Young et al. 2018:59). Moreover, Young et al. (2018:59) point out that a general drawback of word embeddings is that they are highly task-specific, given the great variations in linguistic contexts, and training them from scratch for a new application is very heavy on time and resources. Despite the optimism of some, discussions have started to emerge on the relevance of distributional feature vectors in the long run; a growing consensus in the AI community suggests that adequate representations of words and concepts cannot be inferred from distributional semantics alone (e.g. Kiela et al., 2016; Baroni, 2016; Lucy & Gauthier, 2017); that is, representing the meaning of a word entirely in terms of its connections with other words. Essentially, it still faces the symbol grounding problem discussed earlier (Baroni, 2016:4).

---

[117] That is, the *meaning* of a word is *distributed* over the whole vector, in continuous values across multiple dimensions.
[118] This occurs when the different established senses are related in some way, usually by metaphorical extension (e.g. metonymic expressions like 'the kettle is boiling') or transference (e.g. 'walk through the door').
[119] For instance, idioms like 'cold turkey' or named entities like 'Boston Globe' (Young et al. 2018:59).

Some have argued that the concerns about DSMs not being grounded or embodied are exaggerated, as the co-occurrence patterns in our use of language that are extracted by distributional models reflect the semantic knowledge that we gained in perception, and hence, there is a strong (enough) correlation between linguistic and perceptual information (e.g. Louwerse, 2011; Bryson, 2001). For instance, because ravens are normally black, or mice are more often grey than, say, blue, we are most likely to use the words *black* and *raven*, and *mice* and *grey* together than other colours. Consequently, children can learn useful facts about the appearance of these animals without having perceived them in person (which also explains why subjects born blind can have excellent knowledge of colour terms, e.g. Connoly et al., 2007). Therefore, some have hypothesised that the meaning representations extracted from the (extensive) use of words in texts may be effectively indistinguishable from those derived from perception, making grounding redundant (Bryson, 2001; Bruni et al., 2014:3). In fact, Bryson insists that distributional semantics is sufficient for 'grounding' language, and the role of embodiment has been overstated: "A basic premise of this paper is that human-like semantics can be derived without any particular plant or embodiment" (Bryson, 2001:2).

However, much research has revealed this not to be the case. Many studies have underlined that, although text-derived embeddings capture much of the encyclopaedic, functional and discourse-related properties of word meanings, they tend to miss their more obvious or concrete aspects (e.g. Baroni & Lenci, 2008; Andrews et al., 2009; Riordan & Jones, 2011). For example, whilst we might harvest from text encyclopaedic information about bananas (e.g. being fruit and edible), there may be a lack of more obvious descriptions, such as the fact that they are typically yellow and curvy, as few authors may feel the need to explicitly write down such obvious observations—not to mention more basic properties, like the fact that dogs have heads and eyes (Bruni et al., 2014:3). Contrarily, the same studies show that humans, when asked to describe concepts, typically do so mainly in terms of salient observable (and distinguishing) features. Moreover, the fact that we tend not to describe objects in terms of their salient characteristics unless they are atypical, means that models often model words in very unintuitive ways:

> According to the DSM, the sky is green; flour is black, and violins are blue. Intuitively, the color of violins is so deeply entrenched in our visual experience that few writers will explicitly refer to brown violins, and consequently distributional models will lack crucial evidence about basic object color (Google queries issued on January 5, 2015, returned about 3500 hits for brown violins, more than 5500 for blue violins) (Baroni, 2016:4).

According to Bruni et al. (2014:3), this discrepancy between DSM and our perceptual knowledge is not, in itself, a damning criticism for these models if they succeed at the tasks they are used for. However, if we are interested in the potential implications of such approaches as models of how humans acquire and use language—as is the case for many developers (e.g. Lund & Burgess, 1996; Landauer & Dumais, 1997; Griffiths et al., 2007, Lenci, 2008)—then their complete lack of grounding in perception seriously discredits their psychological plausibility, and exposes them to the same criticisms raised against classic symbolic models (Bruni et al., 2014). This includes the more serious limitation that it is not clear how feature vectors can establish the link between words and the real entities they denote in the external world—one that seems essential for comprehending most linguistic expressions. In the next subsection, I evaluate different methods for 'grounded' semantic modelling in NLU.

## 1.3. Grounded/multimodal approaches

In order to achieve human-like command of natural language in AI, especially in terms of relating language to real-world objects and events, there is a growing recognition that statistical learning methods will have to incorporate perceptual and conceptual modelling of the world. This is explored in an area of NLU called *grounded language learning* (Schubert, 2020). Approaches under this class seem to employ different understandings of what the 'symbol grounding problem' amounts to, and approaches for addressing it range from simple cross-modal approaches that incorporate text and images, to humanoid robots executing commands on real-world objects. In this subsection, I review various such approaches and evaluate their relative adequacy for appropriately addressing the grounding problem, as described in §2 of ch.I.

### 1.3.1. Grounding using images

A popular approach in NLU is to create *multimodal* DSM by simply extending text-based vector-space models to include embeddings of another modality, typically images (e.g. Feng & Lapata, 2010; Leong & Mihalcea, 2011; Bruni et al., 2014; Baroni, 2016). Although some researchers have carried out studies demonstrating the benefits of integrating text-based distributional models with auditory and olfactory information (e.g. Kiela et al., 2015), vision is commonly taken as, what is argued, a reasonable starting point "both for convenience (availability of suitable data to train the models) and because it is probably the dominating modality in determining word meaning" (Bruni et al., 2014:4)—at least for those of us with the ability of sight.

The most effective contemporary visual recognition systems do not capture the contents of images as complex data structures, like 3D figures (Marr, 1982), but as fixed-size vectors mapping various low-level features of an image, as given in the colour properties of (clusters of) pixels (Baroni, 2016:5). For example, the Bag-of-Visual-Words (BoVW) method carves up image content into a set of small clusters (e.g. square areas surrounding each pixel) that are mapped to a pre-determined discrete vocabulary of 'visual words' (e.g. Sivic

& Zisserman, 2003; Yang et al., 2007; Bruni et al., 2014). This technique was inspired by the aforementioned bag-of-words method in information retrieval, whereby textual information is represented as an unordered collection ('bag') of words found in a document. BoVW extends this idea to images by describing them as a collection of discrete regions (i.e. clusters of pixels), capturing their appearance (i.e. pixel colours/intensities) whilst ignoring their spatial structure (similar to ignoring the order of words in a text), which allows it to deal with the various ways in which an object can be captured a picture[120] (Bruni et al., 2014:4).

Similar to word embeddings, these visual features are automatically detected through a learning algorithm that recognises patterns in pixel clusters across its training data, and associates each 'type' of cluster with a distinct 'visual word' (Bruni et al., 2014:4). The whole image is then represented by a vector (with dimensions the size of the visual-word vocabulary) with values recording, in each dimension, the number of relevant pixel clusters mapped to it from the image. Visual words typically capture simple visual attributes, such as colours, oriented edges, visual textures, etc. (Baroni, 2016:5). On the basis of such visual words, classifiers can determine which features are most discriminative of the object to be recognised (Grauman & Leibe, 2011). The embeddings of image features can then be combined (*fused*) with word embeddings into a single vector space, based on the sets of images connected to a word in multimodal corpora. A very simple, effective method of multimodal fusion is to concatenate the visual and textual vectors representing a word, resulting in a new vector that spans both visual and textual dimensions (Bruni et al., 2011; Kiela & Bottou, 2014). These vectors encompass contextual similarities in both linguistic and visual features, and word embeddings are thus 'grounded' in (at least some form/level of) perceptual information (Bruni et al., 2014:11)—or perhaps, rather, enhanced by it.

More sophisticated machine learning algorithms, like deep convolutional networks (e.g. Szegedy et al., 2014; Zeiler & Fergus, 2014), are also increasingly used to learn more abstract feature combinations, as the added 'hidden layers' make it possible to pick up more complex correlations between pixel properties. As text-based DSM, *visual vectors* can exploit the vast amount of image data available on the web, and are typically extracted from pictures on photo-sharing sites such as Flickr (Baroni, 2016:5). The words that label the pictures, which associate the corresponding vectors to the right objects/visual properties, are usually done through manual annotation in the training data, but could also be learned directly from the co-occurrence patterns of words and images (e.g. from image tags on Flickr)—although learning from such noisy data sources can be unreliable (Baroni, 2016:6).

By enhancing word embeddings with notions of the images they are correlated with, it is possible to address some of the issues of pure word embeddings mentioned earlier. Textual and visual data encode different systematic biases about concepts. As such, combining them helps to generate more balanced representations, correcting their respective biases/shortcomings (Baroni, 2016:6). For example, whilst horses might more often be described in text in terms of their functional role as vehicles, their representations in images contain more features pertaining to their appearance, similar to that of other large mammals. Combining the two, one might get a representation that better captures both their functional and observable properties; i.e. through their similarity with visual embeddings of other animals. This also helps to distinguish between objects that are often given similar treatments in texts (e.g. dogs and cats) or images that look similar but have different functions (Baroni, 2016:6).

---

[120] For instance, as objects like buildings can be photographed from many different angles with varying shapes and backgrounds, the BoVW representation of a building might amount to little more than "object with many vertical and horizontal edges" which can effectively generalise to a variety of building types and views (Baroni, 2016:5).

Despite these relative corrections, text corpora and image collections are quite dissimilar to the linguistic and perceptual environments from which we learn word meanings. Moreover, both types of embeddings remain subject to the same challenges of deep learning algorithms I listed in §2.2 of ch.I—such as not being well integrated with prior (real-world) knowledge, not being able to inherently distinguish between causation and correlation, presuming a largely stable word (or at least one that remains close to the training data), and not being fully reliable (see Marcus, 2018). A benefit of visual vectors created through machine learning, is that the object, as well as its surroundings, are taken into account; for example, noticing that all furniture examples are indoors, emphasising the link between the meaning of objects and the contexts in which they occur (Baroni, 2016:7). However, this could also be a limitation, as if all the images the training algorithm was exposed to were of furniture indoors, it may fail to classify a chair as such if it is in a different environment[121].

Moreover, as concrete objects are the easiest to extract from images, most of the work in multimodal DSM approaches limit their focus to concrete nouns, although some work on image-text fusion has explored other features like colour adjectives[122], and spatial/texture concepts (e.g. Hudelot, 2005). However, embeddings based on images have the limitation of not being well suited to deal with more abstract concepts or other linguistic units that have no clear pictorial interpretation, as well as verbs (referring to processes or activities) or events (like different days, seasons, etc.) that depend on progression through time (Baroni, 2016:7). Regneri et al. (2013) try to address these latter examples through using information extracted from videos, which revealed a great improvement over text alone[123] when assessing the similarity of action-depicting verb phrases. However, their test was only done on a small scale, and the sort of richly annotated video data their models require would be very challenging, time-consuming, and expensive to produce on a large scale. According to Baroni (2016:7) the question remains open for future developments of less label-intensive approaches to grounding (the visual aspects) of verbs.

---

[121] For instance, if a classifier is trained on images of trains where there are always clouds in the sky, it may infer that clouds are a necessary part of the classification.
[122] Bruni et al. (2012) found that multimodal DSMs outperform standard DSMs not only on attributing the right colour to concrete objects, but also in distinguishing literal and metaphorical usages of adjectives.
[123] That is, 20% more accuracy when measured to human ratings (Regneri et al., 2013).

As mentioned earlier, the pixel clusters assigned to the same visual word generally tend to be patches with similar low-level appearance, rather than higher object-level identification, which reveals the reason for some of the biases and faulty classifications of 2D images (without real-word knowledge of the objects they represent). As such, Bruni et al. acknowledge that "a truly multimodal representation of meaning should account for the entire spectrum of human senses" (2014:38). However, as is evident from chapter II, this is no small feat, as our senses do not merely include sight and sound, but bodily experiences such as basic sensorimotor concepts, concepts of emotion, of time passing, folk-psychological understandings of people and social situations, etc. In the next subsection, I look at recent attempts to 'ground' language in combinations of robot action, perception, and interaction.

### 1.3.2. Grounding using (inter)action and perception

Apart from the agent-neutral, purely computational approaches to grounding by relating different inputs mathematically, i.e. *Cognitivist Grounding* (Ziemke, 1998:89), there is also a growing interest in approaches in *Embodied/Enactive Grounding* that emphasise the relevance of embodiment and agent-environment mutuality (inspired by the embodied/enactivist paradigm described earlier) (Ziemke, 1998:90). Rather than focusing on low-level properties in images to enhance the feature representations of (mainly) concrete nouns, this class of approaches typically aims to help physical robots learn possible correspondences between parsed instructions and real-world objects, regions, and/or actions, situated in a three-dimensional environment (Paul et al., 2016:1). Whilst the former is useful for specific language processing tasks (e.g. query-answering; having simple conversations; processing simple digital commands) the latter is vital for the increasing interest in letting robots cohabit with us in our factories, workplaces and homes where effective and efficient human-computer collaboration is crucial (Paul et al., 2018:1269). According to Hristov et al. (2017) this requires the capacity to interpret ambiguous, contextual commands, communicated in a way that feels natural and unobtrusive to the (non-expert) human interlocutor. Specifically, the robot must be able to:

- Understand natural language instructions, which might be ambiguous in form and meaning.
- Ground symbols occurring in these instructions within the surrounding physical world.
- Conceptually differentiate between instances of those symbolic terms, based on features pertaining to their grounded instantiation, e.g. shapes and colours of the objects (Hristov et al., 2017:1).

This problem of *physical* grounding has been approached from multiple perspectives. Many models rely on the use of prespecified categories to ground sensory input, such as using a pre-annotated semantic map to interpret instructions (e.g. Matuszek et al., 2013) or using a pre-trained symbol classifier to decide whether a detected object can be labelled with an one of a set of anticipated symbols, such as certain colours or shapes (e.g. Matuszek et al., 2014; Eldon et al., 2016). Unlike human infants, such a system depends on being properly prepared with manually annotated data prior to any online interaction. For a more cognitively realistic approach, the Cross-channel Early Lexical Learning (CELL) model is a simple example of a model that learns directly from raw 'first-person-perspective' sensory data to ground words, particularly relating to object shapes (Roy & Pentland, 2002).

The central aim(s) of CELL is to address three interrelated aspects of early lexical acquisition: (i) how children learn the boundaries of different words in their language from natural speech (ii) how they learn perceptually grounded semantic categories, and (iii) how they learn to associate the two appropriately (Roy & Pentland, 2002:114-116). In essence, CELL achieves this by inferring the most probable word-to-symbol associations by abstracting consistent word-to-context patterns from multiple situations—specifically, correlations between visual input (of objects with different shapes) and words for shape types. As such, the model is mainly apt for learning concrete nouns whose referents can be observed directly, not addressing more abstract or relational concepts. Even so, Roy and Pentland (2002:116-117) explain that the acquisition of word meaning in context is not trivial, given the multiple levels of ambiguity; for instance, firstly, the fact that objects are often referred to in their absence, even in infant-directed speech. This poses a difficulty for such a system that learns associations merely from observed co-occurrences. Secondly, as discussed earlier, ambiguity may also arise from the fact that an unlimited number of referents—including objects as well as (combinations of) their features—can be picked out from the same context (Quine, 1960). To help the system focus on the correct referents, some prior bias can be pre-programmed to favour some features over others[124] (like human perceptual biases/gestalts discussed in §3.1.1 of ch.II). CELL takes an extreme version of this by limiting its focus to only one type of contextual attribute, i.e. object shape.

Further ambiguities arise as audio and visual signals are both perceived through mechanisms that are susceptible to various sources of variation (in perspectives, lighting, tone of voice, etc.) and noise. According to Roy and Pentland (2002:117), similar to humans, example-based computational models can infer notions of central *prototypes* (ideal forms of the category) from multiple observations. In CELL, a probability density function is used to deal with variations input signals by measuring its relative distance from different such prototypes. That is, each prototype has a radius of allowable deviation that determines whether a new percept should fall within that category (Roy & Pentland, 2002:117).

Whilst it provides, in some aspects, a more cognitively realistic model of learning from perceptual input in natural conditions, the CELL model rests on multiple simplifying assumptions that significantly limit its capabilities. Firstly, to simplify visual processing, CELL assumes that each spoken utterance is paired with a single object, and its focus is limited to a certain type of feature (shape) of those objects. According to Roy and Pentland (2002:140) this is somewhat justified by the fact that young infants are known to have a 'shape bias', in that they tend to find words to describe shapes rather than colours and other visual attributes (see Landau et al., 1988). Roy and Pentland (2002:140) suggest that, to cope with more realistic encounters of multiple objects and complex backgrounds, CELL would require a model of visual selection to determine which phenomenon to pair with which utterance—although this still would not account for more abstract concepts/reference to absent or relational entities. Even so, they maintain that visual selection poses a great challenge, as it involves many complex issues, including an interpretation of speaker intent, and the parsing of complex visual scenes (Roy & Pentland, 2002:140).

---

[124] For example, Sankar & Gorin's (1993) model only represented shape and colour attributes, thus constraining their model to only learn words groundable in these input channels.

Finally, another significant assumption in CELL's representation of perceived objects is that the correspondence between different perspectival observations of an object is given. That is, when it receives a set of views of a given object, CELL assumes that all images in the set belong to the same entity, but it is unable to autonomously determine the correspondence between different view-sets of the same object. As two such view-sets will never be identical (due to visual variations in perspective, lighting, etc.), the system faces a  correspondence problem at this level (Roy & Pentland, 2002:140).

Some recent approaches have tried to address a few of these shortcomings. Yu et al. (2005) propose a model that processes spoken natural language input corresponding to images of various objects combined with the direction of a speaker's gaze, as measured by a head-worn eye-tracking device. To test it, speakers narrated stories (in their own words) based on illustrations in a simple children's book. These contained multiple objects, so that for any co-occurring utterance there were multiple possible visual referents. To resolve reference ambiguity, detailed eye-movements of the speakers were traced and automatically analysed to detect their probable points of focus within a given visual scene, which, as in CELL, were then used for online cross-modal associative learning (Yu et al., 2005).

According to Roy (2005:391), this integration of speaker gaze-tracking offers a significant extension beyond the CELL model as it makes use of some kind of interactive/social information, which is known to be crucial in natural language learning/comprehension. Hristov et al., (2017) use a similar approach to integrate user interaction to ground natural language units for a robot in a real-world task-oriented scenario. Rather than using images in a book, their model processes a raw stream of cross-modal input—i.e. linguistic instructions, visual scene perception, and a concurrent trace of 3D eye-tracking fixations—to associate the correct object attributes with their corresponding names. Thereby, it maps spoken (natural language) commands to a planned behaviour. Unlike CELL, this system relies on a predefined set of concepts (for object shapes and colours), and exploits the notion of *intersective modification*; i.e. that more than one symbol can be used to describe an object (Hristov, et al., 2017).

The meanings of the symbols are learned with respect to various observed attributes of a given object, which are limited to a set of low-level features such as colour intensities (in the primary colour channels) and areas of pixel clusters of any specific colour (Hristov, et al., 2017:2). When a new instruction is received, each symbol has a probabilistic classifier that decides which object (and its features) to associate with it. As in CELL, the classifiers are not trained beforehand, but learn from real-time interactions with the speaker: images of the objects are extracted from the high-frequency eye-tracking and video streams, as their corresponding names are extracted from the parsed instructions (Hristov, et al., 2017:2). Emphasising the social aspect of learning, the human participant is simultaneously demonstrating to the robot how a given task should be executed, as well as which visual attributes must be present in the surrounding objects for the task to succeed. For instance, in observing how to make a Greek salad, the agent would not only learn the sequence of steps, but also a sense of the visual features of the different ingredients (and their associated words) (Hristov, et al., 2017:3).

The model consists of an end-to-end process, taking raw linguistic and video data as inputs to learn the meanings of symbols referring to certain (prespecified) shapes or colours. To simplify the processing of (a specific set of attributes of) individual objects—as patches of pixels—the background is removed through a standard background subtraction method, assuming that most of the image will be occupied by a single solid object with a loosely uniform colour. As such, images that have no clear object to detect are considered noise and are discarded (Hristov, et al., 2017). Each symbol becomes associated with certain extracted features, where colour symbols are characterized by the extracted RGB values, and shape symbols from certain patterns of pixels in a patch. A semantic parsing technique is then used to map each natural language instruction (one sentence per instruction) into an abstract representation, in the form a tuple with the format (*action*, *target*, *location*). *Action* corresponds to an element from a predefined set (such as TAKE, PICK, DROP, etc.), *target* corresponds to a list of descriptors for an object/attribute in the environment, as grounded through visua feature extraction (i.e. BLUE, RED, YELLOW; CELL, CUBE, BLOCK), and *location* corresponds to a physical location in the environment. Thereby, the semantic parser will be able to find a mapping from each uttered sentence to a corresponding instruction (Hristov, et al., 2017).

Given a new test image, however, the algorithm fares much better at recognising presented colours than shapes (93% to 56%). According to Hristov, et al. (2017) this owes the fact that, whilst RGB values are able to describe colour values quite precisely, pixel clusters alone are not enough to describe the concept of shape: the algorithm may confuse different object types with each other purely on the basis of having similar sizes. Hence, they also emphasise the need for considering a wider (and more complex) range of feature extractions to pick up finer-grained distinctions between objects (Hristov, et al., 2017). Moreover, as their system focuses on mapping attributes to a pre-specified list of symbols, it fails to deal with novel shapes or colours. As such, they propose that future models should try to account for the learning of novel concepts—perhaps also reliant on user interaction[125]. For instance, if the agent observes a new hue of an existing colour (e.g. teal), or a new colour altogether (e.g. purple) it might prompt the user for a new label (symbol) and record it for future reference[126] (Hristov, et al., 2017). Another limitation of their model is that it can only create compound nouns by combining an adjective (as a salient visual attribute, e.g. *blue*) and a noun (e.g. *cube*), which would not hold for other kinds of adjectives that do not describe immediate visual properties (e.g. *steak knife*). For representing more complex compositional relations, they suggest further refinements, such as using a hierarchical approach (Sun et al., 2014).

---

[125] For exploring another avenue of user-robot interaction, Rosa et al. (2018) offer a system that integrates the use of user-worn smart devices to learn space semantics (i.e. the names of different rooms, like bedroom or kitchen), based on the tracking of a user's different patterns of movement in different rooms.

[126] See Thomason et al. (2017) and Parde et al. (2015) for such approaches.

Despite the simplicity of their model, Hristov, et al. (2017). claim that it provides useful advancements towards learning more complex language structures in future, such as grounding prepositional phrases or learning new actions an online and context-specific (i.e. attribute-specific) manner. To illustrate, once the system is able to distinguish an object by multiple attributes (e.g. *red block*), it could more easily learn what it means for another object to be beneath/next to it (Hristov, et al., 2017). For dealing with more abstract relational concepts, like 'pick up the middle cube in the row of three cubes', Paul et al. (2018) propose a probabilistic model that incorporates notions of abstract spatial concepts (e.g. NEAREST, LEFTMOST, etc.) expressed as probabilistic hierarchical groundings to a specified set of concrete objects (e.g. *block*), as well as notions of cardinality (e.g. *one*, *seven*, etc.) and ordinality (e.g. *first*, *seventh*, etc.) for interpreting complex spoken commands.

Apart from objects and their properties, various approaches have been proposed for the grounding of verb meaning. As mentioned earlier, one approach is to ground verb meaning in the analysis of video sequences. For instance, Siskind (2001) offers a model that grounds of human hands manipulating coloured blocks; that is, understanding verbs as recognisable (perceived) actions, expressed as logical relations. The meanings of basic verbs are modelled as 'temporal schemas' that specify expected sequences of force-dynamic interactions between entities. For instance, the meaning of the utterance 'hand picks up block' can be modelled by the sequence of interactions: "TABLE-SUPPORTS-BLOCK, HAND-CONTACTS-BLOCK, HAND-ATTACHED-BLOCK, HAND-SUPPORTS-BLOCK" (Siskind, 2001). The temporal relations between such force-dynamic features are described using 'Allen relations', which capture thirteen possible logical relations between pairs of time intervals:

> For intervals A and B, Allen relations include: *A ends after B starts*, *A ends exactly as B starts*, *A and B start together but A ends first*, and so on. Higher level actions are defined in terms of these lower level schemas. Thus, *move* is defined as the ordered sequence of the schemas corresponding to '*pick up*' followed by '*put down*' (Roy, 2005:392).

However, according to Roy (2005:392), using logical relations to describe action-sequences makes it hard to differentiate between more nuanced forms of similar action patterns (e.g. *push* versus *shove*). Moreover, verbs refer to both the observation and performance of action; merely being able to classify an action visually is not enough:

> A shortcoming of the standard view of lexical acquisition is that it provides no account of how a child learns to make use of the concepts she learns and the words that label them…even when the network learns perfectly how to classify a domain, it has no mechanism for inference or action (Bailey et al., 1997:1).

This is addressed by Bailey et al. (1997) who developed a model that learns verb meaning in terms of action control structures, called *execution schemas* ('x-schemas' for short), which control sequences of movements of a robot arm. A verb is defined in terms of its associated x-schema (networks of activity patterns) and control parameters. According to Roy (2005), understanding verbs as action/control patterns make it easier to distinguish between their different subtle aspects: the verbs *put down* and *pick up* can be distinguished by the structure of their associated x-schemas (i.e. differences in action patterns), whereas *push* and *shove* may have similar x-schemas, but different force/velocity control parameters (i.e. similar action patterns, executed differently). However, their model faces some shortcomings: for one, it offers no account of the category structure connecting different word senses (Lakoff, 1987). It also fails to explain how verbs (as x-schemas) might be extended to form related phrases like *push through*, or how to map words relating to perceived objects/events to more abstract domains, as we do for metaphors—a limitation faced by the field in general (Bailey et al., 1997:24).

Whilst Siskind's model focuses purely on the visual grounding of verbs, and that of Bailey et al. on verbs as action commands, Roy (2005:392) maintains that a model should ideally be able to account for both aspects at once—either through some kind of bridging structure between perception and control systems, or through the development of a system that can create a single representation integrating perception and action (suggested by research in enactivism).

The intertwined nature of action and perception is not only relevant for grounding verbs but also nouns, as many objects can be differentiated by their functions/affordances; for instance, to differentiate between *round* and *ball* (Roy, 2005:392). For this, Roy (2005) proposes a framework for grounding words (nouns, verbs, and adjectives) in terms of structured networks of sensor and motor primitives: verbs are grounded in sensorimotor control systems, similar to x-schemas. Adjectives (describing object attributes) are grounded in action-relative control expectations; that is, the meaning of different types of attributes are linked to the expected embodied means of engaging with each. For instance, the grounding of a particular colour term (e.g. *blue*) is coupled to the motor control program for turning the robot's gaze towards an object; and the meaning of a word for an attribute of weight, e.g. *light*, is grounded in 'haptic expectations' associated with picking objects up (Roy, 2005:394). Notions of location are also used, and are encoded in terms of spatial coordinates relative to the agent: objects are represented as collections of properties tied to a particular location, including encodings of motor affordances for different ways of interacting with the object. Concerning the round vs. *ball* example, this model takes the meaning of *ball* to subsume both the meaning of *round* (as one of its multiple properties) and all of the movements that may affect the ball (Roy et al., 2004).

Roy et al.'s (2004) model is implemented in Ripley, a robot arm[127] (that includes various sensors to serve as, for instance, 'eyes' and 'ears' ) that is able to convert spoken instructions like 'hand me the red one on your left' into situated action (see **Fig. III.1**). Ripley maintains a dynamic, three-dimensional 'mental model'[128] of its immediate surroundings, which is continually updated based on new sensory information. This includes (i) model of the workspace, i.e. a round tabletop, built into the initial state of the mental model (ii) a model of its own body, consisting of a set of four cylinders connected by joints to mimic the shape and range of the agent's physical positions (which is continually checked against its own joint angles and updated to match), (iii) models of objects situated on the work surface (each described by its position, orientation, shape, colour, mass, and velocity), and (iv) a model of the human interlocutor's body (represented as a simple sphere). At any point, the complete state of the mental model consists of descriptions of all the objects (Roy et al., 2004).

This model gives Ripley a sense of object permanence, noting the position of objects, represented as a multimodal sensory expectation, even when they are out of its immediate perceptual field: "if the robot looks at the appropriate location, its visual system expects to find a visual region; if the robot reaches to the same location, it expects to touch and grasp the object" (Roy, 2005:393). Thereby, the mental model allows it to ground the meaning of verbs, adjectives and concrete nouns using a single, unified representational framework—defined in terms of richly structured sensorimotor representations, i.e. percepts, actions, and affordances (Roy et al., 2004:1379). For instance, adjectives like *heavy* and *light* are defined in terms of their relative ease of being lifted (Roy, et al., 2004:1379). The model takes a connectionist approach, where the most activated function determines which (category) label is assigned. However, most aspects of the lexical structures are manually coded beforehand: only the activation functions (e.g. weight distributions associated with *light*, or colour distributions associated with *red*) are learnt from examples via statistical methods (Roy et al., 2004:1380).

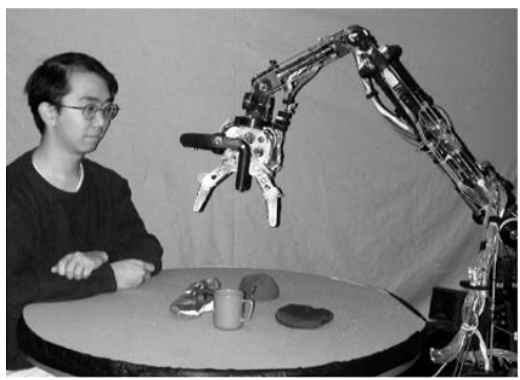

**Fig. III.1**: Ripley perceiving objects on a tabletop (Roy et al., 2004:1377).

---

[127] Roy et al (2004:1375) also describe it as an 'articulated torso' with a 'mouth (gripping mechanism). It has seven degrees of freedom, enabling it to manipulate objects in a three-foot range.
[128] *Mental model*, here, refers to what is essentially a 'cache' of the agent's surroundings as constructed through its perceptual system (Roy et al., 2004:1375).

Another important aim of Roy et al.'s 'mental model' approach is to account for the role of situational context—at least in the physical sense—in language: "As the degree of shared context decreases between communication partners, the efficiency of language also decreases since the speaker is forced to explicate increasing quantities of information that could otherwise be left unsaid" (Roy et al., 2004:1374). Not only does common ground make communication more efficient (by saying things less explicitly), but also helps to resolve ambiguities. In cases where the referents that feature in an instruction are ambiguous, Ripley uses a simple dialogue method to ask for further descriptive terms (within the range of attributes it is capable of processing).

In maintaining a certain level of awareness of the presence and activities of itself and other surrounding entities, including salient aspects of its recent perceptual history, Ripley has a certain range of contextual factors to facilitate the interpretation of simple situated natural language commands, like 'Hand me the cup behind the red ball' (Roy et al., 2004:1374). By representing its external surroundings as a stable backdrop, to which its movement is relativised, it also maintains a stabilised conceptualisation of the external world—which is assumed in much of our use of language. For instance, we normally do not describe objects as moving when we know that our own relative movement caused it to appear that way. We also talk about objects (being in certain locations) even when we are not directly perceiving them at the time (Roy et al., 2004:1374). Moreover, using its sense of awareness of its own location, as well as that of an interlocutor, Ripley is programmed with the ability to use its 'mental model' to interpret spatial expressions from different perspectives, e.g. *your left* versus *my left*—the latter triggering a 'shift of perspective' to the speaker's point of view (Roy, 2004:1375).

For a final point on interaction; some researchers have also started exploring the use of non-verbal cues, like gestures and facial expressions to facilitate human-robot interaction. For instance, Lallée et al. (2013) propose a robot that can utilise a combination of gesture, gaze and speech in the execution of cooperative tasks with a user. For example, as the robot said "You put the box on the left", its gaze was directed in the direction of *you*, *box* and *left* respectively, as they were spoken (Lallée et al., 2013:131). Tasks were also executed using gaze alone, where the human observed as the robot uncovered an object; the robot gazed at the object; and then at the human, to indicate that the human should grasp it; and then gazed at the location where the object should be placed (Lallée et al., 2013:130). This experiment also emphasises the utility of gaze in expressing and interpreting intention. Martinez-Hernandez and Prescott (2016) also explore the use of touch as a non-verbal form of communication: parts of a robot were covered in artificial 'skin' (containing distributed pressure sensors) that is able to measure pressure and duration of touch. A Bayesian framework is used to recognise different (prespecified) types of touch gestures (i.e. *soft*, *hard*, *caress*, and *pinch*), which then lead to different 'facial expressions' on the robot (i.e. *happiness*, *anger*, *shyness*, and *disgust*), which are expressed through a combination of 'eyelid' movement and arrays of LED lights located in its face (see **Fig. III.2**).

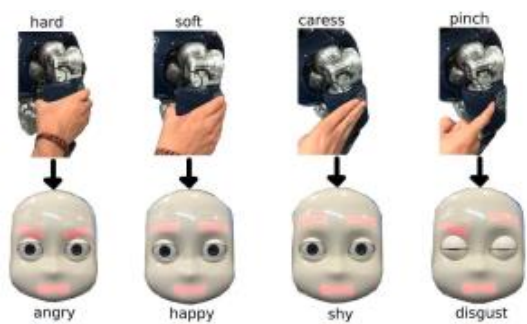

**Fig. III.2**: Reactions to different forms of touch (Martinez-Hernandez & Prescott, 2016:5)

As seen, researchers in AI have been making significant progress in modelling the interactions between language use, action and perception in real-world scenarios; mainly in terms of a limited prespecified set of observable properties like colours, shapes, spatial language; and the execution of basic actions in simple table-top environments. As yet, such models remain at a very basic stage and address only a small fragment of language use—there remain many difficult challenges for including other linguistic aspects such as grammatical composition and the goal-directed use of language in social/cultural contexts (Roy, 2005:394).

Whilst robotic ('embodied'), voice-controlled approaches for grounded language learning may, in some regards, seem more realistic than strictly text-based approaches (or their image-based enhancements), the amount of learning tends to be very constrained, either by the extensive use of hard-coded grammars and pre-specified components to facilitate the grounding process, or by simplifying the setup and environment. As such, their functionality typically does not generalise or scale beyond very limited domains (e.g. Ziemke, 1999; Hermann et al., 2017). Given all the complexities involved in real-world embodied and situated communication, some researchers have turned to computer simulations to explore more natural, autonomous grounded language learning approaches on very basic levels, the final approach I discuss. In contrasting the different capabilities of the constrained 'human-like' and more autonomous 'simulated' approaches, it will be easier to evaluate their respective limitations, in the final section.

### 1.3.3. Grounding using simulation

Inspired by aforementioned work in behaviourism, Hermann et al. (2017) present an agent that learns to bootstrap and generalise semantic knowledge on a purely empirical basis: by being rewarded for successfully carrying out written commands. Rather than a video-based or table-top scenario, their experiments take place in a continuous virtual 3D environment where an 'agent' explores its surroundings as a constant stream of pixel-based visual input, and is given simple textual instructions for finding and picking up objects that fit a given description—potentially in any language. Actively learning through a combination of reinforcement learning (i.e. learning to execute tasks successfully via reward-based trial-and-error) and unsupervised learning (i.e. extracting patterns without being pre-trained on annotated examples), and starting off with little prior knowledge, the agent learns to ground linguistic symbols in its emergent, low-level perceptual representations of the environment and relevant sequences of actions. The agent is then able to apply the language it learnt to interpret novel instructions in unfamiliar environments—exponentially increasing the speed at which it learns new words as its semantic knowledge expands, as is observed in human children (Hermann et al., 2017:1).

Whilst the overall configuration of the environment, and objects within it, can be customised, the precise world experienced by the agent (i.e. the different instantiations of objects, their colours, relative positions, surface patterns, and the environment's general layout) is randomly picked from billions of possibilities, allowing for much more variation than the aforementioned table-top solutions—which also helps agents to abstract the right properties (addressing the issue of *overfitting* mentioned earlier). To enhance the agent's ability to match the appropriate visual and linguistic input, the agent is equipped with a 'word-prediction objective' that estimates words $l_t$ given the visual observation $v_t$ (Hermann et al., 2017:7). This is arguably similar to our common capacity to resolve linguistic ambiguities on the basis of expected referents in a situated contextual conversation.

These experiments mainly take place in a very basic environment consisting of two connected rooms, with two objects in each. To train the agent to learn the meaning of simple referring terms, the virtual environment is configured so that the objects match the descriptions to differing degrees (e.g. same objects, different colours; same colours, different patterns, etc., placed at varying positions). Through thousands of trial-and-error iterations, the agent may start to infer an algorithm that abstracts the correct attribute. Even such a simple referencing task involves a lot of complexity, as maximising rewards across multiple episodes of training requires the agent to "efficiently explore the environment and inspect candidate objects (requiring the execution of hundreds of inter-dependent actions) while simultaneously learning the (compositional) meanings of multi-word expressions and how they pertain to visual features of different objects" (Hermann et al., 2017:4). Once an appropriate grounding has been abstracted, the agent may then generalise it to deal with novel situations, such applying a familiar predicate to novel objects, or interpreting novel instructions in terms of the productive/systematic composition of words and short phrases—paying mind to word order, to execute tasks in the right sequence (Hermann et al., 2017:4-13). They also taught it to learn inter-entity relationships (i.e. *next to* and *in room*), which again relied on the ability to interpret words in the correct order.

In the experiment shown in **Fig. III.3**, the agent starts in position 1 and is given the instruction "pick the red object next to the green object". It is rewarded 1 if it moved to and selected the correct object and −1 if it picked the wrong one. A new episode started after the agent selected an object, or if it failed to select any object after 300 steps (Hermann et al., 2017:5-7).

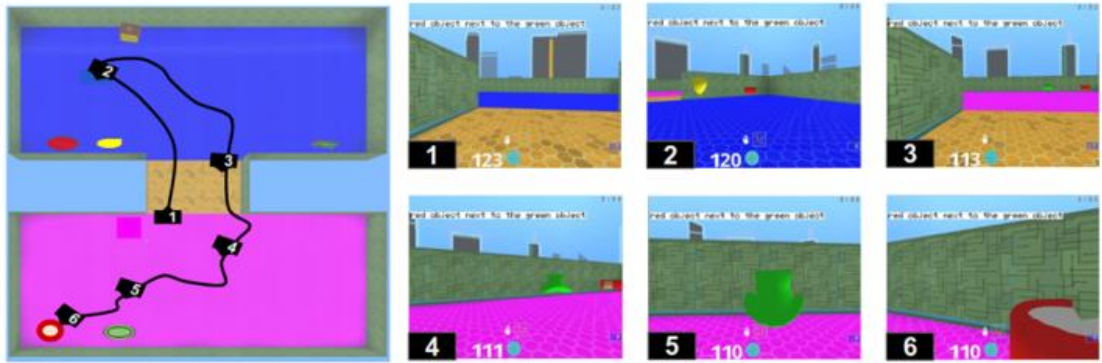

**Fig. III.3**: A top-view of an environment (left), and from the agent's perspective (right) (Hermann et al., 2017:5).

They used a similar strategy to let the agent learn the meaning of relational concepts, specifically relative size (SMALLER/LARGER) and relative shade (LIGHTER/DARKER). Regarding the former, after extensive training, the agent was able to naturally apply notions of relative size to novel shapes with almost perfect accuracy (Hermann et al., 2017:11). However, its ability to generalise lightness and darkness to unfamiliar colours was "significantly above chance but less than perfect" (Hermann et al. (2017:11). Hermann et al. (2017:11) explain that it is especially difficult to interpret *lighter* and *darker* (as we understand it) based purely on an RGB colour space. I would suppose that this may have to do with our real-world, situated experience of lightness and darkness—given the ways in which colour is affected by relative exposure to (sun)light in the real world, or how other substances behave (e.g. the darkening of colour as a liquid increases in density)—rather than being obvious distinctions to draw between types of colours *per se*.

To execute these tasks without any prior programming or hand-coded knowledge, the agent had to first learn different basic skills and capacities beyond the given instruction, including an "awareness of objects as distinct from floors or walls; some capacity to sense ways in which those objects differ; and the ability to both look and move in the same direction" (Hermann et al., 2017:10). Moreover, the agent must infer that solving tasks requires an integration of both visual and linguistic input, without being designed to assume the need for inter-modal interaction. Given the complexity of such learning challenges, the agent requires thousands of episodes of training before an emergence of word learning is evident (Hermann et al., 2017:10)—making it much more complex, expensive and time-consuming to attempt in the real world.

Another influential (and philosophically compelling) example of using agents in a virtual environment to ground language, is Steels' (1995, 1996, 1998) development of an interactive 'language game' (also later called *the naming game* and *discrimination game*) between virtual 'agents' that collaboratively learn to refer to entities in their shared (virtual) environments, using arbitrary labels. Inspired by Wittgenstein's use of the term, as well as other usage-based theories of language, Steels' theoretical framework can be summarised by two hypotheses: that (i) language is an autonomous, and dynamically adaptive system of shared conventions that acquires structure through a self-organising (coordinating) social process, and (ii) language spontaneously gains complexity, driven by the need to optimise communicative success within contingent constraints (such as time limitations, error-prone acoustic transmission, differences in background knowledge between interlocutors, etc.) (Steels, 1995:320).

In Steels' (1995) original experiment, a set of agents are distributed in a 2D grid, and may alter their position at any point. Every agent has a particular orientation (i.e. north, south, east, and west,) and is assumed to have a subjective 'visual field', i.e. direction of orientation, with which it can 'perceive' basic spatial relationships between itself and the other agents in the environment (although, in truth, their *perception* is mere coded information). An agent can use this perception for two purposes: to determine the direction in which another entity is located (for the purpose of referring), and to determine which entities satisfy a given spatial description (for the purpose of interpreting reference). Agents share a situated context in which they can all 'perceive' the relationships between the positions of different agents, although these are perspective-relative. Thereby, the challenge for the agents is to spontaneously develop a new vocabulary to identify one another using labels and spatial descriptions (i.e. *side, front*, *behind*, *straight, right* and *left*) relative to each agent's own position (and orientation). In Steels (1995), however, all agents are orientated north, for the sake of simplicity.

The basic dialogue structure is as follows: one agent (*the initiator*) selects another agent (i.e. *the topic*) out of a set of other agents, together constituting the conversational context. This can proceed in one of two ways: either the initiator can identify it using language (i.e. a label, consisting of two alphabetic letters), or it directly indicates ('points to') the agent in question (i.e. a written expression in the form 'point at a-X'), followed by the label. The initiator then repeats only the agent's unlabelled name (i.e. 'a-*X*'). Another agent, *the recipient*, uses language to identify the chosen topic (i.e. gives the label that he has for that agent). Finally, the initiator indicates whether the receiver succeeded in getting the label right (i.e. 'yes' or 'no') (Steels, 1995:320-321)—effectively creating a reward-based learning process, akin to reinforcement learning.

This process can be extended to pick out objects using the aforementioned spatial descriptions, rather than object names. In this case, each agent already has some shared knowledge of the names of other agents, and use that to determine the names for different spatial descriptions. For instance, the initiator may use itself as the viewpoint of the description: it identifies itself, first through 'pointing' (e.g. 'a-4'), and then giving the label (e.g. 'F A'). It then expresses that the agent of interest is to the left front of his location, by directly specifying that *left* = 'Z E' and *front* = 'G E'. The receiver then uses this information to determine which agent uniquely satisfies that description[129], using the shared name they already have prespecified for it (e.g. 'J O'). This is then confirmed by the initiator (Steels, 1995:322).

---

[129] In cases where more than one agent fits the spatial description, the participants may take turns asking further questions (e.g. is the object in this location?) to shrink the set of possible meanings (Steels, 1995:5).

After many iterations of trial-and-error, the agents many eventually converge on a similar vocabulary, which then allows them to leave out the initiation part for familiar commands (i.e. directly indicating the topic or original spatial terms) and use only the words in their lexicon to execute the dialogue. Steels explains that the vocabulary emerges spontaneously through a dynamic process of self-organisation:

> Self-organization is a mechanism very common in biosystems for reaching coherence among a group of distributed processes. It is based on the amplification and self-enforcement of fluctuations. The fluctuations consist here of random changes to the coupling between words and meanings. The changes are influenced by communicative success, leading to coherence (Steels, 1995:332)

That is, each agent is assumed to initialise and then continuously change its own vocabulary, influenced by communicative success: the greater the success the less likely a change. This creates a feedback-loop—as more agents start using the same word for the same referent, communicative success increases and thus the association between a given word and meaning becomes more stable and coherent, leading to a structured vocabulary (Steels, 1995:324). At first, most conversations do not succeed but meanings propagate, and eventually, words become established: most words (around 80%) get established after about 1000 conversations—although it may take more than four times as long, depending on the contingency of how often words (and their meanings) come up in conversations (Steels, 1995:329). Once there a group of agents have already formed a well-entrenched vocabulary, new agents that are introduced progressively adopt the same conventions (Steels, 1995:330).

Steels (1996) extends this basic method to a model where agents develop a language to refer to each other by means of their different (hand-specified) features (i.e. weight, size, and shape) and their values (e.g. heavy, square, medium, etc.), to which they assign arbitrary labels (i.e. random letter pairs). These features (in their original form) are expressed merely as lists of word pairs, for instance:

```
a-8:
    (weight heavy)
    (size medium)
    (shape round)
```

The agents then use a similar approach than earlier to describe each other; that is, by inventing new labels for the different features of the agent in question, until the agents finally converge on a similar vocabulary. The emergent languages do not have the full complexity of natural languages; for one thing, there is no syntax: 'conversations' merely consists of short interchanges of labels for different agents and features, agent identifiers, feature identifiers, and 'yes' or 'no'. However, the resultant vocabulary does involve expressions with multiple words, ambiguous expressions (i.e. having multiple 'groundings' for a label), and synonymous expressions (e.g. having multiple established labels for a feature/agent) (Steels, 1996:21).

Steels (1996) explains that his research was primarily motivated by scientific/theoretical incentives: to test theories regarding spontaneous language emergence (and its increasing complexity) in a linguistic community, which he considers to be "one of the crucial steps in the origins of intelligence" (1996:180). As a secondary motivation, Steels (1996:180) argues that, in understanding the mechanisms by which a language self-organises, we may come closer towards a bottom-up approach to developing autonomous, artificial intelligence.

Having discussed and compared the major current approaches to NLU, in the next section I summarise my findings in terms of the relative strengths and limitations of various models, specifically in terms of their relative suitability for grounding all of human language use—which I evaluate more thoroughly in the final chapter.

## 2. Chapter summary and reflection

In the previous section, I evaluated some of the major (grounded and ungrounded) approaches to enable natural language understanding in AI. Of these, each has its own relative strengths and suitability for certain tasks, in certain domains, and not all were motivated by the same (theoretical) aims. As such, my evaluation is not to discredit any approach overall, but merely for evaluating them relative to my own aim, which is to explore possibilities in replicating, to the greatest possible extent, human-like 'grounded' natural language acquisition and use in AI agents—one which is, in fact, explicitly shared in many of the approaches I discussed.

In what is perhaps the most basic approach to semantic analysis, (relational or logical) knowledge-based semantic parsing systems, on their own, can be sufficient for such language processing tasks that involve the parsing and processing of commands or queries with very limited complexity. The type using rule-based methods to translate natural language commands into a formal logical language representation might seem the most intuitive from a formal linguistic (and/or CCTM/LOTH) perspective. However, apart from its severe practical limitations—e.g. how time-consuming, and unreliable, it can be to construct extensive lists of rules and exceptions for interpreting ambiguous natural language utterances—that limit its functionality to very specifically-worded, simple commands in extremely narrow domains, the biggest threat to its plausibility (as a model of human language comprehension) is the symbol grounding problem, discussed in §2 of ch.I.

A similar criticism is faced by the type that relies on the use of *relational* knowledge bases, which, from a philosophical perspective, more closely resembles the pure associationist approach of conceptual structure (in the case of lists of real-world relations), or set theory (in the case of bases that indicate semantic relations like types and subtypes). Whilst statistical approaches that train models on the intended result of a given query/command may be easier to train, and more generalisable to different applications, they face their own limitations. That is, they not only face the general challenges of statistical (deep) learning systems that I listed in §2.2 of ch.I, but they are limited in their ability to interpret novel or ambiguous input—given their dependence on familiar examples from a given training set; as well as their lack of integration with real-world (background) knowledge; and commonsense reasoning (based on notions of causation, rather than mere correlation, between input and output). As such, they arguably still fail to 'ground' symbol meaning in an appropriate way.

Turning to vector-space semantics, this approach has been gaining a lot of attention for its success as a (disembodied) usage-based, distributional approach to semantics, both for its practical utility (mainly in modelling 'semantic' similarity) and theoretical grounding. Whilst this token-based approach that models semantics as a function of (linguistic) context seems more suited from a cognitive linguistics point of view (likewise embracing a usage-based approach to language), the graded nature of representations also seems more plausible from a psychological perspective in acknowledging the 'fuzziness' of contextual boundaries. However, models trained purely on linguistic data can only represent as much as is explicitly stated, which can often lead to biases (e.g. modelling objects in terms of atypical features, as those are more likely described), or fail to represent more obvious/concrete features (as people tend not to feel the need to state such features explicitly).

Another major benefit of word embeddings—as continuous numerical representations of features across multiple dimensions (that can be used for different purposes)—is that they are easy to combine with other embeddings of other modalities (like images, sounds, etc.), leading to a single vector representation that optimises across multiple (contextual) parameters at once. That is, they are a simple mathematical means for enabling the automatic extraction of context-sensitive multimodal representations, across vast amounts of data. Most approaches incorporating multimodal fusion has been limited to combining embeddings of text and 2D images, however, which, arguably, does not so much solve the grounding problem as merely *enhance* the word embeddings; i.e. by countering some of its aforementioned biases and shortcomings with information extracted from images.

On the other hand, whilst the visual embeddings may help to point out similarities that text-based embeddings miss (e.g. visual similarities between members of a category) they are likewise exposed to their own biases and limitations. These embeddings typically depend on the extraction of low-level features across whole images, based only on differences in pixel values, and, as such, they lack an understanding of images as representing real-world objects, as well as other important features like body-based affordances, and general background knowledge. Their utility is also largely limited to the modelling of concrete nouns, as they are unfit to deal with more abstract concepts or other linguistic units that have no clear pictorial representation, like verbs. Although verbs could be grounded in video sequences, it was argued that the sort of richly annotated video data their models require would be challenging and expensive to reproduce on a larger scale.

Yet, in principle, various more modalities can be combined, and some have argued that this may be a potential solution to comprehensive semantic grounding. However, I contended that, for a true human-like grounding of our full linguistic capabilities, this would require much more than the combination of basic modalities like sight and sound, but embodied experiences such as basic sensorimotor concepts, concepts of emotion, notions of time passing, folk-psychological understandings of other people, broader cultural/social knowledge, etc. That is, it would require meeting the basic requirements for embodied, embedded, situated, and/or enacted cognition, as discussed in chapter II, rather than the cognitivist idea of combining information from discrete input channels in a disembodied mind.

For this, I evaluated various robotic models that, to different extents, address issues of grounding word meaning in real-world, situated objects, regions, and/or actions. In different relative areas, the models I discussed demonstrated initial progress in multiple crucial aspects of human-like language learning and communication. This includes actively categorising novel object features (albeit limited to certain types) from a raw stream of cross-modal input; using gaze-tracking to help pick out (concrete visual) referents; using a combination of gaze and gestures to communicate nonverbally in a task-oriented scenario; acknowledging that different symbols can be used to describe the same object, depending on the property referred to; grounding more abstract concepts like prepositional phrases, relational concepts, cardinal concepts, and ordinal concepts; grounding verb meaning in terms of observed features, as well as embodied action patterns/control parameters; learning how to ground word meaning in a combination of action-affordances and perception; creating a dynamic mental model of an environment to learn notions of object permanency, a notion of the agent's own body, as well as that of an interlocutor; and learning a notion of speaker-relative perspective for interpreting directions.

However, I found that such systems are still in their early stage, and despite the tentative advancements of different models, each depends on strong prior constraints to allow for its functionality—mostly concerning the types of features they are able to recognise, the environment (most being constricted to a tabletop); and the reliance on the use of hard-coded prior knowledge or prespecified categories/symbols by which to classify sensory input. The use of such constraints are unsurprising, as rather than bootstrapping knowledge for an autonomous, inherently-motivated agent, these systems are basically complex classifiers and command-executors that have to assume all the needed prior knowledge to enable them to execute certain tasks 'like a human'; that is, a biological organism with particular perceptual and cognitive mechanisms that have evolved to survive in a particular environmental niche, situated within a particular social context, and who learnt how to effectively communicate their (particularly human) experiences, aims, and desires in a particular linguistic community. As put by Roy: "We cannot expect that such models and systems will directly explain how people think and communicate: both design and implementation differ dramatically" (2005:394). Thus, many constraints have to be put in place to ensure that the behaviours of these systems remain within the desired boundaries.

To illustrate this point further, I also explored models that restrict their learning to a simple simulated environment where they can learn how to meaningfully bootstrap basic linguistic knowledge with their own goal-directed exploration of the virtual world, building semantic knowledge on thousands of reward-based attempts. This bootstrapped approach takes the other (cognitively implausible) extreme of assuming almost no prior knowledge, cognitive/perceptual biases, biological instincts, or cultural constraints, etc., which naturally takes a lot more attempts to learn than humans who have certain embodied and culturally embedded/situated priors in common. In this category, I investigated two approaches. The first uses reinforcement learning to learn how to abstract certain (again, prespecified) types of visual features (i.e. shape and colour), in simple simulated worlds, with objects placed in different locations, and with different visual features (object/wall/floor colours and object shapes) chosen at random from billions of possibilities. This process is iterated until the agent learns how to map basic instructions (potentially in any language) to the right features in the environment, with anything it 'observes' being possible candidates. Although the process is less constrained than in the real-world table-top scenarios, the setup is unrealistically simple and restricts learning immensely to very basic properties that can be inferred from pixel patches alone.

The second example used a simple coordination method ('language game') between virtual 'agents' to refer to entities in their shared (virtual) environments, using arbitrary labels. Similar to the previous approach, agents learn through thousands of trial-and-error episodes, changing their labels based on their collaborative attempts to autonomously converge on the same names for referents. Whilst this approach explores another realistic bottom-up approach to language acquisition, its utility is mainly theoretical, as neither the 'language' (consisting merely of two-letter labels, with no grammar), nor the 'environment' (defined merely in terms of agents and locations), is much like ours. Moreover, this approach also avoids all the difficult problems related to perception in noisy, dynamic environments, as agents' *perception* merely amounts to coded information of other agents' locations (or their 'features', as lists of words). To apply such a coordination-based approach to language learning in the real world[130], it would reintroduce all the difficulties of embodied learning from the previous examples, such as sticking to the vocabulary (and grammaticality) of an existing natural language, and therewith all of the basic species-specific embodied biases/experiences of humans upon which it was founded (e.g. emotions, the right perceptual biases, sensorimotor capabilities, etc.)—not to mention the problem of reference determination in a complex real-world environment and sociocultural context.

On the basis of my discussion of current major NLU approaches, in the following chapter I offer a more thorough evaluation of the grounding problem introduced in chapter I, using the theoretical insights I gained in chapter II, to explore possibilities for addressing it in AI.

---

[130] See Bleys et al. (2009) for such an attempt.

# Chapter IV: Critical discussion and conclusion

In this chapter, I integrate my findings from the proceeding chapters to evaluate some of the basic requirements for grounded language learning in AI, as well as possible challenges involved in meeting them. On the basis of my discussion of different key approaches/models in NLU in the previous chapter, I revisit the grounding problem introduced in §2 of ch.I, to evaluate different theoretical understandings of the problem, and, accordingly, different (practical) approaches to addressing it. From there, I offer a critical reflection regarding which desiderata remain, and which implementational constraints might need to be put in place in order to enable sufficiently similar grounding of as much as possible of human language use. This is followed by a section in which I evaluate possible challenges in achieving this aim. Finally, I conclude the thesis with a summary of my work in terms of the overall structure and key points of my argument.

## 1. Towards grounded language learning in AI

In the first chapter (§2.2), I expressed that most theorists that take seriously the problem of grounding basically agree on two points: that escaping this internalist trap is 'crucial to the development of truly intelligent behaviour', and that grounding requires agents to be causally coupled with the external world in some way (without the mediation of an external observer), but that the nature of this coupling is disputed. In the first subsection that follows, I evaluate the nature of this dispute more in detail, informed by the various approaches to 'grounded' NLU I discussed in the previous chapter. I also argued, in the first chapter, that an embodied view of cognition might still be compatible with a connectionist approach, as long as the input signals are appropriately 'embodied', and connections are structured in an appropriate way—and that, as such, there might still be hope in, theoretically, modelling aspects of human cognition artificially, although that a total modelling might be more of a theoretical than practical possibility. The processes required, and the extent to which it may be practically feasible, is what I explore in this section—particularly for the sake of grounding natural language use in AI, in a human-like way.

### 1.1. Rethinking grounding

From what I gathered of my findings in chapter III, most of the (*cognitivist* and *embodied*) models I discussed seem to take the view that 'grounding' amounts to something like a simple classification issue: knowing which percepts to map onto which written/spoken linguistic symbol. However, recall from §2.2 of ch.I:

> The grounding problem is, generally speaking, the problem of how to causally connect an artificial agent with its environment such that the agent's behaviour, as well as the mechanisms, representations, etc. underlying it, can be intrinsic and meaningful to itself, rather than dependent on an external designer or observer. It is, for example, rather obvious that your thoughts are in fact intrinsic to yourself, whereas the operation and internal representations of a pocket calculator are extrinsic, ungrounded and meaningless to the calculator itself, i.e. their meaning is parasitic on their interpretation through an external observer/user (Ziemke, 1999:87).

That is, more than merely performing calculations over (multimodal) inputs, a key aspect of the grounding problem, in its original formulation, is making the *meaning* of symbols *intrinsic to the system itself*, and not just seem as such to an external observer (who imbues them with meaning instead). Whilst cognitivist models have approached this problem by merely integrating different modalities, e.g. mapping features in images (pixel data) to symbols, embodied approaches arguably took the same approach a step further with embodied agents that learn map (spoken) natural language symbols to perceptions of objects (and their features, relations), regions, and actions in a real-world situated environment. However, according to Ziemke (1999), this does not suffice, as making something 'intrinsically meaningful' to the agent requires (what he considers) an appropriate understanding of 'situated' and 'embodiment':

- Natural embodiment is more than being-physical. In addition to that it reflects/embodies the history of structural coupling and mutual specification between agent and environment in the course of which the body has been constructed.
- Natural situatedness is more than being physically connected to your environment. It also comprises being embedded conceptually in your own phenomenal world (Umwelt), which is also constructed in the course of the above history of interaction, in congruence with sensorimotor capacities as well as physiological and psychological needs (Ziemke, 1999:97).

In other words, on this view, it is not enough for an agent to merely combine inputs from different modalities, nor to be artificially 'hooked' to an external environment to optimise over multimodal parameters (within given constraints). Hence, even if a system might exhibit the 'right' behaviour, it does not suffice to solve the grounding problem. For that, Ziemke (1999) argues that an agent, as an *integrated and coherent system*, should be embedded in a specific environment that has been made meaningful to it *in terms of* its particular embodied experiences/'phenomenal world'. That is, "a subjective abstraction, interpretation or constructed view of the physical environment that fits the agent's sensorimotor capacities and its physiological and psychological needs" (Ziemke, 1999:96). Without the natural emergence of phenomenal experiences and intrinsic needs (between a co-evolving organism and its environment), he argues that the system "has no concept of what it is doing or what to use the produced labels for, i.e. it is not embedded in any context that would allow/require it to make any meaningful use of these labels" (Ziemke, 1999:90).

On this understanding, organisms are naturally "tailor-made" to act and perceive things in (intrinsically) meaningful ways, through their co-evolution/mutual determination with the environment within which they are embedded—an approach, he argues, that has been largely neglected in embodied AI research (Ziemke, 1999:177). I will refer to this as grounding in the *strong* sense. In contrast, many AI researchers (particularly in NLU) have reinterpreted 'grounding' in operational ways that better suit the practical needs of their users. According to Hudelot et al. (2004), these reinterpretations can typically be distinguished between *the anchoring problem* in robotics, i.e. "the problem of creating and maintaining the correspondence between symbols and sensory data that refer to the same physical object" (Hudelot et al., 2004:2); and the *semantic gap problem* in image indexing/retrieval, i.e. "the lack of coincidence between the information that one can extract from the visual data and the interpretation that the same data has for a user in a given situation" (Hudelot et al., 2004:2).

In essence, what it seems researchers in NLU are attempting to achieve is to *mimic* the particular human-like perceptual and cognitive mechanisms members of our species have evolved through our natural development within a cultural/physical environment—or at least those mechanisms that seem most relevant for the specific language-based task at hand. On the other hand, some researchers in AI are focused more on the issue of autonomy, and, as in the simulation approaches, would like to create agents that develop their own behaviours (and/or means to communicate) though goal-directed, autonomous learning, using their own means of perception and embodied actions.[131] To get the kind of science-fiction inspired AI agents that are *both* autonomous *and* human-like, is an incredibly hard problem that will probably not be solvable until we are able to account for the majority of complex evolved biological mechanisms and processes that make the world intrinsically meaningful to us *in our particular way*—essentially making something of an artificial human clone (like the androids in *Westworld*). That is, because, as discussed earlier, even slight changes in embodiment can have profound effects on an agent's phenomenal experiences of the world. As yet, we do not even understand most of our own physiological/psychological processes, and even if we could succeed in such an ambitious aim as modelling them artificially, it is unclear what the purpose would be (raising the question of whether we could/should treat 'sentient' or *feeling* beings as tools).

Therefore, in keeping with the tradition, I will turn my focus to symbol grounding in the weaker, operational sense; anchoring meaning in real-world phenomena so that it can successfully translate a spoken command into a desired response. For this, I evaluate the desiderata for 'grounding' as much as possible of human language use (drawing from my insights in chapter II) in relevant factors that can be empirically inferred from context by an AI agent, and what the challenges may be for such a weaker reading.

## 1.2. Grounded associative language learning

Perhaps the most important factor for symbol 'anchoring' would be perceptual mechanisms that are sufficiently similar to ours; that is, with the same biases (e.g. gestalt principles, visual illusions/preferences, focal points, focal scope/range, etc.) and constraints (e.g. seeing the same spectrum of colours, learning the difference between light and dark)—and likewise for other kinds of perception like audio (hearing a similar range of frequencies), haptic (e.g. artificial skin[132] that can measure touch or temperature, with prespecified knowledge of which thresholds are extreme for humans), etc. Although it may be possible to encode other senses like taste, those are more experiential than importantly useful for reference determination. Regarding interoception: rather than 'experiencing' affective states, which, for instance, enable us to emphasise with others and predict their intentions/internal states in communication (as discussed in §2.3 of ch.II), a robotic agent should at least be able to ground affect-concepts in features from patterns of others' behaviour (e.g. tone of voice, gestures, word choices, and other subtle behavioural patterns). This could potentially even include the integration of the use of smart devices to measure vitals like heart rate, temperature, etc—which could also be useful for social robots that need to measure when their users are in distress and they need to signal for help (e.g. robots that assist the elderly—see Huang & Liu, 2019, for instance).

---

[131] See Trianni (2008) and Waser (2014) for approaches towards 'autonomous' (and non-humanlike) robots.
[132] That is, pressure and/or heat sensors, as used in Martinez-Hernandez and Prescott (2016).

Another important aspect highlighted by Roy (2005), is that an agent should have the ability to model its surroundings, including itself and others, to retain (and continually update) in memory, and use to imagine things from other perspectives. This could also be useful for grounding basic pronouns like *I*/*my*/*we*/*you*/*us*/*they* etc. Another point, raised by many NLU researchers, is that agents should be able to learn novel concepts *directly* (i.e. without pre-training or hardcoded knowledge), corresponding to a potentially unlimited range of objects and properties, bringing up the issue of reference determination from before. For this, agents should be able to estimate the most probable referent, given the current physical, linguistic, and hopefully social context—although the latter might be a bit more difficult without being grounded in the stronger sense. Nevertheless, even without a 'deeper' understanding of context, distributional embeddings demonstrate that much about context can be learnt purely empirically—further enhanced by the integration of many human-like modalities (such as those listed above)—using methods for multimodal fusion, or even extracting multimodal information directly using connectionist feature-extraction methods. Such a usage-based empirical learning of the (integration of) specific contexts wherein words are used, also seems to fit in with the encyclopaedic view of meaning described in §3.2.1 of ch.II (e.g. to learn that a different shade of red applies in different contextual uses, like *red skin*, *red fox*, *red lipstick*, etc.). Structurally, this is akin to the connectionist/associationist view of conceptual structure I put forward earlier, where concepts are understood as patterns of basic multimodal features abstracted from instances in given contexts.

Beyond concrete objects and properties, however, agents should also be able to ground words relating to actions, events, measurements of time (e.g. *later, tomorrow*, *summer*), ordinality, cardinality, relations (e.g. bigger/smaller, left/right, over/under, inside/outside), and certain other (useful) abstract concepts (e.g. *time*). Moreover, if what Lakoff and Johnson (1980) argue is accurate, about much of our concepts being structured in terms of our basic (distinctively human) embodied experiences, what would also be useful is a sufficiently similar physical embodiment. That is, a body within the range of human-like shapes and sizes, with a vertical orientation, and a capacity for basic human-like sensorimotor movements. This is also useful from an enactivist perspective, to enable an agent to perceive human-like affordances in objects—grasping their function as an important part of their meaning to us, as well as to help interpret which properties of an object are relevant for the particular goal-directed discussion (as described in §2.2 of ch.II). This is also emphasised by the aforementioned simulation theory of semantics (§3.2.2 of ch.II), which claims that correlations between words and their affordances for action are an important part of what makes them meaningful to us, and hence, is important for a grounded interpretation of utterances. According to Noë's sensorimotor theory[133] discussed in §2.2 of ch.II, the ability to move around in a 3D environment and view objects from certain perspectives, is also what allows us to make predictions about objects as *whole* or unified entities. Computationally, this might potentially be addressed with research in the area of *predictive processing*[134], although a description of this approach is beyond the scope of this thesis.

This is not to say that having a specific 'human-like' bodily shape and perceptual capacities is necessary for language grounding, as even within the human species there is great variation in size, shape, abilities and other capacities, and this variation does not necessarily make language comprehension impossible. However, as AI systems are only grounded in the weaker (symbol-anchoring) sense, perception and sensorimotor capacities are the main tools they have at their disposal. That is, they take an empirical approach (without the other aspects of *stronger* grounding like inherent physiological/psychological needs or affective states) that, arguably, significantly limits their ability to gauge speaker intention (based on the discussion in §2.3 of ch.II). As such, I would argue that the more 'typical' such perceptual mechanisms can be, the better.

Finally, beyond grounding symbols, a significant challenge that remains is to enable agents to learn how to construct, and interpret, sentences in terms of grammar. That includes understanding the logical relations that hold between different units in a sentence, as well as how to interpret the meaning of compositions that do not necessarily reduce to the meanings of their parts (recall the *steak knife* example from before). Therefore, rather than simply learning the meanings of individual symbols and combining them systematically, agents should learn the contextual combinations of symbols to produce familiar phrases, or figurative expressions,[135] or how different aspects of a feature is relevant in different combinations (as I argued earlier). Research of this sort is already being explored in compositional distributional semantics (e.g. Sadrzadeh & Grefenstette, 2011; Baroni, 2013). In an earlier paper, I evaluate the difficulties of processing syntactical ambiguities more in-depth (Alberts, 2019).

This list is by no means exhaustive, but is merely a rough sketch of the basic requirements for grounded language learning by AI agents, assuming a weaker sense of the word (as described above). Even so, there remain some significant practical challenges, which I evaluate in the following subsection.

---

[133] That is, a theory that accounts for our ability to perceive objects as whole and three-dimensional, despite our narrow perspectives, in terms of our ability to grasp how our actions can allow different aspects of entities to enter into our field of vision; grasping how *what we can do* affects *what we see*.

[134] See Clark (2013), Hohwy (2016), and Wiese and Metzinger (2017) for an introduction into the programme.

[135] This utterance, rather than symbol-based approach to grammar is also the one taken in cognitive linguistics (see Evans, 2019:127).

### 1.3. Challenges

Even with as many human-like perceptive/behavioural constraints in place as possible, and sophisticated multimodal usage-based learning algorithms, it is still unclear how much of language could be effectively interpreted using purely empirical methods for 'grounding' meaning. Taking Lakoff and Johnson's conceptual metaphor theory as an example, they acknowledge that their descriptions of *which* of our basic bodily experiences influence our structuring of which concepts *in which ways*, is merely speculation (Lakoff & Johnson, 1980:17). As such, there may be many more conceptual metaphors that they failed to recognise as such, as, for us, "no metaphor can ever be comprehended or even adequately represented independently of its experiential basis" (Lakoff & Johnson, 1980:17). Hence, basic similarities in visual/haptic/auditory perception and sensorimotor capacities, implemented a robot with a human-like bodily shape and size, may not constitute a sufficient basis to structure all the conceptual metaphors we employ in language. Moreover, as yet, learning how to metaphorically structure concepts by means of others is a significant challenge in itself that has not been solved in AI—although some are aiming to work in that direction (e.g. Bailey et al., 1997, using their x-schemas described earlier).

In addition, there lies a significant challenge in dealing with the 'graded grammaticality' of natural language utterances, described earlier. Recall from §3.4 of ch.II, that, in cognitive linguistics, a (culturally and contextually embedded) utterance is considered *unit-like* in that it represents the expression of a *coherent idea*, involving a combined range of linguistic and non-linguistic strategies. As such, speakers often find it easy to omit some of the grammaticality requirements of a well-formed sentence. This is performed in context by a member of a particular linguistic community in order to achieve a particular interactional (set of) goal(s). Thus, as I mentioned, the interactional and goal-directed nature of language use, as well as its situated context—which interacts with the speaker's intentions and scaffolds interpretation—are all considered central to the usage-based account that cognitive linguists assume. Again, this draws upon (*but is not limited to*) the conventional (coded) meaning of a particular word/construction, which, in cognitive linguistics, is understood as an idealised abstraction from a range of contextualised (pragmatic) uses.

This points to an important limitation of a weaker (anchoring) version of symbol grounding: with grounding in the stronger sense, the hearer might be able to extract the *coherent idea* communicated by means of an utterance, through a combination of capacities like experiencing simulations, drawing inferences from linguistic/physical/social contexts, reading non-verbal behavioural cues, including other social cognitive capacities to *empathise* with the speaker's mental state/intentions (recall 'mirror neurons' described in §2.3 of ch.II). Whilst some of these might be approximated empirically (like learning correlations between words and different linguistic/physical contexts), it is unclear how much of our inferential capacities depend on our ability to *empathise* with the needs, desires and intentions of others of our species. Moreover, without the ability to emphasise with the psychological/physiological needs involved in goal-directed communication, nor with the experiential content of interoceptive concepts (e.g. *love*, *hunger*, *sadness*), tasks like analysing sentiment (and inferring non-literal meaning like sarcasm) is notoriously difficult for AI systems (see Chakriswaran et al., 2019).

Turning to concepts that can be inferred from direct perception, most of the papers I evaluated acknowledge the complexity of labelling the correct referent amongst many possible objects, properties, relations, etc. in conversation—a task that is also significantly more difficult without the ability to grasp the deeper intentional meaning of utterances, as if grounded in a stronger sense. From a computational perspective, a robust approach might be something like the simulated reinforcement learning-based approaches whereby an agent learns to abstract the right feature from thousands of examples that vary in all ways but one; however, to apply this in the real world would be incredibly expensive and time-consuming.

To reiterate, these general challenges are not necessarily damning for the field as a whole, as the challenges and requirements depend on the particular aims of any given application, which tend to be more limited in scope. However, various researchers have emphasised the need for designing a more general capacity in AI systems to ground natural language utterances in the real world, specifically to meet the increasing demands for social robots to enter our homes and workspaces and execute tasks like assisting the elderly, keeping people company, or executing grounded instructions. To achieve such aims, and perhaps out of curiosity to see how far we can go towards developing agents with a human-like command of natural language, these points may be worth considering.

## 2. Thesis conclusion

In this thesis, I carried out a novel, interdisciplinary analysis into the various complex factors involved in human natural-language acquisition, use and comprehension, with the aim of uncovering some of the basic requirements that would need to be considered if we were to try and develop AI agents with similar capacities. Developing on previous work in which I explored the complexities and challenges involved in enabling AI systems to deal with the *grammatical* (i.e. syntactic and morphological) irregularities and ambiguities inherent in natural language (Alberts, 2019), I turned my focus here towards appropriately inferring the content of symbols themselves, as 'grounded' in real-world objects, properties, events, actions, etc. For this, I first introduced the general theoretical problems I aimed to address, discussing the co-development of AI and the controverted strands of computational theories of mind in cognitive science, and the grounding problem (or 'internalist trap') faced by them.

In order to unpack and address the issue, I offered an extensive critical discussion of the relevant theoretical literature; that is, the major philosophical/psychological debates regarding the nature of *concepts*; theories regarding how concepts are acquired, used, and represented in the mind—including my own account, grounded in current (cognitively plausible) connectionist theories of thought. From there, I aimed to gain a better understanding of the relevant embodied (e.g. cognitive, perceptive, sensorimotor, affective, etc.) factors involved in human cognition, which I discussed drawing from current scientific research on 4E Cognition and embodied social cognition. On that general basis, I focused more specifically on grounded theories of language, drawing from the cognitive linguistics research enterprise that aims to construct a naturalist, cognitively plausible understanding of human concept and language acquisition.

Integrating my findings from these various disciplines, I presented a general theoretical basis upon which to evaluate more practical considerations for its implementation in AI. Here, I presented a critical investigation into the different major approaches in the area of Natural Language Understanding (as well as their integrations), and evaluated the respective strengths and shortcomings of different approaches, in terms of specific models. I then re-evaluated the grounding problem and the different ways in which it has been interpreted by theorists and AI researchers, distinguishing between a strong and weak reading. I presented arguments for why addressing the stronger version in AI seems, both practically and theoretically, problematic. Instead, drawing from the theoretical insights I gathered, I considered some of the key requirements for 'grounding', in the weaker sense, as much as possible of human language-use capability in robotic AI agents, including implementational constraints that might need to be put in place to achieve this through empirical means. Finally, I evaluated some of the challenges that may be involved, if indeed the aim were to meet all the requirements specified.

There have been various attempts in different disciplines to investigate the nature of human concept/language use and acquisition, and how this may be mimicked in AI systems—often with different disciplines largely diverging on their understanding of the issues involved, the appropriate means for addressing them, and their definitions of key terms. For a thorough evaluation, however, one arguably requires an integration of work in various relevant disciplines: as I have argued, more than linguistics alone, our embodied experience of the world is crucial for understanding our use of language. This requires research into our general cognitive/perceptual/sensorimotor/emotional capacities (as explored in 4E theories of cognition), as well as our species-specific capabilities to categorise, memorise, label and communicate our (sufficiently similar) perceptual experiences (as explored in cognitive linguistics). Beyond theory, to consider the possibilities of its implementation in AI requires an understanding of the technical capabilities (and limitations) in the field— as well as, arguably, some philosophical reflection on the implications. In this thesis, I aimed to bring all these major perspectives together (the philosophical, the psychological, the cognitive scientific, and the computational) to unpack some key desiderata for an interdisciplinary, cognitively plausible (and computationally tractable) model of human language acquisition and use, as well as the constraints for its practical implementation in AI—a project that, to my knowledge, has never been carried out in this depth. Whilst the results I offer are speculative, I hope that it at least serves to illuminate and some of the complexities involved, and some important aspects worth considering if one is interested in thoroughly addressing these issues—which might be crucial for future research into effective, real-world human-computer interaction.